%% file: main.tex
\documentclass{article}

\input{template}

\input{preamble}

\input{template_adapter}
\makeatletter
\renewcommand{\template@authormetadata}{%
  \centering
  \template@authorlist\par
  \vspace{0.35em}%
  {\normalfont\normalsize\scshape
    \template@affiliationlist\par
  }%
  \iftemplate@showresources\template@resources\fi
  \vspace{0.2em}%
  {\normalfont\small
    \makebox[\linewidth][c]{%
      $\dagger$: Project Lead\hspace{1.25em}
      \textcolor{black}{\Letter}: Correspondence
    }
  }%
}
\makeatother
\begin{document}

\title{Thinking Inertia: LLMs Keep Thinking When Told Not To}

\author[1]{Dianqiao Lei}
\author[2]{Kevin Qinghong Lin$^\dagger$\textsuperscript{\Letter}}
\author[3]{Pan Lu}
\author[2]{Philip Torr\textsuperscript{\Letter}}
\author[3]{James Zou\textsuperscript{\Letter}}
\affiliation[1]{Tsinghua University}
\affiliation[2]{University of Oxford}
\affiliation[3]{Stanford University}

\abstract{
Large Language Models (LLMs) increasingly ship with explicit ``thinking modes'', yet their counterpart, ``\nothink{}'', has received far less attention.
In this work, we study LLMs' ``\nothink{}'' behavior along two axes.
\textbf{a. How to measure \nothink{}?}
Prior work typically defines \nothink{} through proxies such as a disabled thinking mode or the absence of long traces.
These proxies are unreliable: disabled thinking modes still emit extended reasoning, or long traces often contain uninformative filler rather than genuine inference.
We instead decompose each LLM response into an answer and accompanying visible text, and evaluate it at three levels:
{(i)} an \ETRName{} for strict answer-only compliance;
{(ii)} an instruction-aware \QRelName{} to calculate similarity between the question and visible text;
and {(iii)} an additional LLM-as-judge \EIRName{} for visible explicit inference.
Together, these levels separate answer-only output, relevant but non-inferential text, and explicit inference. 
Human studies demonstrate the effectiveness of these metrics.
\textbf{b. How does no-thinking vary across tasks and models?}
We evaluate six prompting intervention settings on six widely used LLMs (both open-source and commercial models) across three question types: Boolean, multiple-choice, and open-ended.
We find that explicit no-think controls cannot reliably eliminate visible explicit inference; 
models instead exhibit \textbf{``Thinking Inertia''}: \textit{visible explicit inference persists even under strict no-thinking controls, and becomes progressively more prevalent as the answer space opens};
On the another hand, accuracy remains relatively stable on Boolean and multiple-choice tasks, whereas open-ended tasks expose a trade-off between answer-only compliance and task accuracy.
Finally, rewriting the same questions across answer spaces shows that supplying candidate answers makes answer-only responses substantially easier to produce.
Taken together, these findings establish no-thinking as a non-trivial capability of LLMs: stopping explicit reasoning cannot be assumed from model settings or instructions alone, and deserves systematic evaluation alongside reasoning ability itself.

}

\paperResources
\maketitle
\vspace{-0.4em}
\enlargethispage{0cm}
\begin{figure}[H]
    \centering
    \captionsetup{font=small,skip=2pt}
    \includegraphics[width=0.90\textwidth]{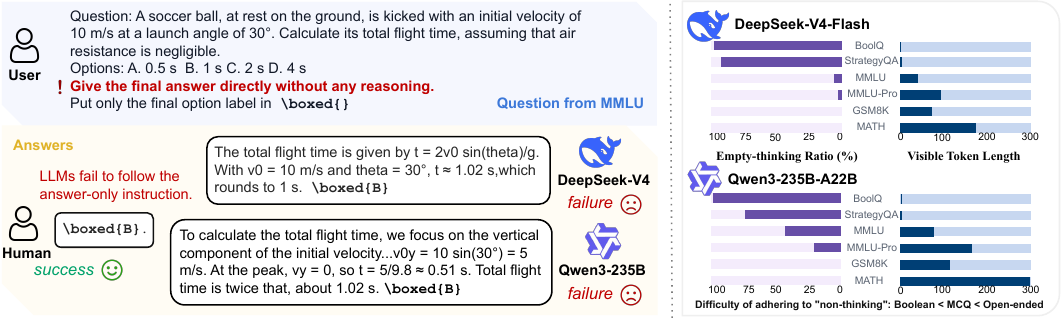}
    \caption{
    \textbf{No-thinking controls do not guarantee response-level no-thinking.}
    \textbf{(Left)} On a representative MMLU question, DeepSeek-V4-Flash and Qwen3-235B-A22B are asked for only the final option, yet expose question-relevant pre-answer text. Appendix Figure~\ref{fig:prompt-comprehension-app} confirms the instruction's clear answer-only reading.
    \textbf{(Right)} Under native no-thinking, green \ETR{} and orange visible-length bars across the six benchmarks show Boolean tasks closer to answer-only than multiple-choice and open-ended tasks.
    }
    \label{fig:motivation-teaser}
\end{figure}

\input{section/intro}

\input{section/method}

\input{section/exps}
\input{section/relatedwork}
\input{section/discussion}

\input{section/conclusion}
\clearpage
\printbibliography

\clearpage
\input{section/appendix}

\end{document}

%% file: template.tex
\usepackage{float}
\usepackage[margin=1.0in]{geometry}
\usepackage{abstract}
\usepackage[utf8]{inputenc}

\usepackage[dvipsnames,svgnames,table,xcdraw]{xcolor}
\usepackage{relsize}
\usepackage{tcolorbox}
\usepackage{listings}
\usepackage{subcaption}
\usepackage{comment}
\usepackage{graphicx}
\usepackage{booktabs}
\usepackage{multirow}
\usepackage{todonotes}
\usepackage{enumitem}
\usepackage{url}
\usepackage{mathpazo}
\usepackage{amsmath,amssymb,amsthm,amsfonts,bm}
\usepackage[toc,page,header]{appendix}
\usepackage{fancyhdr}
\usepackage[
  backend=bibtex,
  natbib=true,
  maxbibnames=99,
  minalphanames=3
]{biblatex}
\usepackage[colorlinks=true]{hyperref}
\usepackage[T1]{fontenc}
\usepackage{newunicodechar}
\usepackage{auth_detailed}

\usepackage{mathtools}
\usepackage{longtable,tabularx}
\usepackage{placeins}
\usepackage{threeparttable}
\usepackage{adjustbox}
\usepackage{pifont}

\counterwithout{figure}{section}
\counterwithout{table}{section}

\definecolor{logocolor}{RGB}{30,0,178}

\hypersetup{
  colorlinks=true,
  urlcolor=BrickRed,
  citecolor=ForestGreen,
  linkcolor=BrickRed
}

%% file: preamble.tex
\usepackage{array}
\usepackage{makecell}
\usepackage{wrapfig}
\usepackage{needspace}
\usepackage{etoc}
\usepackage{afterpage}
\usepackage{pdfpages}
\usepackage{pgfplots}
\usepackage{marvosym}
\usepackage{fontawesome5}
\usetikzlibrary{arrows.meta,calc}
\usepgfplotslibrary{groupplots}
\pgfplotsset{compat=1.18}

\newunicodechar{–}{--}
\newunicodechar{—}{---}
\newunicodechar{‘}{`}
\newunicodechar{’}{'}
\newunicodechar{“}{``}
\newunicodechar{”}{''}

\newcommand{\includepdfafterpage}[2][]{%
  \afterpage{%
    \begingroup
      \let\savednewpage\newpage
      \def\newpage{\let\newpage\savednewpage}%
      \includepdf[#1]{#2}%
    \endgroup
  }%
}

\makeatletter
\newcommand{\longdash}[1][2em]{%
  \makebox[#1]{$\m@th\smash-\mkern-7mu\cleaders\hbox{$\mkern-2mu\smash-\mkern-2mu$}\hfill\mkern-7mu\smash-$}}
\makeatother
\newcommand{\omitskip}{\kern-\arraycolsep}

\newcommand{\eg}{\textit{e.g.,}}
\newcommand{\nothink}{no-thinking}

\newcommand{\ETR}{ETR}
\newcommand{\QRel}{QRel.}
\newcommand{\EIR}{EIR}
\newcommand{\ETRName}{Empty-Thinking Rate}
\newcommand{\QRelName}{Question--Pre-answer Relevance}
\newcommand{\EIRName}{Explicit Inference Rate}

\tcbuselibrary{skins, breakable}
\newtcolorbox{findingbox}{
  colback=blue!5,
  colframe=blue!40,
  boxrule=0.5pt,
  arc=2pt,
  left=6pt, right=6pt, top=4pt, bottom=4pt,
  breakable
}

\definecolor{accHeat}{RGB}{80,145,210}
\definecolor{srHeat}{RGB}{236,150,72}
\definecolor{simHeat}{RGB}{90,170,105}
\definecolor{eirHeat}{RGB}{150,110,185}
\definecolor{dropHeat}{RGB}{145,115,190}
\definecolor{negDrop}{RGB}{220,120,120}
\definecolor{accLine}{RGB}{31,119,180}
\definecolor{srLine}{RGB}{230,126,34}
\definecolor{simLine}{RGB}{44,160,44}
\definecolor{q4Line}{RGB}{31,119,180}
\definecolor{q32Line}{RGB}{214,39,40}
\definecolor{q235Line}{RGB}{44,160,44}
\definecolor{deepseekLine}{RGB}{148,103,189}
\definecolor{gptLine}{RGB}{90,90,90}
\definecolor{geminiLine}{RGB}{255,127,14}
\definecolor{purple}{RGB}{144,153,196}
\definecolor{paperTableBg}{HTML}{f5f5f5}
\definecolor{paperTableBorder}{HTML}{e5e5e5}
\definecolor{contactIcon}{HTML}{D93025}
\definecolor{projectIcon}{HTML}{1A73E8}
\definecolor{githubIcon}{HTML}{24292F}

\newcommand{\resourceLink}[5]{%
  \noindent\textcolor{#1}{#2}\hspace{0.35em}\textbf{#3:} \href{#4}{#5}\par
}

\newcommand{\resourcelist}{}
\newcommand{\paperResources}{%
  \def\resourcelist{%
    \vspace{0.35em}
    {\small\sffamily\setlength{\parskip}{0.15em}%
    \resourceLink{projectIcon}{\faGlobe}{Project Page}{https://thinking-inertia.github.io/}{thinking-inertia.github.io/}
    \resourceLink{githubIcon}{\faGithub}{GitHub}{https://github.com/thinking-inertia/code}{github.com/thinking-inertia/code}
    }%
  }%
}
\newtcolorbox{maintablebox}[1][]{
  enhanced,
  width=\linewidth,
  colback=paperTableBg,
  colframe=paperTableBorder,
  arc=3mm,
  boxrule=0.4pt,
  left=6pt,
  right=6pt,
  top=6pt,
  bottom=6pt,
  boxsep=0pt,
  #1
}

\newcommand{\accv}[2]{\cellcolor{accHeat!#1}#2}
\newcommand{\srv}[2]{\cellcolor{srHeat!#1}#2}
\newcommand{\simv}[2]{\cellcolor{simHeat!#1}#2}

\newcommand{\graytxt}[1]{\textcolor{gray}{#1}}

\providecommand{\cmark}{\ding{51}}
\providecommand{\xmark}{\ding{55}}

\providecommand{\accval}[2]{}
\providecommand{\etrval}[2]{}
\providecommand{\simval}[2]{}
\providecommand{\eirval}[2]{}

\renewcommand{\accval}[2]{%
  \begingroup
  \setlength{\fboxsep}{0.6pt}%
  \colorbox{accHeat!#1}{\strut #2}%
  \endgroup
}

\renewcommand{\etrval}[2]{%
  \begingroup
  \setlength{\fboxsep}{0.6pt}%
  \colorbox{srHeat!#1}{\strut #2}%
  \endgroup
}

\renewcommand{\simval}[2]{%
  \begingroup
  \setlength{\fboxsep}{0.6pt}%
  \colorbox{simHeat!#1}{\strut #2}%
  \endgroup
}

\renewcommand{\eirval}[2]{%
  \begingroup
  \setlength{\fboxsep}{0.6pt}%
  \colorbox{eirHeat!#1}{\strut #2}%
  \endgroup
}

%% file: section/intro.tex
\section{Introduction}
\label{sec:introduction}

\begin{figure*}[!h]
    \centering
    \includegraphics[width=\textwidth]{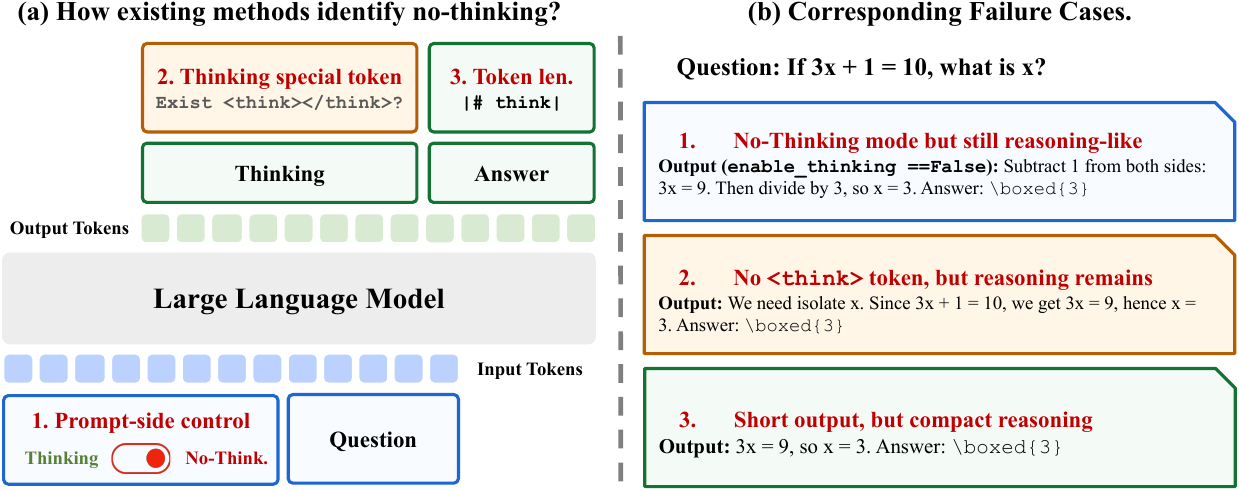}
\caption{
\textbf{Surface proxies fail to identify response-level \nothink{}.}
\textbf{(a)} Three common proxies: 1. prompt-side control, 2. special reasoning tokens (\eg~\texttt{<think>}), and 3. generation token length.
\textbf{(b)} On the same algebra question, disabling the thinking mode, omitting special thinking tokens, or producing a short output can each leave a visible inferential step before the final answer.
These cases motivate separately evaluating strict answer-only compliance, pre-answer relevance, and visible explicit inference.
}
\label{fig:teaser}
\end{figure*}

LLMs have benefited substantially from extended reasoning. From early chain-of-thought prompting~\citep{wei2022cot,kojima2022zeroshot} to more recent test-time scaling~\citep{snell2024scaling,brown2024largelanguagemonkeys}, allocating additional tokens to intermediate computation before returning an answer has significantly improved performance across a wide range of tasks. 
The same mechanism, however, raises inference cost in both latency and token consumption, motivating a line of work on shortening, budgeting, or compressing reasoning traces~\citep{aggarwal2025l1,kang2024c3ot,han2024tokenbudget,xu2025chainofdraft}.
Beyond efficiency, knowing when to answer directly without extended reasoning is also important for human-like adaptive reasoning: humans often respond from intuition and deliberate only when needed.
Together, these efficiency and adaptivity considerations motivate explicit control over reasoning behavior.
As illustrated in Figure~\ref{fig:teaser} (a), current frontier systems expose native thinking switches and reasoning-effort settings, while users can request direct answers through prompting~\citep{openai2025reasoningmodels,yang2025qwen3,DeepSeek-AI2024DeepSeekV3TR}.

Despite these explicit controls, as Figure~\ref{fig:teaser} (b) illustrates, a frontier model with its reasoning channel disabled can still emit question-conditioned content in the ordinary answer channel or produce a short response that nonetheless contains a compact, task-specific reasoning step.
Moreover, the community still \textit{lacks a clear definition of what a ``\nothink{}'' response should look like}.
In practice, \nothink{} is typically inferred from surface proxies: the prompt is placed in a no-think regime~\citep{leviathan2025prompt};
the response does not return a separate reasoning trace, or the \texttt{<think>} block is absent~\citep{yang2025qwen3,DeepSeek-AI2024DeepSeekV3TR}; or the output is simply short~\citep{aggarwal2025l1,kang2024c3ot,han2024tokenbudget,munkhbat2025concise,xu2025chainofdraft}.
These proxies conflate instruction control with the suppression of visible reasoning. 
For example, a generic lead-in such as ``{Let me think about this. The answer should be \texttt{<ans>}}'' carries no task-specific content, yet would be counted as reasoning by surface proxies.
Conversely, a single compact equation may expose an explicit inferential step even when the response is short or contains no special reasoning tokens.

Motivated by this conflation, we study \nothink{} as an observable response behavior and inspect the pre-answer text exposed before the final answer.\footnote{Here, \nothink{} is used as a response-level concept, distinct from latent reasoning processes within the LLM.}
In general question answering, pre-answer text may reflect reasoning, paraphrasing, fact restatement, or simple assertion.
We therefore define three progressively more specific criteria for characterizing such text.
\textbf{(a)} a \emph{discrete} view of whether the complete visible output reaches the strict answer-only boundary, measured by the \textbf{\ETRName{} (\ETR{})};
\textbf{(b)} a \emph{continuous} view of whether any remaining pre-answer text $T$ is relevant to the question $Q$, measured by instruction-aware Qwen3-Embedding-4B via \textbf{\QRelName{} (\QRel{})}; and
\textbf{(c)} a \emph{reasoning} view of whether $T$ contains an explicit inferential step that addresses $Q$, measured by a blinded GPT-5.5 judge through the \textbf{\EIRName{} (\EIR{})}.
The judge separates explicit inference from paraphrases and relevant non-inferential text, while task accuracy independently evaluates the final answer $A$. Thus, \QRel{} provides a relevance signal rather than final evidence of reasoning; conclusions about visible reasoning and thinking inertia rely on \EIR{} together with \ETR{}.
This view defines a testable behavioral framework for comparing interventions, benchmarks, model families, and model scales.

Based on these metrics, our experiments further examine \textit{whether \nothink{} can be controlled, regularized, and generalized.}
We first ask \textit{Q1.} whether enabling or disabling the thinking mode actually suppresses visible pre-answer content.
We then \textit{Q2.} compare several interventions\footnote{We use \emph{intervention} to denote a controlled change to the model's thinking configuration, the output constraints, or the question's answer interface.} to test whether stronger answer-only constraints can make responses more answer-only without damaging task accuracy. Finally, we evaluate \textit{Q3.} how this behavior changes across answer spaces, benchmark domains, and LLMs of different families and scales.
For M1--M5, $T$ is pre-answer text. Assistant-prefill M6 uses the first boxed answer and all returned readable text outside its initial declaration (Section~\ref{sec:no-think-performance}).
The detailed intervention spectrum is introduced in Section~\ref{sec:s3-spectrum}; at a high level, it separates native no-think behavior, reasoning re-elicitation, semantic suppression, and structural prefix forcing.
Our human validation supports all three metrics: a manual boundary audit corroborates \ETR{}, the five annotators' relevance judgments align closely with \QRel{}, and their explicit-inference labels achieve 98.8\% mean binary agreement with the primary GPT-5.5 judge used to compute \EIR{}.
The human annotations also reproduce the main answer-space and intervention trends, further supporting our central conclusions.
The \EIR{} results also remain robust when evaluated by an independent judge from a different model family.

Our results highlight four findings.
\textbf{(i)} Explicit no-think controls are not reliable \nothink{} controls: even a frontier LLM such as DeepSeek-V4-Flash (284B) retains visible explicit inference in 99.9\% of open-ended responses under its disabled-thinking setting.
\textbf{(ii)} LLMs exhibit \emph{``thinking inertia''}: visible inference is organized more by the task's answer space than by the switch itself. Under both native and strict no-thinking settings, it remains lowest on Boolean tasks, intermediate on multiple-choice tasks, and highest on open-ended tasks.
\textbf{(iii)} On open-ended tasks, stronger answer-only instructions reveal what we call a \nothink{}--performance trade-off: stronger suppression improves answer-only compliance and reduces visible inference, but can also reduce task accuracy. Specifically, across the five LLMs with fully disabled thinking, strict answer-only prompting raises mean \ETR{} from near zero to roughly 40\% while reducing mean accuracy by about 15 percentage points.
\textbf{(iv)} Answer space matters: for the same question, fact-like and candidate-visible formats make answer-only responses easier to produce than open-generation formats, showing that the pattern is not explained by benchmark difficulty or answer length alone.

In summary, our work makes two contributions.
First, we define \nothink{} as a measurable response-level behavior and provide a framework for evaluating it.
Second, through comprehensive experiments across task families and diverse LLMs, we identify systematic behaviors including thinking inertia, the \nothink{}--performance trade-off, and answer-space dependence.
Together, these contributions support the broader goal of developing LLMs that, like humans, can adaptively determine when to reason and when to answer directly.

%% file: section/method.tex
\section{Measuring No-Thinking}
\label{sec:measurement-and-spectrum}

\subsection{Observations}
\label{sec:s3-motivation}

No-thinking in LLMs does not have a concrete definition.
In practice, existing work often identifies it through three surface proxies, as summarized on the left side of Figure~\ref{fig:teaser}: 
(\textbf{i}) \emph{prompt-side control}, where the model is run in a no-think setting or is not prompted to think step by step \cite{leviathan2025prompt}; 
(\textbf{ii}) \emph{visible-format control}, where no-thinking is inferred from the absence of an explicit \texttt{<think>} block or a separate reasoning trace \cite{yang2025qwen3,DeepSeek-AI2024DeepSeekV3TR}; and
(\textbf{iii}) \emph{length-based control}, where shorter generations produced by length control or chain-of-thought compression are treated as reduced reasoning \cite{aggarwal2025l1,kang2024c3ot}. 
These proxies are convenient and easy to monitor, but they characterize the response by its surface prompting condition, visible format, or length, rather than by the actual content of the LLM response (such as the question-conditioned information it actually exposes).

Each of these formulations, however, has a simple failure case, illustrated on the right side of Figure~\ref{fig:teaser}.
(\textbf{i}) A no-think instruction can still yield explanatory, question-dependent text before the final answer; disabling a thinking mode therefore does not by itself remove visible reasoning.
(\textbf{ii}) Removing explicit \texttt{<think>} tags only removes a particular format; reasoning may still appear as ordinary language in the answer channel.
(\textbf{iii}) Making an output short does not make it no-thinking, since a compact response can still encode the essential intermediate step needed to answer the question.
Thus, the problem is not that these proxies are useless, but that none of them defines the behavioral object we want to study.
We therefore operationalize \nothink{} at the response level and progressively distinguish strict answer-only compliance, question relevance, and visible explicit inference.
This observable definition does not make claims about latent computation inside the model, because extending the analysis to latent reasoning would substantially broaden the scope and make the boundary of no-thinking difficult to define clearly.

\begin{figure}[t]
    \centering
    \includegraphics[width=\linewidth]{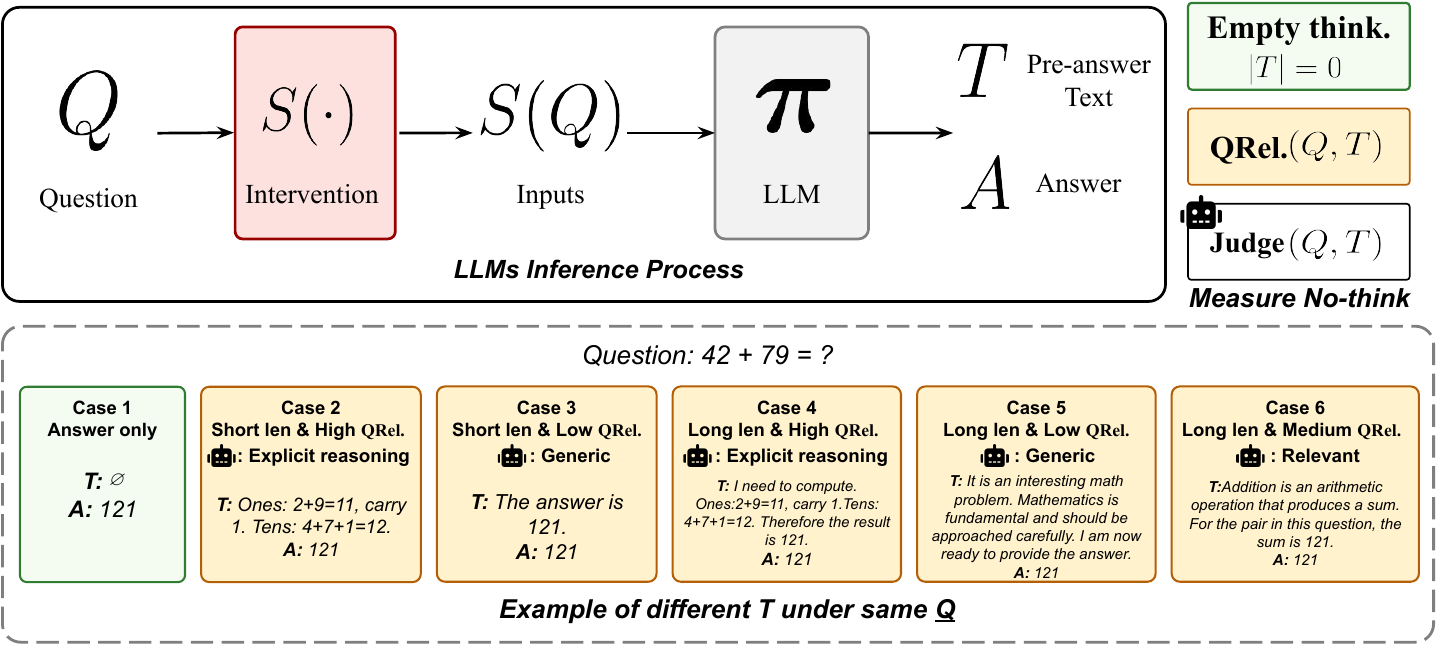}
\caption{
\textbf{Response-centric metrics for measuring no-thinking.}
\textbf{Top:} the LLM inference process. A question $Q$ is wrapped by an intervention $S(\cdot)$ to form the model input $S(Q)$, which the model $\pi$ maps to a response decomposed into pre-answer text $T$ and final answer $A$. We evaluate the visible response through three levels: strict answer-only compliance (\ETR{}), question--pre-answer relevance (\QRel{}), and LLM judge explicit inference (\EIR{}); accuracy independently evaluates $A$.
\textbf{Bottom:} six representative response cases to the same question \emph{``$42 + 79 = ?$''} vary in length, relevance, and inferential content while sharing the same final answer. Case~1 is answer-only (empty $T$), whereas Cases~2--6 use the non-empty-$T$ labels from the mutually exclusive rubric in Table~\ref{tab:reasoning-rubric}: Cases~2 and~4 are explicit reasoning, Cases~3 and~5 are generic/off-topic, and Case~6 is relevant non-inferential.
Table~\ref{tab:reasoning-rubric} additionally defines paraphrase-only and unclear; representative real outputs are provided in Appendix~\ref{app:five-response-regimes}.
}
\label{fig:framework-measurement}
\end{figure}

\subsection{Problem Formulations}
\label{sec:s3-framework}

The observations above suggest that prompt-side heuristics alone are insufficient for studying no-thinking in a controlled way.
We therefore ground our analysis in the response-centric framework shown in Figure~\ref{fig:framework-measurement}.

Formally, let $\pi$ denote a language model and $Q$ a task sample. The query is wrapped by an intervention $S(\cdot)$ (\eg~a system prompt such as ``please answer \texttt{<Question>} directly'') before being passed to the model, which, for the ordinary M1--M5 response order, produces an observable response
\begin{equation}
\pi\bigl(S(Q)\bigr) = \left(T, A\right)
\label{eq:response-decomp}
\end{equation}
where $A$ is the last parseable final answer and $T$ comprises all readable visible text preceding it. We use $(T,A)$ only as a decomposition of the observable response, without making claims about task-relevant computation in unobserved model states. The strict compliance check additionally scans the complete output, so substantive text after $A$ is not treated as answer-only. Appendix~\ref{app:measurement-audits} audits question restatement and answer-first edge cases. Assistant prefilling may alter this order, so we evaluate M6 using its first boxed answer and all residual text (Section~\ref{sec:no-think-performance}).

\noindent\textbf{Visible pre-answer text $T$.}
$T$ contains all readable visible content before the final answer $A$, whether it appears in a special reasoning block, the ordinary answer channel, or a short lead-in. It may include an explicit derivation, a compact reasoning step, a paraphrase, a relevant assertion, or a generic template such as ``The answer is''. These cases motivate evaluating the content of $T$ rather than equating every non-empty $T$ with reasoning.

\noindent\textbf{Question type and answer space $(Q, A)$.}
We write each task sample as $Q = (q, c, \mathcal{A})$, where $q$ is the question, $c$ is optional context, and $\mathcal{A}$ is the admissible answer space, with $A \in \mathcal{A}$. For example, $\mathcal{A} = \{\texttt{yes}, \texttt{no}\}$ for Boolean verification, $\mathcal{A} = \{\texttt{A}, \texttt{B}, \texttt{C}, \texttt{D}\}$ for multiple-choice, and $\mathcal{A}$ permits free-form generation for open-ended tasks. This lets us test whether no-thinking behavior changes under different answer interfaces.

\noindent\textbf{Intervention $S(\cdot)$.}
$S(\cdot)$ denotes an external intervention applied to $Q$ before generation, producing the model input $S(Q)$. We consider two broad classes: 
(\textbf{i}) \emph{question-irrelevant interventions}, which change the instruction, output constraint, or explicit thinking configuration while keeping $Q$ fixed (\eg~prepending ``Answer directly without thinking'');
and (\textbf{ii}) \emph{question-relevant interventions}, which modify the task instance itself, such as its surface form or answer space $\mathcal{A}$ (\eg~converting an open-ended question into a multiple-choice one). 

Together, this unified framework allows us to study \nothink{} through strict compliance, relevance, and explicit inference rather than through surface proxies alone.

\subsection{Three-Level Response Evaluation}
\label{sec:s3-measurement}
Under our formulation, $T$ serves as the primary observable target, while task accuracy separately evaluates the final answer $A$.
We therefore organize the $T$ measurement into three progressively more specific levels (as depicted in Figure~\ref{fig:framework-measurement}): strict answer-only compliance, pre-answer relevance, and visible explicit inference. At the response level, Levels~1 and~3 yield binary scores, whereas Level~2 yields a continuous score; their benchmark-level aggregates are defined below.

\noindent\textbf{Level 1: \ETRName{}.}
We first ask whether each output is strictly answer-only. For output $i$, let $e_i$ denote the response-level empty-thinking indicator. For a benchmark with $N$ outputs under a fixed model and intervention, we define the response-level score and the \textbf{\ETRName{} (\ETR{})} as
\begin{equation}
e_i=\mathbf{1}\!\left\{|T_i|=0\right\},
\qquad
\mathrm{\ETR}=100\%\times\frac{1}{N}\sum_{i=1}^{N}e_i.
\label{eq:etr}
\end{equation}
Here, $e_i$ is the response-level score, whereas \ETR{} is its benchmark-level rate. Operationally, $e_i=1$ only when $A_i$ is parseable and the complete visible output contains no substantive text outside its permitted answer wrapper; otherwise, $e_i=0$. Thus, empty $T_i$ does not override a malformed final-answer format, and substantive text following $A_i$ also violates strict compliance. \ETR{} gives the cleanest operational limit, but cannot distinguish a generic template, a paraphrase, and a question-specific inference once $T_i$ is non-empty.

\noindent\textbf{Level 2: \QRelName{}.}
We next ask how strongly the visible pre-answer text $T$ is semantically related to $Q$. Let $\phi$ denote an embedding encoder. We instantiate $\phi$ with the instruction-aware Qwen3-Embedding-4B~\citep{zhang2025qwen3embedding}, because its query instruction lets us specify the intended relation between $Q$ and $T$: $Q$ is encoded as the query under the instruction ``Given a question, retrieve pre-answer text that contains reasoning relevant to answering the question,'' while $T$ is encoded as the candidate passage. Here, $\phi(Q)$ denotes the resulting instruction-conditioned query embedding and $\phi(T)$ the candidate-passage embedding produced by the same encoder. After $\ell_2$-normalizing each embedding, i.e., dividing each vector by its Euclidean norm, we define the response-level \textbf{\QRelName{} (\QRel{})} score as
\begin{equation}
\mathrm{\QRel}(Q_i,T_i)=
\begin{cases}
0, & |T_i|=0,\\
\cos\!\bigl(\phi(Q_i),\phi(T_i)\bigr), & |T_i|>0.
\end{cases}
\label{eq:fpara}
\end{equation}
The first branch assigns empty $T_i$ a score of zero, while the second directly uses the cosine similarity for non-empty $T_i$. For each model--benchmark--intervention setting, the reported \QRel{} is the arithmetic mean of these response-level scores over all $N$ outputs, including the zeros assigned to empty $T_i$. The instruction-aware objective makes relevance to answering $Q_i$, rather than surface overlap alone, the target relation.

However, \QRel{} measures relevance rather than inference: a paraphrase or a relevant answer assertion can receive a high score without containing an inferential step. Appendix~\ref{app:measurement-audits} audits whether near-verbatim or semantic restatement drives the aggregate results, while Level~3 keeps such text outside the explicit-inference category.

\noindent\textbf{Level 3: \EIRName{}.}
Level 1 checks a structural rule and Level 2 measures functional relevance; neither directly separates reasoning from relevant non-inferential text. We therefore use a blinded LLM judge to classify each non-empty $T_i$. The judge receives only $(Q_i,T_i)$; model identity, intervention mode, final answer $A_i$, correctness, and \QRel{} are hidden. It returns one of the five mutually exclusive labels in Table~\ref{tab:reasoning-rubric}, together with an exact supporting span from $T_i$. The span records the evidence for the selected label and makes the decision auditable, but does not enter the metric. Table~\ref{tab:reasoning-rubric} presents the judge instruction and rubric together.

\begin{table}[H]
\begin{maintablebox}[left=4pt,right=4pt,top=4pt,bottom=4pt]
\centering
\small
\setlength{\tabcolsep}{4pt}
\renewcommand{\arraystretch}{1.10}
\captionsetup{skip=2pt}
\caption{\textbf{Blinded LLM-judge instruction and rubric for non-empty pre-answer text $T$.} Empty $T$ is handled separately and counts as no visible explicit inference.}
\label{tab:reasoning-rubric}
\begin{tcolorbox}[
  enhanced,
  colback=blue!3,
  colframe=black!75,
  boxrule=0.45pt,
  arc=1.5mm,
  attach boxed title to top left={xshift=5pt,yshift=-2mm},
  boxed title style={colback=black!85,colframe=black!85,arc=1mm,left=3pt,right=3pt,top=1pt,bottom=1pt},
  title={Prompt for the LLM Judge},
  fonttitle=\small\bfseries,
  coltitle=white,
  left=5pt,
  right=5pt,
  top=6pt,
  bottom=3pt,
  before skip=2pt,
  after skip=6pt
]
\raggedright
Judge whether the \texttt{PRE-ANSWER TEXT} contains reasoning that addresses the \texttt{QUESTION}, select the corresponding rubric label, and record an exact supporting span.
\end{tcolorbox}
\begin{tabularx}{\linewidth}{@{}p{0.04\linewidth}p{0.22\linewidth}X@{}}
\toprule
\textbf{Index} & \textbf{Label} & \textbf{Criterion} \\
\midrule
1 & Generic/off-topic & Generic lead-in, formatting text, or off-topic content with no question-specific inference. \\
2 & Paraphrase-only & Restates $Q$ or supplied facts without connecting them through an inference step. \\
3 & Relevant non-inferential & Gives a relevant fact, definition, description, or answer assertion without an inference step. \\
4 & \textbf{Explicit reasoning} & Contains deduction, computation, comparison, elimination, rule application, evidence-to-conclusion, or causality addressing $Q$. \\
5 & Unclear & Cannot be classified reliably from the visible text. \\
\bottomrule
\end{tabularx}
\end{maintablebox}
\end{table}
\FloatBarrier

A relevant restatement or answer assertion alone is therefore not reasoning, whereas one concrete inferential step can qualify even when it is short, shallow, or ultimately incorrect.
For the metric, let $\operatorname{Judge}(Q_i,T_i)$ denote the binary score induced by the five-way label returned when the blinded judge receives $(Q_i,T_i)$: it is 1 only when the judge selects \emph{Explicit reasoning}, and 0 for the other four labels.
Empty $T_i$ is not sent to the judge and is assigned $\operatorname{Judge}(Q_i,T_i)=0$ by defacult. For a benchmark with $N$ outputs under a fixed model and intervention, we define the \textbf{\EIRName{} (\EIR{})} as
\begin{equation}
\mathrm{\EIR}=100\%\times\frac{1}{N}\sum_{i=1}^{N}\operatorname{Judge}(Q_i,T_i).
\label{eq:eir}
\end{equation}
Thus, $\operatorname{Judge}(Q_i,T_i)$ is the response-level binary score,
whereas \EIR{} is its benchmark-level statistics. 

\paragraph{Human study.}
We complement the original pre-answer evaluation with a blinded human study that considers two aspects: \textit{first}, whether the three metrics align with human judgments; and \textit{second}, whether the main conclusions drawn from them persist under direct human annotation.
\textbf{(i) Metric validation.} A manual boundary audit checks the strict visible boundary underlying \ETR{}, while five annotators independently provide graded relevance judgments for comparison with \QRel{} and explicit-inference labels for comparison with the Level~3 judge. Additionally, we use a second LLM judge from a different model family to check that the Level~3 results are not specific to the primary judge. Additionally, we use a second LLM judge from a different model family to check that the Level~3 results are not specific to the primary judge.
\textbf{(ii) Conclusion validation.} Building on this agreement, we test whether the main answer-space and intervention trends persist under direct human annotation. The detailed protocol and results are reported in Section~\ref{sec:human-validation} and Appendix~\ref{app:human-validation}.

\subsection{Thinking Behavior Control}
\label{sec:s3-spectrum}
To systematically cover the possible setups,
we design a six-mode spectrum (Table~\ref{tab:six-modes}) that organizes modes along a single axis of \emph{no-think intensity}, ranging from $(-)$ for thinking-encouraging modes and $(0)$ for the native no-thinking mode to $(+\!+\!+)$ for the strongest suppression.
All modes share the same task stem and final-answer request; only the thinking switch and the question-irrelevant intervention change.

\input{table/baseline_new}

\textbf{Thinking variants} (\textsc{Think-On}, \textsc{CoT}).
\textsc{Think-On} (M1) keeps the official thinking interface enabled, while \textsc{CoT} (M3) disables it but adds the instruction \emph{``Think step by step''} to re-elicit reasoning in the ordinary answer channel. Both elicit thinking rather than suppress it, and serve as opposite-direction references for the no-think space (gray rows in Table~\ref{tab:six-modes}).

\textbf{No-thinking variants} (\textsc{Think-Off}, \textsc{Soft}, \textsc{Strict}, \textsc{Prefix}).
\textsc{Think-Off} (M2) disables \texttt{enable\_thinking} without any additional output-side control, serving as the natural baseline against which all suppression interventions are measured.
\textsc{Soft} (M4) and \textsc{Strict} (M5) keep \texttt{enable\_thinking=False} and add increasingly strong natural-language constraints: \textsc{Soft} allows a short lead-in but forbids explanation, while \textsc{Strict} asks for the answer directly without any reasoning. \textsc{Prefix} (M6) is qualitatively different. It supplies the answer template ``The answer is \texttt{\textbackslash boxed\{}'' as the unfinished assistant response and generates its continuation.

%% file: table/baseline_new.tex
\begin{table*}[t]
\begin{maintablebox}
\centering
\small
\setlength{\tabcolsep}{2pt}
\renewcommand{\arraystretch}{1.14}
\caption{
\textbf{Six-mode question-irrelevant prompt intervention spectrum.}
Modes shown in \graytxt{gray (\textsc{Think-On}, \textsc{CoT})} still elicit thinking behavior and serve as thinking-side references; the remaining modes (\textsc{Think-Off}, \textsc{Soft}, \textsc{Strict}, \textsc{Prefix}) form the no-think intervention space, ranging from the native no-thinking baseline to increasingly strong suppression.
}
\label{tab:six-modes}
\begin{tabularx}{\linewidth}{@{}c l l c X@{}}
\toprule
$\#$
& \textbf{Mode} 
& \textbf{Purpose} $($no-think intensity$)$
& \textbf{\texttt{enable\_thinking}} 
& \textbf{Question-Irrelevant Intervention $S(\cdot)$} \\
\midrule
\graytxt{M1} 
& \graytxt{\textsc{Think-On}} 
& \graytxt{Native \texttt{thinking-on} mode $(-)$} 
& \graytxt{\cmark} 
& \graytxt{None.} \\
M2 
& \textsc{Think-Off}
& Native \texttt{no-thinking} mode  $(0)$
& \xmark 
& None. \\
\midrule
\graytxt{M3} 
& \graytxt{\textsc{CoT}}
& \graytxt{Explicit reasoning $(-)$}  
& \graytxt{\xmark}
& \graytxt{``Think step by step.''} \\
M4 
& \textsc{Soft} 
& Soft regularization $(+)$
& \xmark 
& {``A very short lead-in is allowed, but do not explain.''} \\
M5 
& \textsc{Strict}
& Strict regularization $(++)$
& \xmark 
& {``Give the final answer directly without any reasoning.''} \\
M6 
& \textsc{Prefix}
& Prefix forcing $(+++)$
& \xmark 
& {``The answer is \texttt{\textbackslash boxed\{}''} \\
\bottomrule
\end{tabularx}
\end{maintablebox}
\end{table*}

%% file: section/exps.tex
\section{Experiments}
\label{sec:experiments}

We organize our experiments around three questions.
Starting from the model's native thinking controls, we ask \textbf{Q1.} whether enabling or disabling the thinking mode actually suppresses visible pre-answer content.
We then focus on the native no-thinking mode and ask \textbf{Q2.} how prompt interventions of varying strength affect no-thinking behavior and its relationship with task performance.
Building on these analyses, we conduct finer-grained controlled comparisons and ask \textbf{Q3.} how no-thinking behavior varies across benchmark domains and question types, and how much of this variation can be attributed to answer space.

\Needspace{10\baselineskip}
\subsection{Settings}
\label{sec:exp-setup}

\begin{wraptable}[9]{r}{0.4\textwidth}
\vspace{-0.5em}
\begin{maintablebox}[left=4pt,right=4pt,top=4pt,bottom=2pt]
\centering
\scriptsize
\setlength{\tabcolsep}{1.8pt}
\renewcommand{\arraystretch}{1.02}
\caption{LLM baselines for evaluation.}
\label{tab:model-coverage}
\vspace{-0.6em}
\begin{tabular}{@{}lcc@{}}
\toprule
\textbf{Model}  & \textbf{Size} & \textbf{Thinking mode} \\
\midrule
\texttt{Qwen3-4B} & 4B  & On | Off \\
\texttt{Qwen3-32B} & 32B  & On | Off \\
\texttt{Qwen3-235B-A22B}  & 235B  & On | Off \\
\texttt{DeepSeek-V4-Flash}  & 284B  & High | Off \\
\texttt{GPT-5.4}  & N/A  & High | Off \\
\texttt{Gemini-3-Flash} & N/A & High | Minimal \\
\bottomrule
\end{tabular}
\label{tab:baseline}
\end{maintablebox}
\end{wraptable}

\paragraph{LLM baselines.}
We select LLMs that expose explicit thinking-mode controls, so that \nothink~behavior can be tested by intervention rather than inferred from output length or format.
Table~\ref{tab:baseline} spans the Qwen3 family (4B, 32B, and 235B), DeepSeek-V4-Flash, and proprietary frontier models from OpenAI and Google, covering both model families and scales.
The three Qwen3 sizes support a within-family scaling analysis, while the local 4B and 32B checkpoints provide transparent paired native on/off controls for detailed ablations. All six models are evaluated under M1--M5 from Section~\ref{sec:s3-spectrum}; assistant-prefill M6 covers the three Qwen sizes and DeepSeek.

\paragraph{Benchmarks.}
To evaluate no-thinking across answer spaces, we use three answer-space families, each with an easier and a harder benchmark.
\textit{(i) Boolean verification.} Each sample requires a yes/no judgment. We use BoolQ~\citep{clark2019boolq} and StrategyQA~\citep{geva2021strategyqa}.
\textit{(ii) Multiple-choice answering.} Each sample provides a fixed set of candidate options, and the model returns one option label. We use MMLU~\citep{hendrycks2020measuring} and MMLU-Pro~\citep{wang2024mmlupro}.
\textit{(iii) Open-ended answering.} Each sample requires the model to generate the answer without a provided candidate set. We use GSM8K~\citep{cobbe2021training} and MATH~\citep{hendrycks2021math}.
These benchmarks cover answer spaces ranging from binary judgments through fixed-choice selection to open-ended generation, while using the same response-level evaluation metrics.

\paragraph{Metrics.}
For each model--benchmark--intervention setting, we report \textbf{Task Accuracy (Acc.)} and the three no-thinking metrics defined in Section~\ref{sec:s3-measurement}: the \textbf{\ETRName{} (\ETR{})}, the \textbf{\QRelName{} (\QRel{})}, and the \textbf{\EIRName{} (\EIR{})}.
Metrics include outputs with both empty and non-empty $T$. Empty $T$ contributes zero to \QRel{} and \EIR{}. M1--M5 use the full output set as the denominator. M6 follows the residual-text rule in Section~\ref{sec:no-think-performance}.

\textbf{For \QRel{}}, we use the instruction-aware \texttt{Qwen3-Embedding-4B}~\citep{zhang2025qwen3embedding} with last-token pooling and a maximum sequence length of 32,768, following the query--passage setup in Section~\ref{sec:s3-measurement}.
\textbf{For \EIR{}}, we use \texttt{GPT-5.5} at temperature 0 as the primary blinded judge, following the input and blinding protocol in Section~\ref{sec:s3-measurement}.
Appendix~\ref{app:three-level-validation} tests Level 2 with alternative widely used encoders and replicates Level 3 with Claude Opus 4.8 across all six models, six benchmarks, and six intervention modes.

\paragraph{Human study.}
Our unified blinded human study samples 600 outputs, 100 from each benchmark, while preserving each benchmark's full-corpus empty/non-empty-$T$ proportion; the empty-$T$ boundary is manually audited for all outputs.
Five annotators independently review all 402 non-empty pre-answer texts and assign both a five-point question-relevance score and one Level 3 reasoning label; the 198 empty cases receive relevance score 1 and no visible inference by definition.
The annotation view exposes only $(Q,T)$; model identity, intervention mode, final answer, correctness, and automated evaluator scores are hidden.
Section~\ref{sec:human-validation} reports the results, while Appendix~\ref{app:human-validation} provides the full protocol, rubrics, calculations, and per-annotator statistics.

\begin{figure}[!t]
    \centering
    \includegraphics[width=\linewidth]{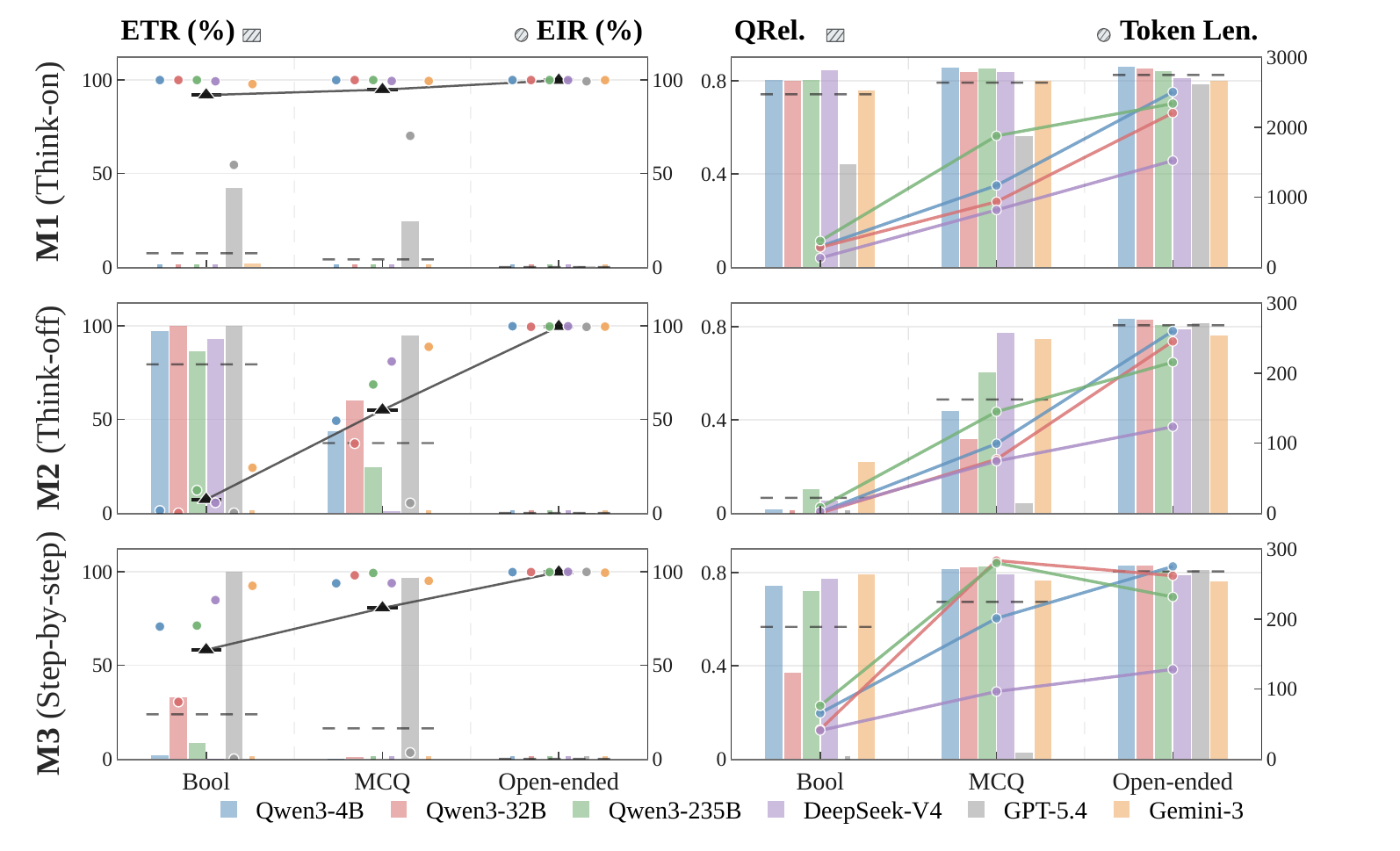}
\caption{
\textbf{Think-on (M1), Think-off (M2), and Step-by-step (M3) across six models and three answer-space families.}
Rows show modes; the columns pair ETR bars with EIR points (left) and QRel. bars with visible pre-answer length for the four models with complete length traces (right).
Gray dashed segments mark six-model means for ETR and QRel.; black triangles connected by dark lines mark the mean EIR; colored marks follow the model legend.
}
\label{fig:thinking-inertia-bars}
\end{figure}

\subsection{Native Think-on vs.\ Think-off}
\label{sec:thinking-inertia}

Figure~\ref{fig:thinking-inertia-bars} compares M1 (\textsc{Think-on}), M2 (\textsc{Think-off}), and M3 (\textsc{Step-by-step}) across six models and three answer-space families (Bool, MCQ, and Open-ended).
The left panels pair ETR with EIR, while the right panels pair QRel. with visible pre-answer length $|T|$.

\paragraph{Native Think-off becomes less effective as the answer space opens.}
We first test whether the native thinking-mode control reliably suppresses visible pre-answer content.
M1 (\textsc{Think-on}) serves as the native thinking-on reference: for most model--task pairs, ETR is near zero, while QRel. and EIR are high.
Under M2 (\textsc{Think-off}), most models reach roughly 80--100\% ETR on Bool, with sharply reduced QRel. and EIR.
MCQ forms an intermediate but heterogeneous regime: ETR ranges from near zero to 95\% across models, while the six-model mean QRel. and EIR values fall between Bool and Open-ended.
On Open-ended tasks, ETR collapses to nearly zero everywhere, while QRel. remains close to its M1 level and EIR remains high.
Across the three metrics, responses become less likely to remain answer-only and more likely to retain question-related, explicitly inferential pre-answer text as the answer space opens.
This observable progression is consistent with greater computational demands: constructing an answer in an open space may require more task-relevant intermediate computation than selecting from narrow or candidate-visible alternatives.

\begin{findingbox}
\textbf{We use ``thinking inertia'' to describe how visible explicit inference persists under no-thinking controls and becomes more common as the answer space opens.}
The native no-thinking setting therefore does not guarantee answer-only responses: ETR is high mainly on Bool, while the six-model mean EIR is lowest on Bool, intermediate on MCQ, and highest on Open-ended tasks.
\end{findingbox}

\paragraph{Step-by-step prompts can recover visible reasoning.}
M3 (\textsc{Step-by-step}) provides the reverse intervention: it asks whether an ordinary step-by-step prompt can re-elicit visible reasoning while the native thinking switch remains off.
Under M3, QRel. and EIR often move back toward their M1 levels, including on Bool for many models; the four available length traces show the same recovery. This indicates that an ordinary step-by-step instruction can re-elicit visible explicit inference even while the native thinking switch remains off.
Nevertheless, the six-model mean EIR preserves the same Bool $<$ MCQ $<$ Open-ended staircase under M3, showing that the answer-space dependence characteristic of \emph{thinking inertia} persists under explicit step-by-step prompting.
GPT-5.4 is the main exception to this recovery pattern, maintaining high answer-only compliance and correspondingly low EIR on both Bool and MCQ.
Thus, the reverse intervention changes the overall level of visible inference but not its answer-space ordering.

\enlargethispage{\baselineskip}
\paragraph{QRel. separates relevance from verbosity.}
In the right panels, pre-answer length under M1 (\textsc{Think-on}) grows sharply from Bool to Open-ended tasks for all four models with complete length traces.
Yet QRel. changes much less across these answer spaces, indicating that it is not merely a proxy for length. QRel. and EIR remain complementary: question-related $T$ may be a paraphrase or relevant assertion, whereas EIR identifies whether it contains an explicit inferential step.
Our M1--M5 audit flags 176 of 477,240 outputs (0.037\%) as near-verbatim repetitions. Setting all their \QRel{} values to zero changes the mean over this corpus by at most 0.00037.
An audit of the stored Level~3 labels identifies 477 M1--M5 outputs (0.100\%) as paraphrase-only; these do not enter the \EIR{} numerator.
Appendix~\ref{app:measurement-audits}, with Tables~\ref{tab:measurement-audit-app} and~\ref{tab:judge-paraphrase-audit-app}, provides the definitions and breakdowns.

\subsection{Performance across No-Think Variants}
\label{sec:no-think-performance}

We next ask how no-think interventions affect response behavior and task performance.

\begin{figure}[!t]
    \centering
    \includegraphics[width=\linewidth]{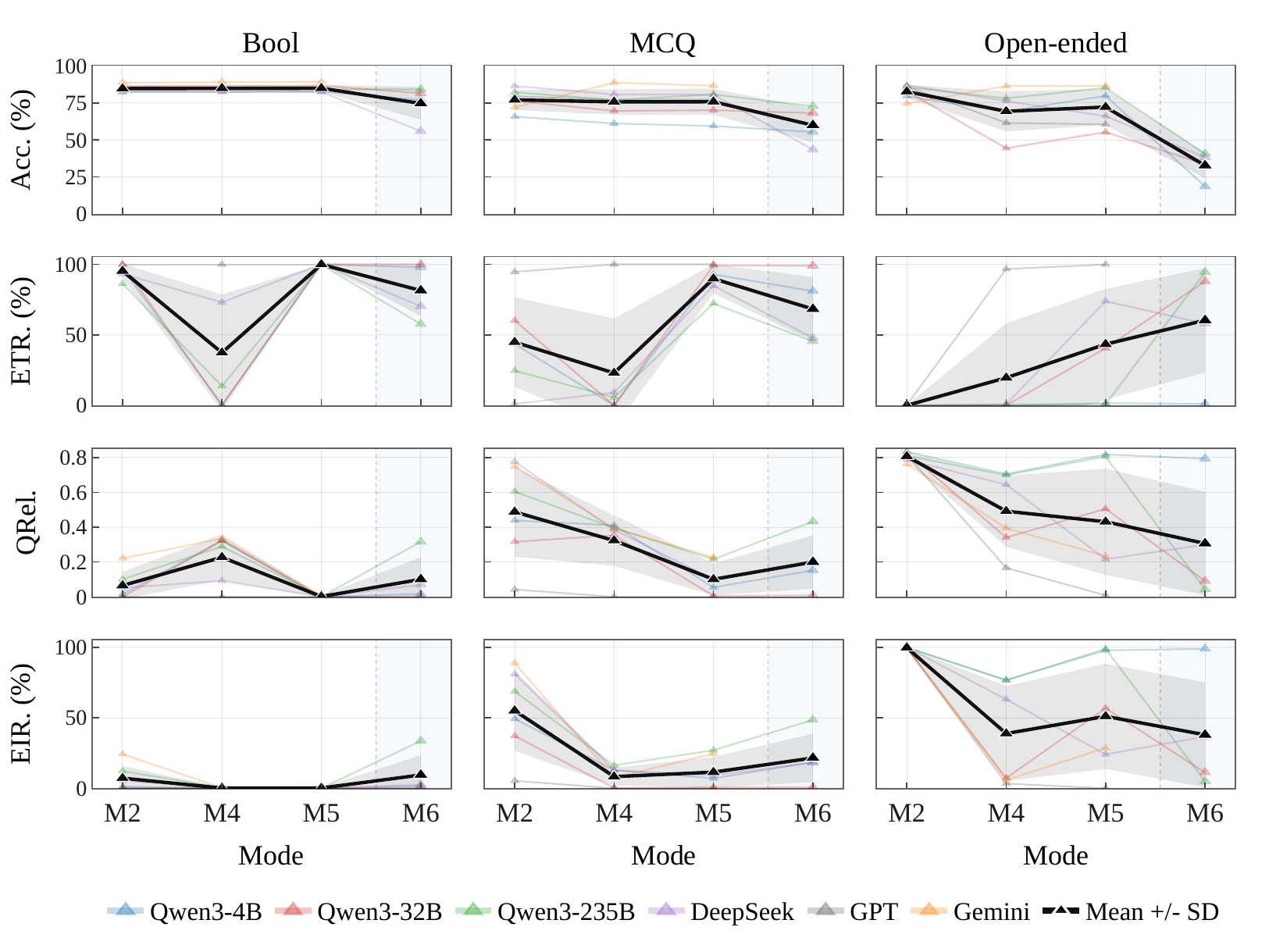}
    \caption{
    \textbf{Natural-language suppression and assistant prefilling across answer spaces.}
    Columns show Bool, MCQ, and Open-ended tasks; rows show Acc., \ETR{}, \QRel{}, and \EIR{}.
    M2/M4/M5 use pre-answer text; M6 uses residual text.
    Blue shading marks M6's distinct assistant-prefill control.
    Colored marks show models; black marks and gray bands give their mean and one standard deviation.
    Appendix~\ref{app:m6-prefill-protocol} gives evaluation and aggregation details.
    }
\label{fig:frontier-family-control}
\end{figure}

\paragraph{Open-ended tasks expose a trade-off between answer-only compliance and accuracy.}
Across M2, M4, and M5 in Figure~\ref{fig:frontier-family-control}, Bool accuracy remains nearly flat at around 85\%, MCQ accuracy varies modestly, and Open-ended accuracy varies most across modes.
On Open-ended tasks, moving from M2 to M5 raises mean \ETR{} from near zero to roughly 40\% and reduces mean accuracy by about 15 percentage points across models with a disabled-thinking condition.
Over the complete evaluation corpus for these two modes, \EIR{} likewise decreases from 99.69\% to 62.00\% (Appendix Table~\ref{tab:cross-judge-eir-app}).
Bool and MCQ show smaller accuracy changes, even as \EIR{} falls from 5.15\% to 0.12\% on Bool and from 45.57\% to 5.39\% on MCQ.
Thus, stronger natural-language suppression reduces visible explicit inference in every answer space, but Open-ended tasks retain substantial inference and show the clearest accuracy cost.

\paragraph{Evaluating assistant-prefilled responses.}
M6 starts the assistant response with \texttt{The answer is \textbackslash boxed\{} and lets the model continue from that unfinished prefix.
We find that M6 responses do not consistently follow the pre-answer $(T,A)$ structure observed under M1--M5. Literal answer-first responses with a text suffix are rare among boxed M1--M5 outputs (Appendix Table~\ref{tab:measurement-audit-app}).
In M6, alongside answer-only responses, we observe $(A,T)$ and $(A,T,A')$, as well as inference within the initial answer box, schematically \texttt{\textbackslash boxed\{T,A\}}. The initial box may also be empty or unclosed (Table~\ref{tab:m6-format-audit-app}).
To evaluate these responses, \textbf{Acc.} scores the answer extracted from the \emph{first prefilled box}, using the same benchmark comparator as M1--M5; a later correction does not replace that answer.
For the behavioral metrics, we remove the first complete \texttt{The answer is \textbackslash boxed\{xxx\}} declaration, including its entire box, and define $T$ as \emph{all remaining returned readable text}: subsequent explanations, repeated or revised answers, and any separately returned reasoning text.
A response counts as answer-only for \textbf{\ETR{}} when it contains a parseable first boxed answer and no substantive residual text.
\textbf{\QRel{}} and \textbf{\EIR{}} use this residual $T$ with the same encoder, judge, and rubric as M1--M5; empty $T$ receives zero for both metrics.
These residual-text scores exclude any inference inside the removed box.
Appendix~\ref{app:m6-prefill-protocol} gives further protocol details, and Appendix~\ref{app:historical-user-prefix} reports the separate user-message-prefix experiment.

\paragraph{Accuracy is lower under M6 prefilling, with losses increasing from Bool to MCQ to Open-ended tasks.}
For the three Qwen sizes and DeepSeek, mean accuracy falls from 83.85\% under M2 to 74.66\% under M6 on Bool, from 77.68\% to 59.93\% on MCQ, and from 83.78\% to 32.76\% on Open-ended tasks: decreases of 9.19, 17.75, and 51.01 percentage points, respectively.
For each of the four models, the decrease from M2 is smallest on Bool and largest on Open-ended tasks.
Compared with M5, mean accuracy decreases by 9.27, 12.75, and 38.84 points, respectively, although individual models do not all follow this ordering.
The three behavioral metrics show broadly similar changes from M5 across models on Bool and MCQ, but diverge on Open-ended tasks.
On Bool and MCQ, M6 generally lowers answer-only compliance and increases residual relevance and inference, while Qwen3-32B remains almost entirely answer-only.
On Open-ended tasks, answer-only compliance rises substantially from M5 for Qwen3-32B and Qwen3-235B, falls for DeepSeek, and remains near zero for Qwen3-4B.
Qwen3-4B combines persistent residual inference with low first-box accuracy, whereas Qwen3-235B is usually answer-only but still suffers a substantial accuracy loss.

\begin{findingbox}
\textbf{On open-ended tasks, no-thinking comes at a cost.}
This pattern is consistent with greater computational demands in Open-ended tasks: compared with native \textsc{Think-off}, strict answer-only prompting lowers \EIR{} but leaves substantial visible explicit inference, while higher answer-only compliance accompanies lower accuracy.
For M6, first-answer accuracy losses grow from Bool through MCQ to Open-ended tasks, where answer-only compliance and residual inference vary sharply across models.
\end{findingbox}

\subsection{Performance across disciplines}

We next address the domain component of \textit{Q3} by holding the answer interface fixed and comparing no-thinking behavior across subject areas.

\paragraph{Higher accuracy does not imply easier no-thinking.}
Table~\ref{tab:mmlu-domain} fixes the answer interface to four-choice MCQ and varies the MMLU subject area.
Across all four models, STEM combines the highest accuracy with the lowest \ETR{} (0.3--36.4\%), arguing against a simple positive relationship between task success and answer-only compliance.
STEM also has the highest \QRel{} (0.509--0.784) and \EIR{} (59.7--85.5\%) in every model block, showing that the domain pattern is not confined to \ETR{}.
Thus, under this fixed interface, answer-only compliance is not simply a proxy for task success, and answer space does not account for all variation in no-thinking behavior.

\begin{table}[t]
\begin{maintablebox}
\centering
\tiny
\setlength{\tabcolsep}{2.0pt}
\renewcommand{\arraystretch}{1.08}

\caption{
\textbf{No-thinking behavior varies by domain even under the same four-choice answer interface.}
Following the official MMLU subject-area split~\citep{hendrycks2020measuring}, we report Humanities, Social Sciences, and STEM and omit the heterogeneous Other category.
The \% column gives each area's share of the full MMLU test split; for each model, the remaining columns report Acc., \ETR{}, \QRel{}, and \EIR{} under M2.
}
\label{tab:mmlu-domain}
\label{tab:domain-gradient}

\resizebox{\linewidth}{!}{%
\begin{tabular}{@{}lc*{16}{c}@{}}
\toprule
& & \multicolumn{4}{c}{\textbf{Qwen3-4B}}
& \multicolumn{4}{c}{\textbf{Qwen3-32B}}
& \multicolumn{4}{c}{\textbf{Qwen3-235B}}
& \multicolumn{4}{c}{\textbf{DeepSeek-V4-Flash}} \\
\cmidrule(lr){3-6} \cmidrule(lr){7-10} \cmidrule(lr){11-14} \cmidrule(l){15-18}
\textbf{MMLU area} & \textbf{\%}
& \textbf{Acc.} & \textbf{\ETR{}} & \textbf{\QRel{}} & \textbf{\EIR{}}
& \textbf{Acc.} & \textbf{\ETR{}} & \textbf{\QRel{}} & \textbf{\EIR{}}
& \textbf{Acc.} & \textbf{\ETR{}} & \textbf{\QRel{}} & \textbf{\EIR{}}
& \textbf{Acc.} & \textbf{\ETR{}} & \textbf{\QRel{}} & \textbf{\EIR{}} \\
\midrule

Humanities & 33.5\%
& \accval{31}{62.5} & \etrval{12}{36.8} & \simval{24}{0.488} & \eirval{28}{56.6}
& \accval{36}{71.6} & \etrval{32}{95.5} & \simval{2}{0.036} & \eirval{2}{4.1}
& \accval{39}{77.1} & \etrval{14}{40.6} & \simval{22}{0.435} & \eirval{25}{49.8}
& \accval{42}{83.5} & \etrval{1}{1.7} & \simval{39}{0.772} & \eirval{41}{82.7} \\

Social Sciences & 21.9\%
& \accval{39}{78.7} & \etrval{26}{78.2} & \simval{8}{0.167} & \eirval{8}{16.2}
& \accval{44}{88.3} & \etrval{29}{87.7} & \simval{5}{0.098} & \eirval{5}{10.7}
& \accval{44}{88.3} & \etrval{14}{41.4} & \simval{23}{0.452} & \eirval{25}{49.6}
& \accval{46}{92.3} & \etrval{1}{2.2} & \simval{39}{0.777} & \eirval{35}{69.2} \\

STEM & 22.5\%
& \accval{42}{83.0} & \etrval{11}{34.3} & \simval{26}{0.527} & \eirval{30}{60.2}
& \accval{46}{91.3} & \etrval{12}{36.4} & \simval{25}{0.509} & \eirval{30}{59.7}
& \accval{45}{90.6} & \etrval{2}{6.2} & \simval{38}{0.752} & \eirval{43}{85.5}
& \accval{47}{94.4} & \etrval{0}{0.3} & \simval{39}{0.784} & \eirval{42}{83.5} \\

\bottomrule
\end{tabular}%
}

\vspace{2pt}
\begin{minipage}{0.98\linewidth}
\footnotesize
\textit{Note.} For API models, Acc./ETR use the complete MMLU aggregate, whereas
QRel./EIR use a 1,000-output domain subset; details are
provided in Appendix Table~\ref{tab:domain-gradient-full-app}.
\end{minipage}
\end{maintablebox}
\end{table}

\subsection{Performance under answer-space rewrites}
\label{sec:answer-space-rewrites}

The preceding experiments vary thinking controls across task families. We now keep the underlying math questions fixed and change only the answer interface, allowing a more controlled comparison.

\begin{figure}[!t]
    \centering
    \vspace{-0.4em}
    \includegraphics[width=\linewidth]{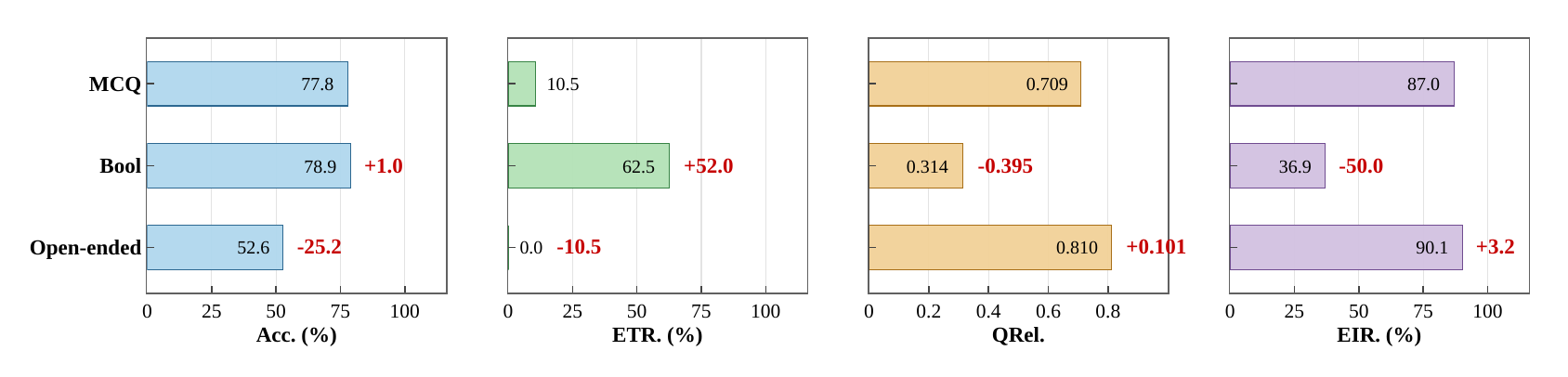}
    \vspace{-0.6em}
\caption{
\textbf{Changing only the answer space reshapes no-thinking behavior on the same questions.}
We rewrite matched numeric-answer items from MMLU and MMLU-Pro for Qwen3-32B under M2 from MCQ into Boolean or open-ended forms.
The four panels report Acc., \ETR{}, \QRel{}, and \EIR{} on the same answer-space variants; red values give the change from the MCQ form, computed before rounding.
}
\label{fig:answer-space-rewrite}
    \vspace{-0.6em}
\end{figure}

\paragraph{Changing only the answer interface reshapes no-thinking behavior.}
Figure~\ref{fig:answer-space-rewrite} rewrites the matched numeric-answer subset of MMLU/MMLU-Pro for Qwen3-32B under M2 from its original MCQ form into Boolean verification or open-ended generation.
The Boolean rewrite leaves accuracy nearly unchanged, from 77.8\% to 78.9\%, but raises \ETR{} from 10.5\% to 62.5\%, lowers \QRel{} from 0.709 to 0.314, and lowers \EIR{} from 86.95\% to 36.90\%.
By contrast, the open-ended rewrite reduces accuracy to 52.6\%, drives \ETR{} to 0\%, and raises \QRel{} to 0.810 and \EIR{} to 90.14\%.
Thus, changing only how the answer is requested substantially changes response-level no-thinking behavior, even for the same underlying questions.
A full-page qualitative example tracing all six modes on one rewritten item is provided on the following page.
\FloatBarrier
\includepdfafterpage[
    pages=1,
    pagecommand={\thispagestyle{plain}}
]{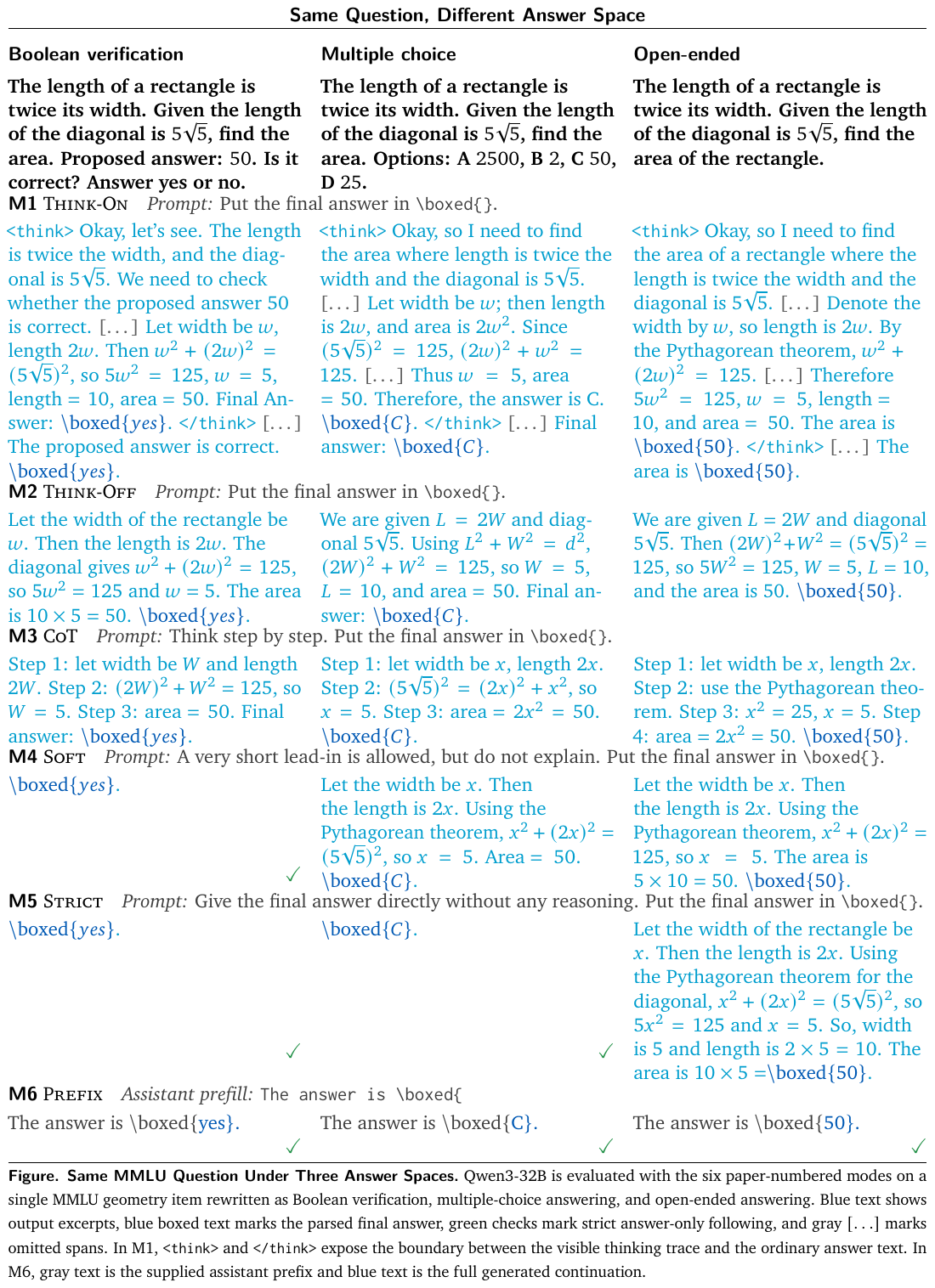}

\paragraph{Answer presentation shapes direct answering.}
This controlled comparison clarifies what the earlier answer-space trend reflects.
Boolean verification gives the model a proposed answer to check, while open-ended generation requires the model to produce the value without such support.
The original MCQ form lies between these cases: it provides options, but the model may still need visible intermediate steps to compare or compute them.
The same ordering appears in the three-level view: Boolean verification has the highest \ETR{} and the lowest \QRel{} and \EIR{}, whereas open-ended generation has no answer-only responses and retains both relevant pre-answer text and explicit inference.
Thus, the answer-space pattern in Sections~\ref{sec:thinking-inertia} and~\ref{sec:no-think-performance} appears both across benchmark families and within the same questions when the answer interface changes.

\begin{findingbox}
\textbf{Even for the same question, the answer space shapes whether an LLM answers directly.}
Candidate-visible formats make answer-only responses easier to produce than open-ended generation, so the broader pattern cannot be reduced to differences between benchmark families.
\end{findingbox}

\subsection{Human Validation}
\label{sec:human-validation}

We report the metric- and conclusion-level human-study results using the validation design in Section~\ref{sec:s3-measurement} and the experimental protocol in Section~\ref{sec:exp-setup}.

\paragraph{Metric validation.}
At \textbf{Level 1}, the strict-compliance decision underlying \ETR{} agrees with the manually inspected empty-$T$ indicator on 565 of 600 outputs (94.2\%); the remaining 35 discrepancies reflect the additional requirement that the final answer be parseable in the permitted format.
At \textbf{Level 2}, the primary instruction-aware Qwen3-Embedding-4B \QRel{} scores reach a mean item-level Spearman correlation of $\rho=0.829$ with the five annotators' relevance judgments, compared with 0.809 for all-MiniLM-L6-v2 (Table~\ref{tab:human-qrel-main}).

\begin{wraptable}[9]{r}{0.50\textwidth}
\vspace{-0.5em}
\begin{maintablebox}[left=4pt,right=4pt,top=4pt,bottom=4pt]
\centering
\scriptsize
\setlength{\tabcolsep}{3.6pt}
\renewcommand{\arraystretch}{1.02}
\captionsetup{font=scriptsize,skip=3pt}
\caption{Human and LLM-judge explicit-inference rates across the six sampled M2/M5 answer-space cells. All entries use the all-output denominator within each cell; the Human mean row averages the five annotator-specific rates.}
\label{tab:human-eir-main}
\begin{tabular}{@{}lccc|ccc@{}}
\toprule
& \multicolumn{3}{c}{\textbf{M2: Think-Off}} & \multicolumn{3}{c}{\textbf{M5: Strict}} \\
\textbf{Evaluator} & \textbf{Bool} & \textbf{MCQ} & \textbf{Open} & \textbf{Bool} & \textbf{MCQ} & \textbf{Open} \\
\midrule
GPT-5.5 & 4.0\% & 29.6\% & 100.0\% & 0.0\% & 11.8\% & 57.1\% \\
Claude Opus 4.8 & 0.0\% & 29.6\% & 100.0\% & 0.0\% & 11.8\% & 53.6\% \\
\textbf{Human mean} & \textbf{4.0\%} & \textbf{30.4\%} & \textbf{100.0\%} & \textbf{0.0\%} & \textbf{11.8\%} & \textbf{57.1\%} \\
\bottomrule
\end{tabular}
\end{maintablebox}
\vspace{-0.5em}
\end{wraptable}

At \textbf{Level 3}, the human explicit-inference labels achieve 98.8\% mean binary agreement with the primary GPT-5.5 judge and 96.7\% with the independent Claude Opus 4.8 judge across all 600 outputs.
Table~\ref{tab:human-eir-main} further shows that the corresponding all-output rates closely track the human mean across the six sampled M2/M5 answer-space cells; Appendix~\ref{app:human-validation} provides the per-annotator agreement and reliability statistics.

\begin{wraptable}[6]{r}{0.54\textwidth}
\vspace{-2em}
\begin{maintablebox}[left=4pt,right=4pt,top=4pt,bottom=4pt]
\centering
\scriptsize
\setlength{\tabcolsep}{3.2pt}
\renewcommand{\arraystretch}{1.02}
\captionsetup{font=scriptsize,skip=3pt}
\caption{Human relevance calibration on the unified 600-output sample. Each entry is an item-level Spearman correlation over all 600 outputs; empty $T$ receives relevance score 1 and \QRel{} 0, and the final column reports the mean across annotators.}
\label{tab:human-qrel-main}
\begin{tabular}{@{}lrrrrrr@{}}
\toprule
& \textbf{A1} & \textbf{A2} & \textbf{A3} & \textbf{A4} & \textbf{A5} & \textbf{Mean} \\
\midrule
Qwen3-Embedding-4B \QRel{} & 0.828 & 0.831 & 0.830 & 0.833 & 0.823 & \textbf{0.829} \\
all-MiniLM-L6-v2 & 0.794 & 0.811 & 0.814 & 0.827 & 0.798 & 0.809 \\
\bottomrule
\end{tabular}
\end{maintablebox}
\vspace{1em}
\end{wraptable}
\vspace{0.8em}

\paragraph{Conclusion validation.}
The mean human rates in Table~\ref{tab:human-eir-main} preserve both experimental trends.
Under M2 and M5, explicit inference rises from Bool to MCQ to Open-ended; relative to M2, M5 lowers the rate in every answer space, including from 100.0\% to 57.1\% on Open-ended tasks.
\par
\vspace{0.7em}
\noindent
These direct annotations therefore reproduce the two conclusions drawn in Sections~\ref{sec:thinking-inertia} and~\ref{sec:no-think-performance}: visible explicit inference becomes harder to suppress as the answer space opens and remains common on Open-ended tasks even under the strictest natural-language instruction.

\begin{findingbox}
\textbf{Human judgments support the three-level evaluation and its central conclusions.}
The boundary audit corroborates \ETR{}, relevance judgments correlate with \QRel{}, and explicit-inference labels agree strongly with the primary and independent LLM judges. The same annotations reproduce the answer-space staircase and persistent visible inference on Open-ended tasks under M5.
\end{findingbox}

%% file: section/relatedwork.tex
\section{Related Work}
\label{sec:related}

\paragraph{Efficient and Adaptive Reasoning.}
Test-time computation can substantially improve language-model reasoning~\citep{snell2024scaling,brown2024largelanguagemonkeys}, motivating methods that allocate this computation more selectively. One line of work controls the length or budget of visible reasoning through length-conditioned reinforcement learning~\citep{aggarwal2025l1}, token-budget prompting~\citep{han2024tokenbudget}, budget forcing~\citep{muennighoff2025s1}, and self-training for concise traces~\citep{munkhbat2025concise}. Another compresses intermediate reasoning by distilling shorter chains~\citep{kang2024c3ot} or using terse drafts~\citep{xu2025chainofdraft}; a recent survey provides a broader taxonomy~\citep{sui2025efficient}. Empirical studies further show that more reasoning is not uniformly beneficial: unnecessarily long traces can impair efficiency and even accuracy on simple problems, while the useful reasoning length depends on both task difficulty and model capability~\citep{chen2024overthinking,wu2025whenmore,zhang2025system1}. Recent methods therefore train models to decide whether extended reasoning is needed~\citep{fang2025thinkless,zhang2025adaptthink,jiang2025hybridthink,tu2025learningwhentothink}. Selective reasoning has also been studied in vision-language models~\citep{wang2025think,lei2026thinklessactearly}.

Our scope differs from optimizing the amount of test-time computation. We ask whether a requested no-thinking state is realized in the \emph{observable response}, following the common think/no-think focus on exposed CoT or reasoning channels rather than unverifiable latent cognition. We disable provider thinking interfaces or impose answer-only formats, then separately measure strict answer-only compliance, pre-answer relevance, and visible explicit inference. This distinction matters because intermediate computation need not be expressed in natural language, as demonstrated by latent-reasoning methods such as Coconut~\citep{hao2025coconut}, and because a visible CoT need not faithfully reflect the computation that produced an answer~\citep{turpin2023unfaithful,lanham2023faithfulness}.
Recent work also studies whether concealed reasoning can be reconstructed or exposed. Trace Inversion synthesizes distillation-useful reasoning traces from black-box inputs, answers, and optional summaries without access to hidden CoT~\citep{zhang2026stealreasoning}. Encrypted reasoning blocks exposed by proprietary APIs can also be transferred to weaker models within the same provider and decoded into plaintext~\citep{panfilov2026stealing}.
Zero-CoT Probe suppresses explicit reasoning through assistant prefilling in open-weight models and strict prompting in closed-source models to probe memorization and data contamination~\citep{lan2026illusionreasoningexposingevasive}.
Prompt repetition offers a complementary input-side technique for improving non-reasoning models and evaluates final-answer accuracy, length, and latency~\citep{leviathan2025prompt}. It does not ask whether visible question-conditioned text remains, nor does it distinguish paraphrases and relevant assertions from inference. Our contribution instead decomposes the response into pre-answer text and the final answer and evaluates input- and output-side interventions through \ETR{}, \QRel{}, \EIR{}, task accuracy, and same-question answer-space controls.

\paragraph{Instruction Following.}
Instruction following also underpins our intervention spectrum: the ``no-think,'' ``answer-only,'' and prefix-forcing modes are all instructions whose effectiveness depends on how reliably models obey natural-language constraints. Instruction tuning~\citep{wei2022finetuned,sanh2022multitask,chung2024scaling} and human-feedback alignment~\citep{ouyang2022instructgpt,bai2022constitutional} train models to follow open-ended directives, while self-generated and synthetic instruction data~\citep{wang2023selfinstruct,taori2023alpaca} broaden the instruction distribution. Verifiable-instruction benchmarks such as IFEval~\citep{zhou2023ifeval} and FollowBench~\citep{jiang2024followbench} evaluate compliance with output-format and content constraints, and related benchmarks study model behavior under complex or multi-constraint instructions~\citep{he2024complexbench,qin2024infobench}. Our work uses this perspective in a targeted way: we treat ``produce no \textit{visible} reasoning'' as one such verifiable output constraint, and study whether surface compliance actually removes visible question-conditioned payload rather than only matching the requested format.

%% file: section/discussion.tex
\section{Discussion and Conclusion}
\label{sec:discussion}

We argue that ``no-thinking'' in LLMs should be evaluated as a response-level property rather than inferred from surface proxies such as missing reasoning tags, no-think settings, or short generations. By decomposing each output into pre-answer text $T$ and final answer $A$ (the first boxed answer $A$ and residual text $T$ for M6), we progressively measure strict answer-only compliance, pre-answer relevance, and visible explicit inference, while task accuracy evaluates $A$ independently. Across local Qwen3 models, scaled variants, and frontier APIs, we find that explicit no-think controls do not reliably remove visible explicit inference; models exhibit ``\emph{thinking inertia}'' organized by answer space; stronger answer-only constraints reveal a \emph{no-thinking--performance trade-off} between compliance and accuracy; and compressibility is governed by the structural support of the question itself, not by question difficulty or model capability alone. These findings suggest that building LLMs that can stop thinking on demand requires more than hiding reasoning traces or enforcing shorter outputs: it requires evaluating when direct answering is actually supported by the task.

\subsection{Scope and Limitations}
\label{sec:discussion-limitations}

Our analysis deliberately takes the \emph{observable response} as its unit of study.
This choice provides a common behavioral basis across local models and proprietary APIs whose internal computation is neither equally exposed nor directly comparable.
Our claims therefore concern visible output, not the absence of task-relevant computation from hidden states; provider signals are incorporated when available without treating them as equivalent descriptions of internal reasoning.

Our six-model, six-benchmark suite is designed to isolate answer-space structure under deterministic decoding whenever supported.
Generalization to multilingual, code, tool-use, multi-turn, and stochastic settings, as well as robustness to evolving proprietary endpoints, remains open.
\QRel{} is an operational relevance complement to the exact \ETR{} boundary rather than a standalone reasoning detector. \EIR{} directly separates explicit inference from paraphrases and relevant non-inferential text, but remains dependent on the visible-text rubric and judge calibration; Appendix~\ref{app:three-level-validation} tests this dependence with an independent judge family, and Appendix~\ref{app:measurement-audits} audits near-verbatim copying, semantic paraphrases, and response order. These boundaries delimit the claims while preserving a reproducible, provider-agnostic basis for studying \nothink{}.

\subsection{Implications and Future Work}
\label{sec:discussion-future}

No-thinking is a distinct controllable capability rather than the default complement of a reasoning mode.
Direct answering depends on answer-space support and computational necessity rather than on a switch or model scale alone.
This suggests task-aware routing: answer directly when the interface provides sufficient support, and preserve explicit computation when suppression would sacrifice accuracy.
Visible compliance should not, however, be equated with lower inference cost when hidden reasoning may remain active.

One direction is to train models for calibrated stopping rather than uniform suppression, jointly optimizing accuracy, answer-only compliance, visible relevance and inference, latency, and actual reasoning-token cost.
Extensions across languages and interactive tasks can test generality, while internal-state analyses can examine whether computation disappears, moves to a hidden channel, or is only omitted from view.
In high-stakes settings, concise outputs may reduce cost and disclosure, but explanations remain important for verification and error detection.
The aim is to make models' decisions about \emph{when} and \emph{how} to expose intermediate computation controllable and auditable, rather than to suppress such computation universally.

%% file: section/appendix.tex
\appendix
\raggedbottom
\newcommand{\legacyM}{\ensuremath{\mathrm{M6}^{\mathrm{u}}}}

\renewcommand{\thetable}{\Alph{section}.\arabic{table}}
\renewcommand{\thefigure}{\Alph{section}.\arabic{figure}}
\renewcommand{\theHtable}{appendix.\Alph{section}.\arabic{table}}
\renewcommand{\theHfigure}{appendix.\Alph{section}.\arabic{figure}}
\setcounter{table}{0}
\setcounter{figure}{0}
\makeatletter
\@addtoreset{table}{section}
\@addtoreset{figure}{section}
\makeatother
\newcommand{\appliteacc}[1]{\begingroup\setlength{\fboxsep}{0.45pt}\colorbox{accHeat!14}{\strut #1}\endgroup}
\newcommand{\applitesr}[1]{\begingroup\setlength{\fboxsep}{0.45pt}\colorbox{srHeat!14}{\strut #1}\endgroup}
\newcommand{\appliteqrel}[1]{\begingroup\setlength{\fboxsep}{0.45pt}\colorbox{simHeat!14}{\strut #1}\endgroup}
\newcommand{\appdrop}[1]{\begingroup\setlength{\fboxsep}{0.45pt}\colorbox{negDrop!14}{\strut #1}\endgroup}
\newcommand{\appaccshade}[2]{\begingroup\setlength{\fboxsep}{0.45pt}\colorbox{accHeat!#1}{\strut #2}\endgroup}
\newcommand{\appsrshade}[2]{\begingroup\setlength{\fboxsep}{0.45pt}\colorbox{srHeat!#1}{\strut #2}\endgroup}
\newcommand{\appqrelshade}[2]{\begingroup\setlength{\fboxsep}{0.45pt}\colorbox{simHeat!#1}{\strut #2}\endgroup}
\newcommand{\appdropshade}[2]{\begingroup\setlength{\fboxsep}{0.45pt}\colorbox{negDrop!#1}{\strut #2}\endgroup}
\newcommand{\apphmcell}[4]{%
  \filldraw[fill=#3, draw=white, line width=0.35pt] (#1,#2) rectangle ++(1,1);
  \node[font=\tiny, inner sep=0pt] at ({#1+0.5},{#2+0.5}) {#4};
}

\section{Appendix Overview}
\label{app:overview}

We report the protocols, robustness checks, aggregate results, question-structure controls, and representative outputs underlying the main-text results. Appendix~\ref{app:m6-prefill-protocol} gives the assistant-prefill M6 protocol and results. We denote the separate user-message-prefix condition by \legacyM{}; its implementation and results are described in Appendix~\ref{app:historical-user-prefix}.

\section{Experimental Details}
\label{app:prompts}

\subsection{Dataset Sizes}
\label{app:dataset-sampling}

Table~\ref{tab:dataset-sampling-app} lists the full benchmark splits. The number of evaluated items depends on the model and intervention.

\begin{table}[H]
\centering
\small
\setlength{\tabcolsep}{7pt}
\renewcommand{\arraystretch}{1.05}
\caption{Full benchmark split sizes, used by the local Qwen3-4B/32B runs under M1--M5 and \legacyM{}.}
\label{tab:dataset-sampling-app}
\begin{tabular}{@{}lll@{}}
\toprule
\textbf{Dataset} & \textbf{Split} & \textbf{Size used} \\
\midrule
BoolQ & Validation & 3,270 \\
StrategyQA & Test & 687 \\
MMLU & Test & 14,042 \\
MMLU-Pro & Test & 12,032 \\
GSM8K & Test & 1,319 \\
MATH & Test & 5,000 \\
\bottomrule
\end{tabular}
\end{table}

For M1--M5 and \legacyM{}, the local Qwen3-4B/32B runs use the full benchmark splits. Each of the four API models uses 1,000 sampled items from BoolQ, MMLU, MMLU-Pro, GSM8K, and MATH, together with all 687 StrategyQA items, giving 5,687 outputs per mode and 34,122 outputs across these six conditions. M6 uses 5,687 items per model for all four models; its item manifest and family counts are given in Appendix~\ref{app:m6-prefill-protocol}.

\subsection{Models and Inference Settings}
\label{app:model-settings}

For M1--M5 and \legacyM{}, we use the models' public thinking-control interfaces. Appendix~\ref{app:m6-prefill-protocol} describes the assistant-prefill interfaces and deployment settings for M6.

\begingroup
\small
\setlength{\tabcolsep}{3.2pt}
\renewcommand{\arraystretch}{1.12}
\setlength{\LTleft}{0pt}
\setlength{\LTright}{\fill}
\setlength{\LTcapwidth}{\textwidth}
\captionsetup{width=\textwidth}
\begin{longtable}{@{}p{0.23\textwidth}p{0.32\textwidth}p{\dimexpr0.45\textwidth-4\tabcolsep\relax}@{}}
\caption{Model interfaces and decoding settings. API controls and ``all modes'' budgets refer to M1--M5 and \legacyM{}; M6 deployment settings are given in Appendix~\ref{app:m6-prefill-protocol}.}
\label{tab:model-settings-app}\\
\toprule
\textbf{Model group} & \textbf{Thinking control} & \textbf{Decoding and evaluation scope} \\
\midrule
\endfirsthead
\multicolumn{3}{@{}l}{\small\itshape\tablename~\thetable{} (continued)} \\
\toprule
\textbf{Model group} & \textbf{Thinking control} & \textbf{Decoding and evaluation scope} \\
\midrule
\endhead
\midrule
\multicolumn{3}{r@{}}{\footnotesize\itshape Continued on next page} \\
\endfoot
\bottomrule
\endlastfoot

Qwen3-4B / Qwen3-32B
& Local Qwen chat-template switch: \texttt{enable\_thinking=true} for Mode~1 and \texttt{false} for M2--M5 and \legacyM{}.
& Temperature 0.0. Response budgets: 4,096 tokens for M1; 1,024 for M2--M5 and \legacyM{}; 8,192 for M6. \\

Qwen3-235B-A22B
& API reasoning control: high effort for Mode~1 and no-reasoning setting for M2--M5 and \legacyM{}.
& Temperature 0.0. Max tokens: 8,192 for all modes. \\

DeepSeek-V4-Flash
& Provider thinking interface: high effort for Mode~1 and disabled for M2--M5 and \legacyM{}.
& Temperature 0.0. Max tokens: 8,192 for all modes. \\

GPT-5.4
& Reasoning effort: high for Mode~1 and none for M2--M5 and \legacyM{}.
& Temperature 0.0. Max tokens: 8,192 for all modes. \\

Gemini-3-Flash-Preview
& Reasoning effort: high for Mode~1 and minimal for M2--M5 and \legacyM{}.
& Temperature 0.0. Max tokens: 8,192 for all modes. The minimal setting is treated as near-off rather than strict off. \\
\end{longtable}
\endgroup

We set temperature to zero wherever the endpoint supports it.
The M1--M5 and \legacyM{} API runs use an 8,192-token response budget. Local Qwen budgets vary by condition, as listed in Table~\ref{tab:model-settings-app}; M6 uses 8,192 tokens for all four models (Appendix~\ref{app:m6-prefill-protocol}).
For Qwen3-235B-A22B, DeepSeek-V4-Flash, and GPT-5.4, truncation affects at most 0.35\% of outputs in any M2/M5 answer-space cell under the 8,192-token budget.
For M1--M5 and \legacyM{}, non-text provider reasoning signals enter the strict answer-only decision, and returned readable reasoning contributes to the visible text.

\section{Evaluation Protocol}
\label{app:evaluation-protocol}

\paragraph{Protocol summary.}
M1--M5 and \legacyM{} use the response-level setup from Section~\ref{sec:s3-measurement}, with pre-answer text $T$ and final answer $A$. Non-text reasoning signals enter the answer-only check, and returned readable reasoning fields contribute to $T$.
For M6, $A$ is the first boxed answer and $T$ contains the residual text from all returned readable fields (Appendix~\ref{app:m6-prefill-protocol}).

\begingroup
\small
\setlength{\tabcolsep}{4pt}
\renewcommand{\arraystretch}{1.08}
\setlength{\LTleft}{0pt}
\setlength{\LTright}{\fill}
\setlength{\LTcapwidth}{\textwidth}
\captionsetup{width=\textwidth}
\begin{longtable}{@{}p{0.22\textwidth}p{0.38\textwidth}p{\dimexpr0.40\textwidth-4\tabcolsep\relax}@{}}
\caption{Scope of the M1--M5 and \legacyM{} result blocks and question-structure controls. Assistant-prefill M6 results are reported in Appendix~\ref{app:m6-prefill-protocol}.}
\label{tab:evaluation-scope-app}\\
\toprule
\textbf{Result block} & \textbf{Models and task scope} & \textbf{Appendix evidence supported} \\
\midrule
\endfirsthead
\multicolumn{3}{@{}l}{\small\itshape\tablename~\thetable{} (continued)} \\
\toprule
\textbf{Result block} & \textbf{Models and task scope} & \textbf{Appendix evidence supported} \\
\midrule
\endhead
\midrule
\multicolumn{3}{r@{}}{\footnotesize\itshape Continued on next page} \\
\endfoot
\bottomrule
\endlastfoot

Local Qwen spectrum
& Qwen3-4B and Qwen3-32B on the six-benchmark suite: BoolQ, StrategyQA, MMLU, MMLU-Pro, GSM8K, and MATH, under M1--M5 and \legacyM{}.
& Global local-model summary and Qwen scaling evidence in Figure~\ref{fig:global-six-mode-app} and Table~\ref{tab:qwen-scaling-modes-app}. \\
Frontier model results
& Qwen3-235B-A22B, DeepSeek-V4-Flash, GPT-5.4, and Gemini-3-Flash-Preview on the same six-benchmark suite, using each provider's public thinking-control interface.
& Frontier native no-think and suppression/forcing results in Figures~\ref{fig:frontier-native-no-think-app} and~\ref{fig:focusness-heatmap-app}, plus the Qwen table in Table~\ref{tab:qwen-scaling-modes-app}. \\
MMLU domain analysis
& Qwen3-4B, Qwen3-32B, DeepSeek-V4-Flash, and Qwen3-235B-A22B, grouped by MMLU supercategory and, where applicable, subject. Acc./\ETR{} use the full split; \QRel{}/\EIR{} use the full split for the local models and 1,000 items for each API model.
& Domain and subject-level analyses in Figure~\ref{fig:domain-gradient-heatmap-app}, Table~\ref{tab:domain-gradient-full-app}, and Table~\ref{tab:mmlu-subject-breakdown-app}. \\
Surface and candidate-visibility interventions
& Qwen3-4B and Qwen3-32B on MMLU variants with choice shuffling, numeric labels, and boolean verification with a proposed answer shown.
& Question-interface controls in Tables~\ref{tab:surface-control-app} and~\ref{tab:candidate-shown-app}. \\
Numeric-source open-generation intervention
& Qwen3-4B and Qwen3-32B on matched numeric-source items, comparing MCQ, boolean verification, and open numeric generation.
& Candidate-removal stress test in Table~\ref{tab:open-numeric-app}. \\
\end{longtable}
\endgroup

\paragraph{Run aggregation.}
Dataset-level aggregates use M1--M5 as defined in Section~\ref{sec:s3-spectrum} and \legacyM{} for the user-message-prefix condition.
The grouped metric tables pool items within each answer family. In Figure~\ref{fig:frontier-family-control}, local Open-ended accuracy for M2, M4, and M5 is the mean of the two benchmark accuracies. M6 pools items within each model and family, then averages the model rates with equal weight.
The frontier M1--M5 and \legacyM{} aggregates use the 8,192-token runs with seed 20260502.
For these pre-answer conditions, \QRel{} uses the complete visible text before $A$, including readable reasoning fields and ordinary pre-answer content.
GPT-5.5 scores all outputs in the M1--M5 plus \legacyM{} corpus. The domain analysis groups the MMLU portion of these outputs by subject area. These \EIR{} scores appear in the global, frontier, scaling, and domain analyses; the question-structure controls examine relevance and consistency.

\section{Validation of the Three-Level Evaluation}
\label{app:three-level-validation}

The full-corpus encoder and judge comparisons validate the pre-answer protocol for M1--M5 and \legacyM{}.

\subsection{Level 2: Encoder Robustness}
\label{app:level2-robustness}

Qwen3-Embedding-4B is our primary Level 2 encoder because its instruction-aware retrieval interface lets the query specify the intended relation between $Q$ and $T$.
We test whether the relevance pattern depends on this representation by rescoring the complete corpus with Qwen3-Embedding-4B and all-MiniLM-L6-v2; Table~\ref{tab:qrel-full-encoder-app} reports the central M2/M5 comparison.

\begin{table}[H]
\centering
\small
\setlength{\tabcolsep}{5.5pt}
\renewcommand{\arraystretch}{1.08}
\caption{\textbf{Full-corpus Level 2 comparison.} Mean relevance scores use all outputs, with empty $T$ assigned zero. Both encoders were applied to outputs from all six models, six benchmarks, and the M1--M5 plus \legacyM{} conditions; the table reports the central M2/M5 contrast.}
\label{tab:qrel-full-encoder-app}
\begin{tabular}{@{}lccc|ccc@{}}
\toprule
& \multicolumn{3}{c}{\textbf{M2: Think-Off}} & \multicolumn{3}{c}{\textbf{M5: Strict}} \\
\textbf{Encoder} & \textbf{Bool} & \textbf{MCQ} & \textbf{Open} & \textbf{Bool} & \textbf{MCQ} & \textbf{Open} \\
\midrule
all-MiniLM-L6-v2 & 0.046 & 0.366 & 0.701 & 0.001 & 0.041 & 0.455 \\
\textbf{Qwen3-Embedding-4B} & \textbf{0.047} & \textbf{0.406} & \textbf{0.854} & \textbf{0.001} & \textbf{0.045} & \textbf{0.546} \\
\bottomrule
\end{tabular}
\end{table}

We also compare MiniLM, BGE-base-en-v1.5, and E5-base-v2 on a fixed, balanced 3,600-output sample~\citep{wang2020minilm,reimers-2019-sentence-bert,xiao2023cpack,wang2022e5}.
The sample contains 50 outputs from each of 72 cells: six models, three answer spaces, and M1/M2/M3/M5.
We use each encoder's recommended pooling and query/passage prefixes and assign a score of zero to empty $T$.
These encoders have different retrieval interfaces from Qwen3-Embedding-4B's reasoning-focused query instruction, so the comparison tests sensitivity to encoder choice.

\begin{table}[H]
\centering
\small
\setlength{\tabcolsep}{5.5pt}
\renewcommand{\arraystretch}{1.08}
\caption{\textbf{Level 2 sensitivity to commonly used encoders} on the balanced 3,600-output sample. Absolute cosine scales are encoder-specific; the relevant comparisons are the directions within each row.}
\label{tab:qrel-encoder-sensitivity-app}
\begin{tabular}{@{}lccc|ccc@{}}
\toprule
& \multicolumn{3}{c}{\textbf{M2: Think-Off}} & \multicolumn{3}{c}{\textbf{M5: Strict}} \\
\textbf{Encoder} & \textbf{Bool} & \textbf{MCQ} & \textbf{Open} & \textbf{Bool} & \textbf{MCQ} & \textbf{Open} \\
\midrule
all-MiniLM-L6-v2 & 0.119 & 0.460 & 0.677 & 0.052 & 0.132 & 0.439 \\
BGE-base-en-v1.5 & 0.284 & 0.424 & 0.759 & 0.154 & 0.262 & 0.518 \\
E5-base-v2 & 0.263 & 0.431 & 0.890 & 0.251 & 0.365 & 0.638 \\
\bottomrule
\end{tabular}
\end{table}

Across both comparisons, every encoder preserves Bool $<$ MCQ $<$ Open under M2 and M5 $<$ M2 in every answer space.

\subsection{Level 3: Cross-Family Judge Robustness}
\label{app:level3-robustness}

GPT-5.5 is the primary Level 3 judge.
We apply Claude Opus 4.8 to the complete corpus with the same blinded input, prompt, five-way rubric, and supporting-span requirement, covering all six models and benchmarks under M1--M5 plus \legacyM{}.
For each non-empty $T$, both judges receive only $(Q,T)$; empty $T$ is assigned no visible inference.
\begin{table}[H]
\centering
\small
\setlength{\tabcolsep}{5.2pt}
\renewcommand{\arraystretch}{1.08}
\caption{\textbf{Cross-family replication of the main \EIR{} result.} Values are all-output explicit-inference rates; GPT-5.5 is the primary judge and Claude Opus 4.8 is the independent cross-family judge.}
\label{tab:cross-judge-eir-app}
\begin{tabular}{@{}lccc|ccc@{}}
\toprule
& \multicolumn{3}{c}{\textbf{M2: Think-Off}} & \multicolumn{3}{c}{\textbf{M5: Strict}} \\
\textbf{Judge} & \textbf{Bool} & \textbf{MCQ} & \textbf{Open} & \textbf{Bool} & \textbf{MCQ} & \textbf{Open} \\
\midrule
GPT-5.5 & 5.15\% & 45.57\% & 99.69\% & 0.12\% & 5.39\% & 62.00\% \\
Claude Opus 4.8 & 4.24\% & 40.73\% & 98.55\% & 0.10\% & 5.11\% & 61.02\% \\
\bottomrule
\end{tabular}
\end{table}

\paragraph{Separating paraphrases from explicit inference.}
The five-way GPT-5.5 rubric distinguishes semantic paraphrases from explicit inference, including paraphrases with little lexical overlap with the question.
Table~\ref{tab:judge-paraphrase-audit-app} reports the M2/M5 rates using all outputs as the denominator for both categories.
Across M1--M5, GPT-5.5 classifies 477 of 477,240 outputs (0.100\%) as paraphrase-only. Appendix~\ref{app:historical-user-prefix} reports the corresponding total with \legacyM{} included.
Only the explicit-reasoning label contributes to \EIR{}; paraphrases and relevant non-inferential text contribute zero.

\begin{table}[H]
\centering
\small
\setlength{\tabcolsep}{5.5pt}
\renewcommand{\arraystretch}{1.08}
\caption{\textbf{Explicit inference and paraphrase-only rates under the primary GPT-5.5 judge.} Both rates use all outputs in the corresponding setting as the denominator; empty $T$ contributes to neither numerator.}
\label{tab:judge-paraphrase-audit-app}
\begin{tabular}{@{}lrrr@{}}
\toprule
\textbf{Mode} & \textbf{Answer space} & \textbf{Explicit reasoning} & \textbf{Paraphrase-only} \\
\midrule
M2 & Bool & 5.15\% & 0.03\% \\
   & MCQ  & 45.57\% & 0.11\% \\
   & Open & 99.69\% & 0.02\% \\
\midrule
M5 & Bool & 0.12\% & 0.01\% \\
   & MCQ  & 5.39\% & 0.03\% \\
   & Open & 62.00\% & 1.00\% \\
\bottomrule
\end{tabular}
\end{table}

Both judges give Bool $<$ MCQ $<$ Open under M2 and M5, and M5 $<$ M2 in every answer space.
On Open-ended tasks under M5, Opus identifies visible explicit inference in 61.02\% of outputs, compared with GPT-5.5's 62.00\%.

Appendix~\ref{app:historical-user-prefix} reports agreement between the two judges over the full M1--M5 and \legacyM{} corpus.

\subsection{Human Validation Protocol}
\label{app:human-validation}

The human study evaluates 600 outputs from M2 and M5.

\paragraph{Sampling and blinding.}
The sample contains 100 outputs from each benchmark. After rounding, the proportions of empty and non-empty $T$ match those in that benchmark's full corpus. We manually audit whether $T$ is empty for all 600 outputs. The annotation view exposes only $(Q,T)$ and hides dataset, model, mode, final answer, correctness, \QRel{}, and judge labels. Empty $T$ receives relevance score 1 and no visible inference by definition. Five annotators independently review the same 402 non-empty texts.

\paragraph{Level 2 relevance annotation.}
For every non-empty $T$, annotators assign one ordinal score: \textbf{1}, no relevant content (generic, off-topic, or unrelated text); \textbf{2}, weak relevance (mostly boilerplate or only loosely related text); \textbf{3}, on-topic but non-evidential content (a relevant restatement, definition, or assertion without question-specific support); \textbf{4}, question-specific support (at least one relevant piece of evidence, comparison, computation, or inference); or \textbf{5}, substantive multi-step support (multiple connected question-specific steps or a developed derivation). This score evaluates graded pre-answer relevance and is kept separate from the Level 3 reasoning decision.

For annotator $a$, let $h_{ai}$ be the human score and $q_i$ the \QRel{} score for output $i$. We compute item-level Spearman correlation
\begin{equation}
\rho_a=\mathrm{Corr}\!\left(\mathrm{rank}(h_{ai}),\mathrm{rank}(q_i)\right),
\end{equation}
where ties receive their average rank. Table~\ref{tab:human-qrel-main} reports every $\rho_a$ and their arithmetic mean $\bar{\rho}=\frac{1}{5}\sum_{a=1}^{5}\rho_a$.

\paragraph{Level 3 reasoning annotation.}
Independently of the relevance score, annotators choose exactly one of the same five mutually exclusive labels used by GPT-5.5 and Opus 4.8: generic/off-topic, paraphrase-only, relevant non-inferential, explicit reasoning, or unclear. The criteria are identical to Table~\ref{tab:reasoning-rubric}; in particular, only deduction, computation, comparison, elimination, rule application, evidence-to-conclusion, or causality enters the explicit-reasoning category. A restatement or relevant assertion without an inferential step therefore does not enter the \EIR{} numerator. Annotators also record a supporting span from $T$.

\paragraph{Aggregation and agreement.}
All rates in Table~\ref{tab:human-eir-main} use the full number of sampled outputs in each M2/M5 answer-space cell, including empty $T$, and the human row is the arithmetic mean of the five annotator-specific rates. We compute per-example five-way agreement using exact rubric-label matches and binary agreement by collapsing the rubric into explicit reasoning versus all other labels. Krippendorff's $\alpha$ is computed across the five annotators on the 402 independently reviewed non-empty texts: nominal distance for the five-way and binary reasoning labels, and ordinal distance for the five-point relevance scores.

\begin{table}[H]
\centering
\scriptsize
\setlength{\tabcolsep}{4.0pt}
\renewcommand{\arraystretch}{1.04}
\caption{Per-annotator relevance--\QRel{} calibration on the 600-output human sample. Values are item-level Spearman correlations.}
\label{tab:human-qrel-appendix}
\begin{tabular}{@{}lrrrrrr@{}}
\toprule
\textbf{Encoder} & \textbf{A1} & \textbf{A2} & \textbf{A3} & \textbf{A4} & \textbf{A5} & \textbf{Mean} \\
\midrule
Qwen3-Embedding-4B & 0.828 & 0.831 & 0.830 & 0.833 & 0.823 & \textbf{0.829} \\
all-MiniLM-L6-v2 & 0.794 & 0.811 & 0.814 & 0.827 & 0.798 & 0.809 \\
\bottomrule
\end{tabular}
\end{table}

\begin{table}[H]
\centering
\scriptsize
\setlength{\tabcolsep}{4.0pt}
\renewcommand{\arraystretch}{1.04}
\caption{Per-annotator agreement with the blinded Level 3 judges on all 600 outputs. Five-way agreement requires the same rubric label; binary agreement compares explicit-reasoning versus other labels.}
\label{tab:human-judge-appendix}
\begin{tabular}{@{}lrrrrrr@{}}
\toprule
\textbf{Judge / agreement} & \textbf{A1} & \textbf{A2} & \textbf{A3} & \textbf{A4} & \textbf{A5} & \textbf{Mean} \\
\midrule
GPT-5.5 / Five-way & 98.50\% & 97.83\% & 98.17\% & 97.33\% & 97.50\% & \textbf{97.87\%} \\
GPT-5.5 / Binary & 98.67\% & 98.83\% & 99.00\% & 98.83\% & 98.67\% & \textbf{98.80\%} \\
Opus 4.8 / Five-way & 96.00\% & 95.67\% & 95.83\% & 95.00\% & 95.17\% & \textbf{95.53\%} \\
Opus 4.8 / Binary & 96.50\% & 97.00\% & 96.83\% & 96.67\% & 96.50\% & \textbf{96.70\%} \\
\bottomrule
\end{tabular}
\end{table}

\begin{table}[H]
\centering
\scriptsize
\setlength{\tabcolsep}{4.0pt}
\renewcommand{\arraystretch}{1.04}
\caption{Inter-annotator reliability on the 402 non-empty pre-answer texts.}
\label{tab:human-iaa-appendix}
\begin{tabular}{@{}lr@{}}
\toprule
\textbf{Annotation variable} & \textbf{Krippendorff's $\alpha$} \\
\midrule
Five-way reasoning label (nominal) & 0.873 \\
Explicit reasoning vs. other (nominal) & 0.948 \\
Five-point relevance score (ordinal) & 0.919 \\
\bottomrule
\end{tabular}
\end{table}

\begin{table}[H]
\centering
\scriptsize
\setlength{\tabcolsep}{3.6pt}
\renewcommand{\arraystretch}{1.04}
\caption{Per-annotator all-output explicit-inference rates for the six sampled M2/M5 cells. The main-text human row is the column-wise mean.}
\label{tab:human-eir-appendix}
\begin{tabular}{@{}llrrrrrr@{}}
\toprule
\textbf{Mode} & \textbf{Answer space} & \textbf{A1} & \textbf{A2} & \textbf{A3} & \textbf{A4} & \textbf{A5} & \textbf{Mean} \\
\midrule
M2 & Bool & 4.0\% & 4.0\% & 4.0\% & 4.0\% & 4.0\% & \textbf{4.0\%} \\
M2 & MCQ & 29.6\% & 29.6\% & 33.3\% & 29.6\% & 29.6\% & \textbf{30.4\%} \\
M2 & Open & 100.0\% & 100.0\% & 100.0\% & 100.0\% & 100.0\% & \textbf{100.0\%} \\
M5 & Bool & 0.0\% & 0.0\% & 0.0\% & 0.0\% & 0.0\% & \textbf{0.0\%} \\
M5 & MCQ & 11.8\% & 11.8\% & 11.8\% & 11.8\% & 11.8\% & \textbf{11.8\%} \\
M5 & Open & 57.1\% & 57.1\% & 57.1\% & 57.1\% & 57.1\% & \textbf{57.1\%} \\
\bottomrule
\end{tabular}
\end{table}

The evaluation artifacts include the per-example annotations and fixed sample IDs. Across the six cells, the ranking by mean human rate matches the rankings by mean instruction-aware \QRel{} ($\rho=1.000$) and GPT-5.5 \EIR{} ($\rho=1.000$), and nearly matches the ranking by Opus 4.8 \EIR{} ($\rho=0.986$). Judge--annotator agreement is computed item by item over all sampled outputs.

\section{Measurement Robustness Audits}
\label{app:measurement-audits}

We audit response order in 477,240 M1--M5 outputs from six models and six datasets, then examine the 22,748 assistant-prefill M6 outputs under the residual-text protocol.

\subsection{Response Order and Restatement under M1--M5}

\paragraph{Near-verbatim question repetition is rare under a lexical audit.}
We check M1--M5 outputs for near-verbatim repetition of the question.
The detector compares the question stem, without appended option labels or instruction wrappers, with four fields. These are \texttt{extracted\_t} and \texttt{content}, together with \texttt{reasoning} and \texttt{reasoning\_content}.
It removes complete boxed-answer spans, lowercases the text, and tokenizes ASCII letters and numbers.
A record is flagged if any component has at least four tokens, differs from the question's token length by at most a factor of 1.5, and shares at least 80\% of the distinct consecutive four-grams in \emph{both} directions.
This flags 176 of 477,240 records (0.037\%); even assigning all flagged \QRel{} values to zero changes the corpus mean by at most $176/477{,}240=0.00037$.
Including \legacyM{}, the same detector flags 310 of 572,688 outputs (0.054\%; Appendix~\ref{app:historical-user-prefix}).

\paragraph{Semantic paraphrases do not enter the inference numerator.}
The primary judge classifies 477 M1--M5 outputs (0.100\%) as paraphrase-only.
Only the explicit-reasoning label contributes to \EIR{}; paraphrases and relevant non-inferential text do not.
Table~\ref{tab:judge-paraphrase-audit-app} gives the central M2/M5 breakdown.

\paragraph{Answer-first responses with a suffix.}
We audit 445,180 M1--M5 outputs with a parseable \texttt{\textbackslash boxed\{\}} or \texttt{\textbackslash fbox\{\}} in the canonical answer-view content.
The strict lexical check flags a response when the first box begins after at most three letter/number tokens and at least three such tokens follow it.
It finds 62 records (0.014\% of boxed outputs; Table~\ref{tab:measurement-audit-app}).
The low rate supports the pre-answer protocol for boxed M1--M5 responses; this check captures the literal answer-first format defined above.
\ETR{} checks the complete returned output, including any substantive suffix. For \QRel{} and \EIR{}, extraction ends at the selected answer, leaving subsequent explanations outside $T$.
M6 begins with a supplied answer declaration and uses the residual-text protocol below.

\begin{table}[H]
\centering
\scriptsize
\setlength{\tabcolsep}{4pt}
\renewcommand{\arraystretch}{1.10}
\caption{\textbf{M1--M5 format audit.} Copy uses the four-gram detector and all responses as its denominator. Answer-first-plus-tail uses boxed answer-view outputs and the strict lexical rule defined above.}
\label{tab:measurement-audit-app}
\begin{tabular}{@{}lrrr@{}}
\toprule
\textbf{Model} & \textbf{Responses} & \textbf{Copy} & \textbf{Answer-first-plus-tail} \\
\midrule
Qwen3-4B & 181,750 & 117 (0.064\%) & 23/164,895 (0.014\%) \\
Qwen3-32B & 181,750 & 28 (0.015\%) & 27/170,686 (0.016\%) \\
Qwen3-235B-A22B & 28,435 & 3 (0.011\%) & 10/28,273 (0.035\%) \\
DeepSeek-V4-Flash & 28,435 & 10 (0.035\%) & 1/28,272 (0.004\%) \\
GPT-5.4 & 28,435 & 0 (0.000\%) & 0/28,417 (0.000\%) \\
Gemini-3-Flash-Preview & 28,435 & 18 (0.063\%) & 1/24,637 (0.004\%) \\
\midrule
\textbf{All} & \textbf{477,240} & \textbf{176 (0.037\%)} & \textbf{62/445,180 (0.014\%)} \\
\bottomrule
\end{tabular}
\end{table}

\subsection{Assistant-Prefill M6: Structure and Evaluation}
\label{app:m6-prefill-protocol}

\paragraph{Cohort and implementation.}
The M6 evaluation contains 22,748 generations: 5,687 per model for Qwen3-4B, Qwen3-32B, Qwen3-235B-A22B, and DeepSeek-V4-Flash.
Each model uses the same item manifest: 1,000 items from each of BoolQ, MMLU, MMLU-Pro, GSM8K, and MATH, plus all 687 StrategyQA items.
Thus, the per-model family counts are 1,687 Bool, 2,000 MCQ, and 2,000 Open-ended.
The three Qwen models run locally; generation continues from an unfinished assistant response containing \texttt{The answer is \textbackslash boxed\{}.
DeepSeek uses a prefix-continuation interface with model version 0731.
All four models use an 8,192-token response budget, temperature zero, and a configuration that requests thinking to be disabled.
GPT-5.4 and Gemini-3-Flash-Preview are omitted because we could not verify an assistant-continuation interface for these endpoints.
Relative to M2/M4/M5, M6 uses smaller samples and a longer budget for Qwen3-4B and Qwen3-32B (1,024 tokens under M2--M5), local inference for Qwen3-235B in place of its API endpoint, and a different DeepSeek version.

\paragraph{Observable response structures.}
Table~\ref{tab:m6-format-audit-app} partitions all generations by the first prefilled box and subsequent visible structure.
The answer-only test covers all returned readable fields, while the later-box counts use the content channel alone. Separately returned reasoning fields have no assumed position relative to that channel.
A response with a usable first box is answer-only when its residual text is empty. Otherwise, it has no later content-channel box, later boxes that all match the initial box, or at least one later box that differs.
Box comparisons ignore whitespace and letter case and compare literal text; a difference need not change the answer's meaning. Explicit inference in the residual text is assessed separately by the judge.
The unusable category takes precedence and contains 877 empty boxes, 532 unclosed boxes, and 447 complete boxes from which the benchmark answer normalizer extracts no answer.
Of these, 1,854 occur in DeepSeek and two in Qwen3-32B.
There are also 648 outputs with empty residual $T$ but no usable first answer; these receive \ETR{}=0.

\begin{table}[H]
\centering
\scriptsize
\setlength{\tabcolsep}{3pt}
\renewcommand{\arraystretch}{1.12}
\caption{\textbf{Response structures under assistant-prefill M6.} The five categories are mutually exclusive; parentheses give percentages of all outputs in the row. Residual text includes all returned readable text outside the first declaration. Later boxes are counted in the content channel only.}
\label{tab:m6-format-audit-app}
\resizebox{\linewidth}{!}{%
\begin{tabular}{@{}lrccccc@{}}
\toprule
\textbf{Model / family} & \textbf{$N$} & \makecell{\textbf{Answer-only}\\\textbf{(\ETR{})}} & \makecell{\textbf{Residual, no}\\\textbf{later box}} & \makecell{\textbf{Later boxes}\\\textbf{all same}} & \makecell{\textbf{A later box}\\\textbf{differs}} & \makecell{\textbf{Unusable}\\\textbf{first box}} \\
\midrule
Qwen3-4B & 5,687 & 3,293 (57.90\%) & 511 (8.99\%) & 864 (15.19\%) & 1,019 (17.92\%) & 0 (0.00\%) \\
Qwen3-32B & 5,687 & 5,427 (95.43\%) & 42 (0.74\%) & 149 (2.62\%) & 67 (1.18\%) & 2 (0.04\%) \\
Qwen3-235B & 5,687 & 3,771 (66.31\%) & 1,594 (28.03\%) & 252 (4.43\%) & 70 (1.23\%) & 0 (0.00\%) \\
DeepSeek & 5,687 & 3,302 (58.06\%) & 194 (3.41\%) & 60 (1.06\%) & 277 (4.87\%) & 1,854 (32.60\%) \\
\textbf{All} & 22,748 & 15,793 (69.43\%) & 2,341 (10.29\%) & 1,325 (5.82\%) & 1,433 (6.30\%) & 1,856 (8.16\%) \\
\midrule
Bool & 6,748 & 5,497 (81.46\%) & 769 (11.40\%) & 5 (0.07\%) & 9 (0.13\%) & 468 (6.94\%) \\
MCQ & 8,000 & 5,466 (68.33\%) & 1,219 (15.24\%) & 286 (3.57\%) & 99 (1.24\%) & 930 (11.62\%) \\
Open-ended & 8,000 & 4,830 (60.38\%) & 353 (4.41\%) & 1,034 (12.93\%) & 1,325 (16.56\%) & 458 (5.73\%) \\

\bottomrule
\end{tabular}}
\end{table}

\paragraph{First-answer accuracy and residual-text metrics.}
We parse the first prefilled box with balanced braces and pass only that isolated box to the same benchmark answer normalizer and comparator used for M1--M5.
Accuracy depends only on the first box, regardless of later corrections.
An empty, unclosed, or unparseable first answer receives Acc.=0 and \ETR{}=0; \ETR{} requires a complete box with a nonempty normalized prediction.
We remove the entire initial declaration when its box is complete, including all text inside the box. If the box never closes, we retain the complete content.
We join nonempty \texttt{reasoning}, \texttt{reasoning\_content}, readable nonsummary \texttt{reasoning\_details}, and residual content in that order, retaining the first occurrence of each part after whitespace normalization and case folding for duplicate comparison. We then append readable reasoning summaries, applying the same rule to the merged text and each summary as whole parts.
We trim and collapse whitespace in the remaining content and the merged text. A string consisting only of a single full stop (ASCII or ideographic) is treated as empty.
We pass residual $T$ to the Qwen3-Embedding-4B encoder and GPT-5.5 five-way judge, using the same query instruction and judge rubric as for pre-answer text.
Empty $T$ receives \QRel{}=0 and \EIR{}=0 without a model call.
Removing the first box also removes any equations or prose inside it, so residual-text scores exclude inference within that box.
Later answer assertions and corrections remain in $T$ and are visible to the judge.

\paragraph{Denominators and aggregation.}
Acc., \ETR{}, and \QRel{} use all 22,748 outputs. The GPT-5.5 judging API returned content-filter errors for 30 outputs, which are excluded only from \EIR{}, leaving 22,718 outputs. Table~\ref{tab:m6-residual-metrics-app} reports the family-level values and per-group denominators $N_{\mathrm{EIR}}$.
Within a model, tasks are item-weighted. The means and population standard deviations in Figure~\ref{fig:frontier-family-control} give equal weight to each available model.
Gemini is excluded from the M2/M4/M5 \ETR{} means and standard deviations in Figure~\ref{fig:frontier-family-control}.
M6 differs from M2/M4/M5 in model coverage and evaluation scope.

\begin{table}[H]
\centering
\small
\setlength{\tabcolsep}{5pt}
\renewcommand{\arraystretch}{1.06}
\caption{\textbf{M6 with first-box accuracy and evaluation of residual text from all returned readable fields.} Acc., \ETR{}, and \EIR{} are percentages. $N_{\mathrm{EIR}}$ excludes only judge-filtered records; all other columns use $N$.}
\label{tab:m6-residual-metrics-app}
\begin{tabular}{@{}llrrrrrr@{}}
\toprule
\textbf{Model} & \textbf{Family} & \textbf{$N$} & \textbf{Acc.} & \textbf{\ETR{}} & \textbf{\QRel{}} & \textbf{\EIR{}} & \textbf{$N_{\mathrm{EIR}}$} \\
\midrule
Qwen3-4B & Bool & 1,687 & 77.00 & 97.93 & 0.016 & 1.54 & 1,686 \\
Qwen3-4B & MCQ & 2,000 & 55.40 & 81.05 & 0.152 & 18.62 & 1,992 \\
Qwen3-4B & Open-ended & 2,000 & 18.60 & 1.00 & 0.793 & 98.90 & 2,000 \\
\midrule
Qwen3-32B & Bool & 1,687 & 81.56 & 100.00 & 0.000 & 0.00 & 1,687 \\
Qwen3-32B & MCQ & 2,000 & 68.10 & 99.00 & 0.008 & 0.80 & 2,000 \\
Qwen3-32B & Open-ended & 2,000 & 33.55 & 88.00 & 0.091 & 11.25 & 2,000 \\
\midrule
Qwen3-235B & Bool & 1,687 & 84.11 & 57.68 & 0.316 & 33.61 & 1,684 \\
Qwen3-235B & MCQ & 2,000 & 72.75 & 45.45 & 0.433 & 48.49 & 1,986 \\
Qwen3-235B & Open-ended & 2,000 & 40.65 & 94.45 & 0.043 & 4.90 & 2,000 \\
\midrule
DeepSeek & Bool & 1,687 & 55.96 & 70.24 & 0.070 & 2.61 & 1,687 \\
DeepSeek & MCQ & 2,000 & 43.45 & 47.80 & 0.205 & 18.09 & 1,996 \\
DeepSeek & Open-ended & 2,000 & 38.25 & 58.05 & 0.300 & 36.70 & 2,000 \\
\bottomrule
\end{tabular}
\end{table}

\paragraph{Examples and interpretation.}
Table~\ref{tab:m6-format-examples-app} illustrates several M6 response structures. The opening declaration includes the supplied prefix, and a bracketed ellipsis marks omitted text.
The pear example follows $(A,T)$. The crustacean example compares sizes before revising ``yes'' to ``no'', yielding $(A,T,A')$.
In the arithmetic example, the computation appears inside the first box and is therefore removed before residual-text evaluation.

\begin{table}[H]
\centering
\small
\setlength{\tabcolsep}{4pt}
\renewcommand{\arraystretch}{1.18}
\caption{\textbf{Examples of M6 response structure.} Text is reproduced from the stored content field with line breaks normalized. Q235 denotes Qwen3-235B; DS denotes DeepSeek.}
\label{tab:m6-format-examples-app}
\begin{tabularx}{\linewidth}{@{}p{0.26\linewidth}X@{}}
\toprule
\textbf{Source / structure} & \textbf{Observed content} \\
\midrule
Qwen3-32B, MMLU 10039; answer-only & The answer is $\boxed{\mathrm{C}}$. \\
Q235, StrategyQA 206; $(A,T)$ & The answer is $\boxed{\mathrm{no}}$. A pear typically floats in water because its density is less than that of water. \\
Q235, StrategyQA 168; $(A,T,A')$ & The answer is $\boxed{\mathrm{yes}}$. [\ldots] Since the crab's leg span is significantly larger than the dimensions of the mattress, it would not be able to stretch out completely on a king-sized mattress. Correction: The answer is $\boxed{\mathrm{no}}$. \\
Q235, GSM8K 723; computation inside box & The answer is $\boxed{170 + 23 - 20 - 13 = 160}$. \\
DS, BoolQ 2373; empty first box & The answer is $\boxed{\phantom{x}}$ no, because Hawaii is a U.S. state, not a territory. \\
DS, MMLU 12206; unclosed first box & \texttt{The answer is \textbackslash boxed\{B} \\
\bottomrule
\end{tabularx}
\end{table}

\begin{figure}[H]
\centering
\includegraphics[width=0.88\textwidth]{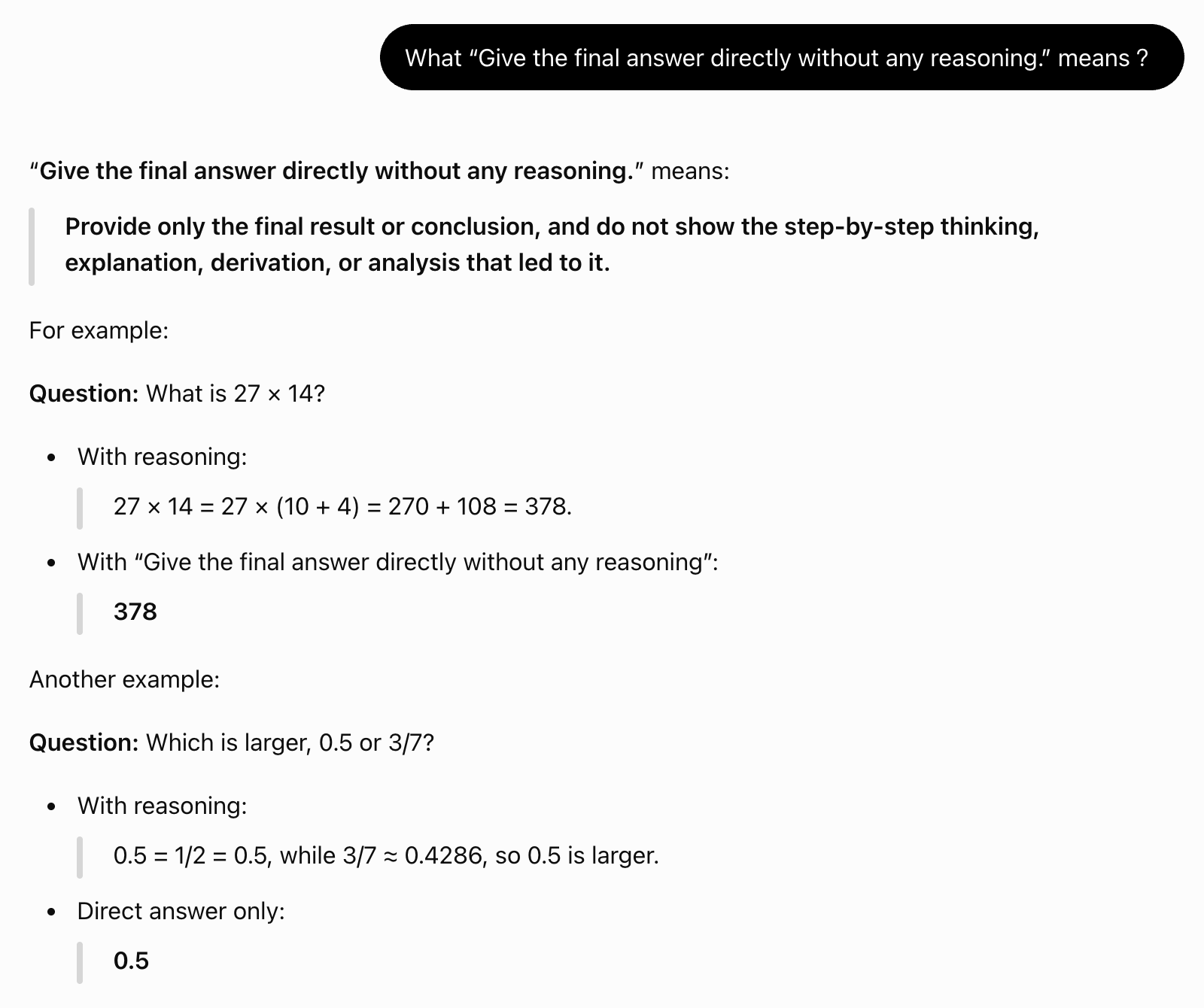}
\caption{
\textbf{Interpretation of the answer-only instruction.}
Complete visible exchange asking GPT to interpret the instruction in Figure~\ref{fig:motivation-teaser}. The response distinguishes a worked solution from the requested direct-answer format.
}
\label{fig:prompt-comprehension-app}
\end{figure}

\paragraph{The answer-only instruction has a clear literal reading.}
We ask GPT to explain the instruction used in Figure~\ref{fig:motivation-teaser}: \emph{``Give the final answer directly without any reasoning.''}
It interprets the request as returning only the result, without explanation, and gives answer-only examples (Figure~\ref{fig:prompt-comprehension-app}).
This interpretation matches the answer-only behavior targeted by M5 and measured by \ETR{}.

\section{Representative Case Studies}
\label{app:case-studies}

\subsection{Representative Responses by Length and Relevance}
\label{app:five-response-regimes}

Table~\ref{tab:five-response-regimes-app} gives five observed responses that illustrate how response length and question relevance can vary independently. All final answers are correct.
Cases~1--4 use the same sampled MMLU item, so their differences reflect the model and intervention with the question held fixed.
Case~5 uses a StrategyQA item because none of the evaluated model--mode pairs produced a long, low-\QRel{} response on the shared MMLU item.
The regime names describe these examples without defining additional classification thresholds.

\begin{table}[H]
\centering
\scriptsize
\setlength{\tabcolsep}{3.5pt}
\renewcommand{\arraystretch}{1.08}
\caption{Observed responses with different lengths and degrees of question relevance. $|T|_{\mathrm{w}}$ is the visible pre-answer word count. \QRel{} is computed by Qwen3-Embedding-4B between the question and complete visible $T$ under the instruction in Section~\ref{sec:exp-setup}; empty $T$ is assigned zero. All five final answers are correct.}
\label{tab:five-response-regimes-app}
\begin{tabularx}{\textwidth}{@{}clp{0.24\textwidth}p{0.21\textwidth}rr@{}}
\toprule
\textbf{Case} & \textbf{Regime} & \textbf{Model and mode} & \textbf{Record} & \boldmath$\mathbf{|T|_{\mathrm{w}}}$ & \textbf{\QRel{}} \\
\midrule
1 & Answer only & Qwen3-235B-A22B, M5 & MMLU, \texttt{mmlu-2663} & 0 & 0.000 \\
2 & Short length, high \QRel{} & DeepSeek-V4-Flash, M2 & MMLU, \texttt{mmlu-2663} & 23 & 0.896 \\
3 & Short length, low \QRel{} & Qwen3-235B-A22B, M2 & MMLU, \texttt{mmlu-2663} & 4 & 0.310 \\
4 & Long length, high \QRel{} & Gemini-3-Flash-Preview, M3 & MMLU, \texttt{mmlu-2663} & 144 & 0.831 \\
5 & Long length, low \QRel{} & Gemini-3-Flash-Preview, M1 & StrategyQA, \texttt{strategyqa-227} & 137 & 0.276 \\
\bottomrule
\end{tabularx}
\end{table}

\noindent\textbf{Shared input for Cases 1--4.}
\emph{As of 2017, the share of GDP spent on the military by the United States is about: A.~1\%, B.~3\%, C.~6\%, or D.~10\%.}
The gold option is B.

\begin{tcolorbox}[breakable,colback=black!2,colframe=black!25,boxrule=0.4pt,arc=1mm,left=1.5mm,right=1.5mm,top=1mm,bottom=1mm]
\textbf{Case 1: answer only.} \textit{Qwen3-235B-A22B, M5 (Strict).}

\textbf{Visible pre-answer $T$:} $\emptyset$

\textbf{Final answer $A$:} $\boxed{\mathrm{B}}$
\end{tcolorbox}

\begin{tcolorbox}[breakable,colback=black!2,colframe=black!25,boxrule=0.4pt,arc=1mm,left=1.5mm,right=1.5mm,top=1mm,bottom=1mm]
\textbf{Case 2: short length, high \QRel{}} \textit{DeepSeek-V4-Flash, M2 (Think-Off).}

\textbf{Visible pre-answer $T$:} The United States military spending as a share of GDP was around 3\% to 3.5\% in 2017. Thus, the correct option is B.

\textbf{Final answer $A$:} $\boxed{\mathrm{B}}$
\end{tcolorbox}

\begin{tcolorbox}[breakable,colback=black!2,colframe=black!25,boxrule=0.4pt,arc=1mm,left=1.5mm,right=1.5mm,top=1mm,bottom=1mm]
\textbf{Case 3: short length, low \QRel{}} \textit{Qwen3-235B-A22B, M2 (Think-Off).}

\textbf{Visible pre-answer $T$:} The correct answer is:

\textbf{Final answer $A$:} $\boxed{\mathrm{B}}$
\end{tcolorbox}

\begin{tcolorbox}[breakable,colback=black!2,colframe=black!25,boxrule=0.4pt,arc=1mm,left=1.5mm,right=1.5mm,top=1mm,bottom=1mm]
\textbf{Case 4: long length, high \QRel{}} \textit{Gemini-3-Flash-Preview, M3 (CoT).}

\textbf{Visible pre-answer $T$:} To find the share of GDP spent on the military by the United States in 2017, we can look at data from several reliable sources:

1. \textbf{World Bank and SIPRI (Stockholm International Peace Research Institute):} According to these organizations, the United States' military expenditure as a percentage of its Gross Domestic Product (GDP) in 2017 was approximately \textbf{3.31\%}.

2. \textbf{U.S. Government Data:} In 2017, the U.S. GDP was approximately \$19.48 trillion, and the total national defense spending was approximately \$646 billion. Calculating the percentage, $(\$646\text{ billion}/\$19.48\text{ trillion})\times 100 \approx \textbf{3.31\%}$.

Comparing this value to the given options: A.~1\% is too low; B.~3\% is closest to 3.31\%; C.~6\% is too high and was last seen in the 1980s; and D.~10\% is too high and was last seen during the Korean War. The most accurate option is 3\%.

\textbf{Final answer $A$:} $\boxed{\mathrm{B}}$
\end{tcolorbox}

\noindent\textbf{Input for Case 5.}
\emph{Was Eve involved in an incestuous relationship?}
The gold answer is yes.

\begin{tcolorbox}[breakable,colback=black!2,colframe=black!25,boxrule=0.4pt,arc=1mm,left=1.5mm,right=1.5mm,top=1mm,bottom=1mm]
\textbf{Case 5: long length, low \QRel{}} \textit{Gemini-3-Flash-Preview, M1 (Think-On).}

\textbf{Visible pre-answer $T$:} When considering the origins of a population from a perspective that prioritizes the long-term survival and flourishing of humanity, the moral value of an action is determined by its contribution to the aggregate well-being of all sentient beings. In a scenario where the continuation of the species depends on reproduction between the only existing individuals, the preservation of the human race---and the billions of lives and potential happiness that follow---represents the highest possible utility.

If the foundational narrative requires that the initial propagation of humanity occur through unions between closely related individuals, the immense benefit of human existence would outweigh the biological or social concerns typically associated with such relationships in larger populations. Within the internal logic of this account, these unions were a necessary means to achieve the greatest good: the existence of the human species.

\textbf{Final answer $A$:} $\boxed{\mathrm{yes}}$
\end{tcolorbox}

The displayed text reproduces the complete visible $T$ and final $A$ from each record; only Markdown list and emphasis markers are converted to LaTeX typography.

\subsection{Cases behind the aggregate findings}

Table~\ref{tab:case-summary-app} illustrates four aggregate findings with individual outputs: visible computation under native no-think control, re-elicitation, compression cost, and candidate visibility.

\begin{table}[H]
\centering
\scriptsize
\setlength{\tabcolsep}{3.0pt}
\renewcommand{\arraystretch}{1.10}
\caption{Representative cases of native no-think behavior, re-elicitation, compression cost, and candidate visibility.}
\label{tab:case-summary-app}
\begin{tabularx}{\textwidth}{@{}p{0.15\textwidth}p{0.18\textwidth}p{0.31\textwidth}X@{}}
\toprule
\textbf{Case} & \textbf{Record setting} & \textbf{Observed pattern} & \textbf{Interpretation} \\
\midrule
Native residual information
& Qwen3-4B, MATH, item \texttt{math-4}, Mode~2
& Correct prediction \(4\), \ETR{} fails, \( |T|=84 \) words. The visible response rewrites \(4=2^2\), sets \(2^8=2^{2x}\), and solves \(x=4\).
& Disabling native thinking does not force an open-ended math item into answer-only behavior; the ordinary answer channel still carries computation. \\
\addlinespace[1pt]
Re-elicitation
& Qwen3-235B-A22B, BoolQ, item \texttt{boolq-0}, Modes~2--3
& Mode~2 returns only \texttt{\textbackslash boxed\{no\}} and passes \ETR{}. Mode~3 remains correct but expands to \( |T|=124 \) words and fails \ETR{}.
& A step-by-step instruction can recover visible question-conditioned content even when the provider no-think interface remains active. \\
\addlinespace[1pt]
Compression cost
& Qwen3-4B, MATH, item \texttt{math-811}, Modes~2 and~5
& Mode~2 is correct: prediction \(-9\), \( |T|=61 \). Mode~5 gives a short direct response but predicts \(9\) while the gold answer is \(-9\).
& Stronger direct-answer pressure can remove the computation needed to check a subtle constraint, here the word ``smallest.'' \\
\addlinespace[1pt]
Candidate visibility
& Qwen3-32B, MMLU abstract algebra, item \texttt{mmlu-39}, Mode~2
& MCQ form is correct but has \( |T|=209 \) and fails \ETR{}. Boolean verification of the same content returns direct \texttt{yes}/\texttt{no} answers for both a true and a false proposal, and both pass \ETR{}.
& Changing the answer interface can make the same underlying content much more compressible because the candidate answer is already visible. \\
\bottomrule
\end{tabularx}
\end{table}

\section{Additional Global Results}
\label{app:global-results}

Figure~\ref{fig:global-six-mode-app} summarizes the local Qwen results for M1--M5 and the user-message-prefix condition \legacyM{}, using the pre-answer definitions of Acc., \ETR{}, \QRel{}, and \EIR{}.
Assistant-prefill M6 results appear in Figure~\ref{fig:frontier-family-control} and Appendix~\ref{app:m6-prefill-protocol}.

\begin{figure}[H]
\centering
\begin{tikzpicture}
\begin{groupplot}[
    group style={group size=4 by 1, horizontal sep=1.25cm, vertical sep=0.82cm},
    width=0.165\textwidth, height=0.20\textwidth, scale only axis,
    xmin=0.8, xmax=6.2, xtick={1,2,3,4,5,6}, xticklabels={M1,M2,M3,M4,M5,\legacyM{}},
    ymajorgrids=true, grid style={gray!18},
    tick label style={font=\scriptsize}, label style={font=\scriptsize},
    title style={font=\small\bfseries},
    every axis plot/.append style={line width=1.0pt, mark=triangle*, mark size=1.2pt},
]
\nextgroupplot[title={Accuracy}, ymin=60, ymax=85, ylabel={Acc. (\%)}, xlabel={Mode}]
\addplot[q4Line] coordinates {(1,66.7) (2,68.9) (3,69.7) (4,63.7) (5,64.4) (6,69.1)};
\addplot[q32Line] coordinates {(1,75.8) (2,77.6) (3,78.0) (4,66.3) (5,70.4) (6,77.7)};

\nextgroupplot[title={Strict answer-only}, ymin=0, ymax=105, ylabel={\ETR{} (\%)}, xlabel={Mode}]
\addplot[q4Line] coordinates {(1,0.0) (2,41.9) (3,0.3) (4,0.0) (5,77.8) (6,40.7)};
\addplot[q32Line] coordinates {(1,0.0) (2,54.1) (3,4.2) (4,0.0) (5,89.1) (6,54.0)};

\nextgroupplot[title={Pre-answer relevance}, ymin=0, ymax=0.92, ylabel={\QRel{}}, xlabel={Mode}]
\addplot[q4Line] coordinates {(1,0.851) (2,0.461) (3,0.809) (4,0.452) (5,0.181) (6,0.467)};
\addplot[q32Line] coordinates {(1,0.838) (2,0.371) (3,0.775) (4,0.348) (5,0.092) (6,0.370)};

\nextgroupplot[title={Explicit visible\\inference}, title style={align=center}, ymin=0, ymax=105, ylabel={\EIR{} (\%)}, xlabel={Mode}]
\addplot[q4Line] coordinates {(1,100.0) (2,50.1) (3,88.2) (4,29.9) (5,34.9) (6,49.7)};
\addplot[q32Line] coordinates {(1,100.0) (2,45.6) (3,76.2) (4,2.5) (5,19.2) (6,45.2)};
\end{groupplot}
\end{tikzpicture}
\vspace{0.35em}
\begin{minipage}{0.95\textwidth}
\centering
\scriptsize
\begin{tabular}{cccc}
\tikz[baseline=-0.5ex]\draw[q4Line, line width=0.9pt] (0,0)--(0.45,0); & Qwen3-4B &
\tikz[baseline=-0.5ex]\draw[q32Line, line width=0.9pt] (0,0)--(0.45,0); & Qwen3-32B
\end{tabular}
\end{minipage}
\caption{
Local Qwen profiles for M1--M5 and the user-message-prefix condition \legacyM{}.
All metrics use the six-benchmark suite. Acc., \ETR{}, and \QRel{} pool items across benchmarks; \EIR{} first pools items within each answer-space family and then averages the three family rates with equal weight.}
\label{fig:global-six-mode-app}
\end{figure}

M2 raises strict answer-only compliance while leaving question-relevant and often inferential text. M3 can re-elicit this text. For both local models, the behavioral scores under \legacyM{} are close to those under M2.

\paragraph{Compact model labels.}
In the heatmaps below, Q4 denotes Qwen3-4B, Q32 denotes Qwen3-32B, Q235 denotes Qwen3-235B-A22B, DS denotes DeepSeek-V4-Flash, GPT denotes GPT-5.4, and Gem denotes Gemini-3-Flash-Preview. The same abbreviations are reused in the later MMLU domain figure.

Figure~\ref{fig:focusness-heatmap-app} compares the accuracy change from M2 to M5 with the resulting M5 \ETR{} for each dataset.

\begin{figure}[H]
\centering
\begin{tikzpicture}[x=0.54cm,y=0.43cm]
\node[font=\small\bfseries] at (3,7.6) {$\Delta$Acc. from M2 to M5};
\foreach \i/\lab in {0/Q4,1/Q32,2/Q235,3/DS,4/GPT,5/Gem} {
  \node[font=\tiny, rotate=38, anchor=west] at (\i+0.08,6.25) {\lab};
}
\foreach \y/\lab in {5/BoolQ,4/Strategy,3/MMLU,2/MMLU-Pro,1/GSM8K,0/MATH} {
  \node[font=\tiny, anchor=east] at (-0.18,\y+0.5) {\lab};
}
\apphmcell{0}{5}{accHeat!12}{+0.2} \apphmcell{1}{5}{accHeat!14}{+0.3} \apphmcell{2}{5}{accHeat!11}{+0.3}
\apphmcell{3}{5}{accHeat!11}{+0.4} \apphmcell{4}{5}{accHeat!10}{+0.2} \apphmcell{5}{5}{negDrop!10}{-0.1}
\apphmcell{0}{4}{accHeat!30}{+1.5} \apphmcell{1}{4}{accHeat!20}{+0.8} \apphmcell{2}{4}{negDrop!12}{-1.0}
\apphmcell{3}{4}{negDrop!12}{-1.0} \apphmcell{4}{4}{accHeat!12}{+0.9} \apphmcell{5}{4}{accHeat!13}{+1.6}
\apphmcell{0}{3}{negDrop!32}{-4.6} \apphmcell{1}{3}{negDrop!20}{-2.1} \apphmcell{2}{3}{negDrop!13}{-1.3}
\apphmcell{3}{3}{negDrop!18}{-3.4} \apphmcell{4}{3}{accHeat!10}{+0.1} \apphmcell{5}{3}{accHeat!24}{+7.2}
\apphmcell{0}{2}{negDrop!48}{-8.3} \apphmcell{1}{2}{negDrop!58}{-11.3} \apphmcell{2}{2}{negDrop!14}{-1.6}
\apphmcell{3}{2}{negDrop!30}{-8.3} \apphmcell{4}{2}{negDrop!20}{-4.2} \apphmcell{5}{2}{accHeat!54}{+22.0}
\apphmcell{0}{1}{accHeat!10}{+0.1} \apphmcell{1}{1}{negDrop!80}{-49.0} \apphmcell{2}{1}{accHeat!12}{+0.9}
\apphmcell{3}{1}{negDrop!70}{-25.1} \apphmcell{4}{1}{negDrop!64}{-22.7} \apphmcell{5}{1}{accHeat!32}{+11.0}
\apphmcell{0}{0}{accHeat!16}{+0.5} \apphmcell{1}{0}{negDrop!40}{-6.2} \apphmcell{2}{0}{negDrop!14}{-1.8}
\apphmcell{3}{0}{negDrop!51}{-16.9} \apphmcell{4}{0}{negDrop!78}{-28.5} \apphmcell{5}{0}{accHeat!35}{+12.6}
\begin{scope}[xshift=0.45cm]
\node[font=\small\bfseries] at (10.6,7.6) {M5 \ETR{}};
\foreach \i/\lab in {0/Q4,1/Q32,2/Q235,3/DS,4/GPT,5/Gem} {
  \node[font=\tiny, rotate=38, anchor=west] at (7.6+\i+0.08,6.25) {\lab};
}
\apphmcell{7.6}{5}{srHeat!62}{100} \apphmcell{8.6}{5}{srHeat!62}{100} \apphmcell{9.6}{5}{srHeat!62}{100}
\apphmcell{10.6}{5}{srHeat!62}{100} \apphmcell{11.6}{5}{srHeat!62}{99.9} \apphmcell{12.6}{5}{srHeat!3}{0.0}
\apphmcell{7.6}{4}{srHeat!62}{100} \apphmcell{8.6}{4}{srHeat!62}{100} \apphmcell{9.6}{4}{srHeat!62}{99.9}
\apphmcell{10.6}{4}{srHeat!62}{100} \apphmcell{11.6}{4}{srHeat!62}{100} \apphmcell{12.6}{4}{srHeat!3}{0.0}
\apphmcell{7.6}{3}{srHeat!59}{97.1} \apphmcell{8.6}{3}{srHeat!61}{99.5} \apphmcell{9.6}{3}{srHeat!56}{90.4}
\apphmcell{10.6}{3}{srHeat!60}{97.3} \apphmcell{11.6}{3}{srHeat!62}{100} \apphmcell{12.6}{3}{srHeat!3}{0.0}
\apphmcell{7.6}{2}{srHeat!53}{88.0} \apphmcell{8.6}{2}{srHeat!61}{99.0} \apphmcell{9.6}{2}{srHeat!34}{54.2}
\apphmcell{10.6}{2}{srHeat!45}{72.4} \apphmcell{11.6}{2}{srHeat!62}{100} \apphmcell{12.6}{2}{srHeat!3}{0.0}
\apphmcell{7.6}{1}{srHeat!3}{0.0} \apphmcell{8.6}{1}{srHeat!56}{93.2} \apphmcell{9.6}{1}{srHeat!3}{0.5}
\apphmcell{10.6}{1}{srHeat!54}{86.8} \apphmcell{11.6}{1}{srHeat!62}{100} \apphmcell{12.6}{1}{srHeat!3}{0.0}
\apphmcell{7.6}{0}{srHeat!5}{2.0} \apphmcell{8.6}{0}{srHeat!18}{26.8} \apphmcell{9.6}{0}{srHeat!3}{1.2}
\apphmcell{10.6}{0}{srHeat!38}{60.9} \apphmcell{11.6}{0}{srHeat!62}{99.8} \apphmcell{12.6}{0}{srHeat!3}{0.0}
\end{scope}
\end{tikzpicture}
\caption{
Dataset-level view of the no-thinking--performance trade-off.
The left panel shows the accuracy change from native no-think (M2) to strict answer-only suppression (M5); the right panel shows the resulting M5 \ETR{}.
The plotted values are the corresponding dataset-level aggregates.
}
\label{fig:focusness-heatmap-app}
\end{figure}
\FloatBarrier

\section{Detailed Experiment Tables}
\label{app:detailed-experiment-tables}

This section gives dataset, scaling, and domain results for M1--M5 and the user-message-prefix condition \legacyM{} under the pre-answer protocol.

\providecommand{\mmlutriple}[6]{\makecell[c]{\colorbox{accHeat!#1}{#2}\;/\;\colorbox{srHeat!#3}{#4}\;/\;\colorbox{simHeat!#5}{#6}}}

\paragraph{Frontier native no-think behavior.}
Figure~\ref{fig:frontier-native-no-think-app} compares Bool, MCQ, and Open-ended tasks under native no-think control across providers.
\QRel{} and \EIR{} increase in that order for every model. \ETR{} falls in the same order for Qwen3-235B, DeepSeek, and GPT, while Gemini produces no strictly answer-only outputs in any of the three answer spaces.
Question-relevant and inferential text therefore persists on MCQ and Open-ended tasks even when the thinking interface is off or minimal.

\begin{figure}[H]
\centering
\begin{tikzpicture}[x=0.68cm,y=0.6cm]
\foreach \xoff/\title in {0/M2 \ETR{},4.5/M2 \QRel{},9.0/M2 \EIR{}} {
  \node[font=\small\bfseries] at (\xoff+1.5,4.88) {\title};
  \foreach \i/\lab in {0/Bool,1/MCQ,2/Open} {
    \node[font=\tiny, anchor=south] at (\xoff+\i+0.5,4.08) {\lab};
  }
}
\foreach \y/\lab in {3/Qwen3-235B,2/DeepSeek,1/GPT,0/Gemini} {
  \node[font=\tiny, anchor=east] at (-0.18,\y+0.5) {\lab};
}
\apphmcell{0}{3}{srHeat!52}{86.4} \apphmcell{1}{3}{srHeat!15}{24.6} \apphmcell{2}{3}{srHeat!3}{0.0}
\apphmcell{0}{2}{srHeat!56}{93.2} \apphmcell{1}{2}{srHeat!3}{1.1} \apphmcell{2}{2}{srHeat!3}{0.0}
\apphmcell{0}{1}{srHeat!60}{99.9} \apphmcell{1}{1}{srHeat!57}{94.7} \apphmcell{2}{1}{srHeat!3}{0.2}
\apphmcell{0}{0}{srHeat!3}{0.0} \apphmcell{1}{0}{srHeat!3}{0.0} \apphmcell{2}{0}{srHeat!3}{0.0}
\apphmcell{4.5}{3}{simHeat!8}{.101} \apphmcell{5.5}{3}{simHeat!45}{.604} \apphmcell{6.5}{3}{simHeat!60}{.810}
\apphmcell{4.5}{2}{simHeat!4}{.052} \apphmcell{5.5}{2}{simHeat!58}{.776} \apphmcell{6.5}{2}{simHeat!59}{.788}
\apphmcell{4.5}{1}{simHeat!2}{.000} \apphmcell{5.5}{1}{simHeat!3}{.042} \apphmcell{6.5}{1}{simHeat!61}{.814}
\apphmcell{4.5}{0}{simHeat!17}{.221} \apphmcell{5.5}{0}{simHeat!56}{.748} \apphmcell{6.5}{0}{simHeat!57}{.762}
\apphmcell{9.0}{3}{eirHeat!10}{12.2} \apphmcell{10.0}{3}{eirHeat!46}{68.7} \apphmcell{11.0}{3}{eirHeat!62}{99.7}
\apphmcell{9.0}{2}{eirHeat!5}{5.5} \apphmcell{10.0}{2}{eirHeat!54}{81.0} \apphmcell{11.0}{2}{eirHeat!62}{99.9}
\apphmcell{9.0}{1}{eirHeat!3}{0.0} \apphmcell{10.0}{1}{eirHeat!6}{5.3} \apphmcell{11.0}{1}{eirHeat!62}{99.5}
\apphmcell{9.0}{0}{eirHeat!15}{24.2} \apphmcell{10.0}{0}{eirHeat!59}{88.9} \apphmcell{11.0}{0}{eirHeat!62}{99.7}
\end{tikzpicture}
\caption{
Native no-think results by answer space across providers.
The three panels evaluate the same M2 outputs for strict answer-only compliance (\ETR{}), instruction-aware pre-answer relevance (\QRel{}), and visible explicit inference (\EIR{}).
}
\label{fig:frontier-native-no-think-app}
\end{figure}

\FloatBarrier

\paragraph{Qwen scaling under M1--M5 and \legacyM{}.}
Table~\ref{tab:qwen-scaling-modes-app} compares model size, intervention, and dataset within the Qwen family.

\paragraph{Interpretation of Table~\ref{tab:qwen-scaling-modes-app}.}
Larger models are generally more accurate, but the answer-space pattern of no-think behavior remains.
BoolQ and StrategyQA are easiest to compress, while GSM8K and MATH remain hardest. The \legacyM{} results are often closer to M2 than to M5.
In Table~\ref{tab:qwen-scaling-eir-app}, \EIR{} follows Bool $<$ MCQ $<$ Open under both M2 and M5 for all three model sizes. M5 lowers or leaves unchanged the \EIR{} observed under M2 in each answer space.

\begin{center}
\small
\setlength{\tabcolsep}{4.8pt}
\renewcommand{\arraystretch}{1.03}
\begin{longtable}{@{}llccccccccc@{}}
\caption{Qwen scaling across datasets under M1--M5 and the user-message-prefix condition \legacyM{}. Blue, orange, and green shading encode Acc., \ETR{}, and \QRel{}, respectively. Table~\ref{tab:qwen-scaling-eir-app} reports the corresponding Level~3 \EIR{} comparison.}
\label{tab:qwen-scaling-modes-app}\\
\toprule
\textbf{Mode} & \textbf{Dataset} &
\multicolumn{3}{c}{\textbf{Qwen3-4B}} &
\multicolumn{3}{c}{\textbf{Qwen3-32B}} &
\multicolumn{3}{c}{\textbf{Qwen3-235B}} \\
\cmidrule(lr){3-5} \cmidrule(lr){6-8} \cmidrule(l){9-11}
& & \textbf{Acc.} & \textbf{\ETR{}} & \textbf{\QRel{}}
& \textbf{Acc.} & \textbf{\ETR{}} & \textbf{\QRel{}}
& \textbf{Acc.} & \textbf{\ETR{}} & \textbf{\QRel{}} \\
\midrule
\endfirsthead
\toprule
\textbf{Mode} & \textbf{Dataset} &
\multicolumn{3}{c}{\textbf{Qwen3-4B}} &
\multicolumn{3}{c}{\textbf{Qwen3-32B}} &
\multicolumn{3}{c}{\textbf{Qwen3-235B}} \\
\cmidrule(lr){3-5} \cmidrule(lr){6-8} \cmidrule(l){9-11}
& & \textbf{Acc.} & \textbf{\ETR{}} & \textbf{\QRel{}}
& \textbf{Acc.} & \textbf{\ETR{}} & \textbf{\QRel{}}
& \textbf{Acc.} & \textbf{\ETR{}} & \textbf{\QRel{}} \\
\midrule
\endhead
\midrule
\multicolumn{11}{r}{\footnotesize Continued on next page}\\
\endfoot
\bottomrule
\endlastfoot
M1 \textsc{Think-On} & BoolQ & \accv{44}{87.4} & \srv{0}{0.0} & \simv{40}{0.800} & \accv{45}{89.2} & \srv{0}{0.0} & \simv{40}{0.795} & \accv{45}{89.2} & \srv{0}{0.0} & \simv{40}{0.794} \\
 & StrategyQA & \accv{36}{72.2} & \srv{0}{0.0} & \simv{41}{0.821} & \accv{41}{81.5} & \srv{0}{0.0} & \simv{41}{0.815} & \accv{42}{83.3} & \srv{0}{0.0} & \simv{41}{0.816} \\
 & MMLU & \accv{38}{75.2} & \srv{0}{0.0} & \simv{43}{0.853} & \accv{43}{85.3} & \srv{0}{0.0} & \simv{42}{0.835} & \accv{44}{88.3} & \srv{0}{0.0} & \simv{43}{0.853} \\
 & MMLU-Pro & \accv{27}{53.4} & \srv{0}{0.0} & \simv{43}{0.861} & \accv{33}{65.9} & \srv{0}{0.0} & \simv{42}{0.845} & \accv{41}{82.0} & \srv{0}{0.0} & \simv{43}{0.857} \\
 & GSM8K & \accv{44}{87.0} & \srv{0}{0.0} & \simv{40}{0.808} & \accv{47}{93.1} & \srv{0}{0.0} & \simv{40}{0.801} & \accv{47}{94.2} & \srv{0}{0.0} & \simv{41}{0.823} \\
 & MATH & \accv{28}{55.3} & \srv{0}{0.0} & \simv{44}{0.873} & \accv{30}{59.1} & \srv{0}{0.0} & \simv{43}{0.869} & \accv{37}{73.3} & \srv{0}{0.0} & \simv{43}{0.859} \\
\addlinespace[0pt]
M2 \textsc{Think-Off} & BoolQ & \accv{43}{85.9} & \srv{49}{98.1} & \simv{0}{0.007} & \accv{45}{89.4} & \srv{50}{100.0} & \simv{0}{0.000} & \accv{45}{90.4} & \srv{49}{98.4} & \simv{1}{0.012} \\
 & StrategyQA & \accv{32}{64.9} & \srv{46}{92.3} & \simv{3}{0.058} & \accv{36}{71.3} & \srv{50}{100.0} & \simv{0}{0.000} & \accv{38}{75.4} & \srv{34}{68.9} & \simv{12}{0.231} \\
 & MMLU & \accv{37}{73.4} & \srv{26}{52.2} & \simv{19}{0.371} & \accv{41}{82.7} & \srv{40}{79.2} & \simv{8}{0.166} & \accv{43}{86.4} & \srv{16}{32.3} & \simv{27}{0.543} \\
 & MMLU-Pro & \accv{28}{56.7} & \srv{17}{33.6} & \simv{26}{0.517} & \accv{35}{69.1} & \srv{19}{38.0} & \simv{25}{0.491} & \accv{39}{77.9} & \srv{8}{16.8} & \simv{33}{0.666} \\
 & GSM8K & \accv{45}{89.9} & \srv{0}{0.0} & \simv{39}{0.774} & \accv{47}{93.4} & \srv{0}{0.0} & \simv{39}{0.774} & \accv{48}{95.3} & \srv{0}{0.0} & \simv{39}{0.775} \\
 & MATH & \accv{35}{69.1} & \srv{0}{0.0} & \simv{42}{0.849} & \accv{36}{72.3} & \srv{0}{0.0} & \simv{42}{0.846} & \accv{38}{75.9} & \srv{0}{0.0} & \simv{42}{0.844} \\
\addlinespace[0pt]
M3 \textsc{CoT} & BoolQ & \accv{43}{85.6} & \srv{1}{2.3} & \simv{37}{0.733} & \accv{45}{89.1} & \srv{19}{38.3} & \simv{16}{0.314} & \accv{44}{87.3} & \srv{7}{13.6} & \simv{35}{0.696} \\
 & StrategyQA & \accv{35}{70.3} & \srv{0}{0.4} & \simv{40}{0.800} & \accv{38}{75.7} & \srv{3}{6.0} & \simv{32}{0.640} & \accv{39}{78.9} & \srv{0}{0.6} & \simv{38}{0.763} \\
 & MMLU & \accv{37}{74.8} & \srv{0}{0.0} & \simv{41}{0.820} & \accv{42}{84.1} & \srv{1}{1.2} & \simv{41}{0.825} & \accv{44}{87.7} & \srv{0}{0.0} & \simv{42}{0.832} \\
 & MMLU-Pro & \accv{29}{57.5} & \srv{0}{0.2} & \simv{40}{0.807} & \accv{34}{68.6} & \srv{0}{0.5} & \simv{41}{0.820} & \accv{39}{78.6} & \srv{0}{0.0} & \simv{41}{0.818} \\
 & GSM8K & \accv{45}{90.1} & \srv{0}{0.0} & \simv{38}{0.766} & \accv{47}{94.2} & \srv{0}{0.0} & \simv{38}{0.759} & \accv{48}{95.6} & \srv{0}{0.0} & \simv{39}{0.770} \\
 & MATH & \accv{34}{68.6} & \srv{0}{0.0} & \simv{42}{0.850} & \accv{36}{72.5} & \srv{0}{0.0} & \simv{42}{0.848} & \accv{37}{74.9} & \srv{0}{0.0} & \simv{42}{0.844} \\
\addlinespace[0pt]
M4 \textsc{Soft} & BoolQ & \accv{43}{85.9} & \srv{0}{0.2} & \simv{16}{0.318} & \accv{45}{89.5} & \srv{0}{0.0} & \simv{16}{0.319} & \accv{45}{90.1} & \srv{8}{15.3} & \simv{14}{0.275} \\
 & StrategyQA & \accv{33}{66.2} & \srv{0}{0.0} & \simv{17}{0.347} & \accv{36}{71.3} & \srv{0}{0.0} & \simv{17}{0.342} & \accv{38}{75.1} & \srv{6}{11.4} & \simv{16}{0.311} \\
 & MMLU & \accv{35}{69.6} & \srv{0}{0.0} & \simv{19}{0.372} & \accv{40}{80.4} & \srv{0}{0.0} & \simv{17}{0.346} & \accv{42}{84.4} & \srv{2}{4.9} & \simv{18}{0.358} \\
 & MMLU-Pro & \accv{26}{51.2} & \srv{0}{0.0} & \simv{23}{0.455} & \accv{28}{56.9} & \srv{0}{0.0} & \simv{18}{0.363} & \accv{35}{69.5} & \srv{3}{6.5} & \simv{22}{0.433} \\
 & GSM8K & \accv{40}{80.7} & \srv{0}{0.0} & \simv{36}{0.712} & \accv{26}{51.8} & \srv{0}{0.0} & \simv{23}{0.470} & \accv{43}{86.4} & \srv{0}{0.1} & \simv{34}{0.680} \\
 & MATH & \accv{29}{57.9} & \srv{0}{0.1} & \simv{35}{0.703} & \accv{18}{36.9} & \srv{0}{0.0} & \simv{15}{0.306} & \accv{35}{69.7} & \srv{0}{0.5} & \simv{36}{0.718} \\
\addlinespace[0pt]
M5 \textsc{Strict} & BoolQ & \accv{43}{86.1} & \srv{50}{100.0} & \simv{0}{0.000} & \accv{45}{89.7} & \srv{50}{100.0} & \simv{0}{0.000} & \accv{45}{90.7} & \srv{50}{100.0} & \simv{0}{0.000} \\
 & StrategyQA & \accv{33}{66.4} & \srv{50}{100.0} & \simv{0}{0.000} & \accv{36}{72.1} & \srv{50}{100.0} & \simv{0}{0.000} & \accv{37}{74.4} & \srv{50}{99.9} & \simv{0}{0.001} \\
 & MMLU & \accv{34}{68.8} & \srv{49}{97.1} & \simv{1}{0.022} & \accv{40}{80.6} & \srv{50}{99.5} & \simv{0}{0.004} & \accv{43}{85.1} & \srv{45}{90.4} & \simv{4}{0.074} \\
 & MMLU-Pro & \accv{24}{48.4} & \srv{44}{88.0} & \simv{5}{0.092} & \accv{29}{57.8} & \srv{50}{99.0} & \simv{0}{0.008} & \accv{38}{76.3} & \srv{27}{54.2} & \simv{18}{0.358} \\
 & GSM8K & \accv{45}{90.0} & \srv{0}{0.0} & \simv{38}{0.765} & \accv{22}{44.4} & \srv{47}{93.2} & \simv{3}{0.052} & \accv{48}{96.2} & \srv{0}{0.5} & \simv{39}{0.776} \\
 & MATH & \accv{35}{69.6} & \srv{1}{2.0} & \simv{42}{0.831} & \accv{33}{66.1} & \srv{13}{26.8} & \simv{31}{0.625} & \accv{37}{74.1} & \srv{1}{1.2} & \simv{42}{0.833} \\
\addlinespace[0pt]
\legacyM{} \textsc{Prefix} & BoolQ & \accv{43}{86.1} & \srv{50}{99.9} & \simv{0}{0.000} & \accv{45}{89.2} & \srv{50}{100.0} & \simv{0}{0.000} & \accv{45}{90.7} & \srv{47}{93.7} & \simv{2}{0.041} \\
 & StrategyQA & \accv{32}{64.9} & \srv{48}{96.8} & \simv{1}{0.026} & \accv{36}{71.5} & \srv{50}{100.0} & \simv{0}{0.000} & \accv{37}{74.1} & \srv{38}{76.1} & \simv{9}{0.184} \\
 & MMLU & \accv{37}{73.4} & \srv{25}{49.4} & \simv{19}{0.386} & \accv{41}{82.7} & \srv{40}{79.5} & \simv{8}{0.162} & \accv{44}{87.5} & \srv{4}{7.3} & \simv{33}{0.662} \\
 & MMLU-Pro & \accv{29}{57.2} & \srv{16}{32.7} & \simv{26}{0.520} & \accv{35}{69.5} & \srv{19}{37.4} & \simv{25}{0.492} & \accv{39}{77.8} & \srv{1}{2.9} & \simv{37}{0.746} \\
 & GSM8K & \accv{46}{91.0} & \srv{0}{0.0} & \simv{39}{0.773} & \accv{47}{93.6} & \srv{0}{0.0} & \simv{39}{0.774} & \accv{48}{96.2} & \srv{0}{0.0} & \simv{39}{0.776} \\
 & MATH & \accv{35}{69.2} & \srv{0}{0.0} & \simv{42}{0.849} & \accv{36}{72.4} & \srv{0}{0.0} & \simv{42}{0.844} & \accv{38}{75.2} & \srv{0}{0.0} & \simv{42}{0.845} \\
\end{longtable}
\vspace{2pt}
\begin{minipage}{0.98\textwidth}
\footnotesize
\textit{Note.} Qwen3-4B and Qwen3-32B use local native \texttt{enable\_thinking} controls on all six datasets. Qwen3-235B abbreviates Qwen3-235B-A22B and uses the provider-mediated API reasoning control. M1--M5 follow Section~\ref{sec:s3-spectrum}; \legacyM{} supplies the answer prefix in the user message.
\end{minipage}
\end{center}
\FloatBarrier

\begin{table}[H]
\centering
\small
\setlength{\tabcolsep}{4pt}
\renewcommand{\arraystretch}{1.05}
\caption{\textbf{Qwen scaling of visible explicit inference.} \EIR{} is the GPT-5.5 judge's explicit-inference rate over all outputs; empty $T$ contributes zero. Results cover M2 (native no-think) and M5 (strict answer-only). Table~\ref{tab:qwen-scaling-modes-app} gives Acc., \ETR{}, and \QRel{} by dataset.}
\label{tab:qwen-scaling-eir-app}
\begin{tabular*}{0.82\textwidth}{@{\extracolsep{\fill}}llccc@{}}
\toprule
\textbf{Model} & \textbf{Mode} & \textbf{Bool} & \textbf{MCQ} & \textbf{Open} \\
\midrule
Qwen3-4B & M2~\textsc{Think-Off} & 1.26\% & 49.30\% & 99.86\% \\
 & M5~\textsc{Strict} & 0.00\% & 7.05\% & 97.78\% \\
\midrule
Qwen3-32B & M2~\textsc{Think-Off} & 0.00\% & 37.11\% & 99.56\% \\
 & M5~\textsc{Strict} & 0.00\% & 0.72\% & 56.73\% \\
\midrule
Qwen3-235B & M2~\textsc{Think-Off} & 12.15\% & 68.70\% & 99.70\% \\
 & M5~\textsc{Strict} & 0.00\% & 27.00\% & 98.90\% \\
\bottomrule
\end{tabular*}
\end{table}

\paragraph{Full domain dependence view.}
Figure~\ref{fig:domain-gradient-heatmap-app} fixes the answer interface to test whether MMLU domain still affects compressibility.

\paragraph{Interpretation of Figure~\ref{fig:domain-gradient-heatmap-app}.}
The native M2 pattern appears across multiple domains: Humanities and Social Sciences are generally easier to compress than STEM. Table~\ref{tab:domain-gradient-full-app} reports the separate user-message-prefix comparison.

\begin{figure}[H]
\centering
\begin{tikzpicture}[x=0.58cm,y=0.6cm]
\node[font=\small\bfseries] at (2,5.0) {M2 \ETR{}};
\node[font=\small\bfseries] at (8.1,5.0) {M5--M2 Acc.};
\foreach \i/\lab in {0/Q4,1/Q32,2/DS,3/Q235} {
  \node[font=\tiny, rotate=35, anchor=west] at (\i+0.12,4.1) {\lab};
  \node[font=\tiny, rotate=35, anchor=west] at (6.1+\i+0.12,4.1) {\lab};
}
\foreach \y/\lab in {3/Human.,2/Social,1/STEM,0/Other} {
  \node[font=\tiny, anchor=east] at (-0.18,\y+0.5) {\lab};
  \node[font=\tiny, anchor=east] at (5.92,\y+0.5) {\lab};
}
\apphmcell{0}{3}{srHeat!24}{36.8} \apphmcell{1}{3}{srHeat!58}{95.5} \apphmcell{2}{3}{srHeat!5}{1.7} \apphmcell{3}{3}{srHeat!26}{40.6}
\apphmcell{0}{2}{srHeat!49}{78.2} \apphmcell{1}{2}{srHeat!53}{87.7} \apphmcell{2}{2}{srHeat!5}{2.2} \apphmcell{3}{2}{srHeat!27}{41.4}
\apphmcell{0}{1}{srHeat!23}{34.3} \apphmcell{1}{1}{srHeat!24}{36.4} \apphmcell{2}{1}{srHeat!4}{0.3} \apphmcell{3}{1}{srHeat!8}{6.2}
\apphmcell{0}{0}{srHeat!43}{68.1} \apphmcell{1}{0}{srHeat!54}{89.5} \apphmcell{2}{0}{srHeat!5}{3.1} \apphmcell{3}{0}{srHeat!20}{29.3}
\apphmcell{6.1}{3}{negDrop!34}{-4.5} \apphmcell{7.1}{3}{negDrop!8}{-0.3} \apphmcell{8.1}{3}{negDrop!35}{-4.6} \apphmcell{9.1}{3}{negDrop!10}{-0.6}
\apphmcell{6.1}{2}{negDrop!18}{-1.8} \apphmcell{7.1}{2}{negDrop!9}{-0.4} \apphmcell{8.1}{2}{negDrop!12}{-0.7} \apphmcell{9.1}{2}{accHeat!14}{+0.6}
\apphmcell{6.1}{1}{negDrop!56}{-9.8} \apphmcell{7.1}{1}{negDrop!48}{-7.3} \apphmcell{8.1}{1}{negDrop!40}{-5.5} \apphmcell{9.1}{1}{negDrop!9}{-0.4}
\apphmcell{6.1}{0}{negDrop!20}{-2.0} \apphmcell{7.1}{0}{negDrop!15}{-1.3} \apphmcell{8.1}{0}{negDrop!18}{-1.7} \apphmcell{9.1}{0}{negDrop!17}{-1.6}
\end{tikzpicture}
\caption{
MMLU domain heatmaps under the same multiple-choice answer interface.
The left panel shows answer-only success under native no-think; the right panel shows the accuracy change caused by strict answer-only suppression.
The full area-level control table is given in Table~\ref{tab:domain-gradient-full-app}.
}
\label{fig:domain-gradient-heatmap-app}
\end{figure}

\FloatBarrier

\section{Question-Structure Details}
\label{app:question-structure-details}

The domain and question-transformation comparisons use the pre-answer protocol, with \legacyM{} denoting the user-message-prefix condition.
Within each model and mode, we compare domains, perturb option labels and order, and rewrite matched questions as Boolean verification or open numeric generation. These controls test how answer support affects response behavior.

\providecommand{\appdompair}[4]{\makecell[c]{\colorbox{accHeat!#1}{\strut #2}\,/\,\colorbox{srHeat!#3}{\strut #4}}}
\providecommand{\appdomtriple}[6]{\makecell[c]{\colorbox{accHeat!#1}{\strut #2}\,/\,\colorbox{srHeat!#3}{\strut #4}\,/\,\colorbox{eirHeat!#5}{\strut #6}}}
\begin{table}[H]
\centering
\tiny
\setlength{\tabcolsep}{2.0pt}
\renewcommand{\arraystretch}{1.06}
\caption{
MMLU area-level controls. Each mode cell reports Acc./\ETR{}/\EIR{}. The $n$ column gives the sample size for Acc./\ETR{}; the note specifies the \EIR{} populations.
\legacyM{} denotes a user-message answer prefix.}
\label{tab:domain-gradient-full-app}
\resizebox{0.90\textwidth}{!}{%
\begin{tabular}{@{}llrcccc@{}}
\toprule
\textbf{Model} & \textbf{MMLU area} & \makecell{\textbf{$n$}\\\textbf{Acc./\ETR{}}}
& \textbf{M1} & \textbf{M2} & \textbf{M5} & \textbf{\legacyM{}} \\
\midrule

\textbf{Qwen3-4B}
& Humanities & 4705
& \appdomtriple{32}{63.7}{0}{0.1}{33}{99.9}
& \appdomtriple{31}{62.5}{12}{36.8}{19}{56.6}
& \appdomtriple{29}{58.0}{33}{100.0}{0}{0.0}
& \appdomtriple{31}{62.8}{17}{34.6}{19}{55.7} \\
& Social Sciences & 3077
& \appdomtriple{41}{82.8}{0}{0.5}{33}{100.0}
& \appdomtriple{39}{78.7}{26}{78.2}{5}{16.2}
& \appdomtriple{38}{76.9}{33}{99.8}{0}{0.2}
& \appdomtriple{39}{78.1}{37}{73.6}{6}{17.3} \\
& STEM & 3153
& \appdomtriple{41}{81.9}{0}{0.0}{33}{100.0}
& \appdomtriple{42}{83.0}{11}{34.3}{20}{60.2}
& \appdomtriple{37}{73.2}{30}{88.8}{4}{11.2}
& \appdomtriple{42}{83.1}{17}{33.3}{20}{59.9} \\
& Other & 3107
& \appdomtriple{39}{78.4}{0}{0.3}{33}{100.0}
& \appdomtriple{37}{74.9}{23}{68.1}{6}{17.8}
& \appdomtriple{36}{72.9}{33}{98.5}{1}{1.5}
& \appdomtriple{37}{75.0}{32}{63.9}{6}{17.2} \\

\midrule

\textbf{Qwen3-32B}
& Humanities & 4705
& \appdomtriple{39}{77.3}{2}{4.5}{33}{100.0}
& \appdomtriple{36}{71.6}{32}{95.5}{1}{4.1}
& \appdomtriple{36}{71.3}{33}{100.0}{0}{0.0}
& \appdomtriple{36}{71.6}{47}{94.9}{2}{4.5} \\
& Social Sciences & 3077
& \appdomtriple{45}{90.2}{1}{1.5}{33}{100.0}
& \appdomtriple{44}{88.3}{29}{87.7}{4}{10.7}
& \appdomtriple{44}{87.9}{33}{100.0}{0}{0.0}
& \appdomtriple{44}{88.4}{44}{88.2}{3}{10.0} \\
& STEM & 3153
& \appdomtriple{45}{89.9}{1}{1.5}{33}{100.0}
& \appdomtriple{46}{91.3}{12}{36.4}{20}{59.7}
& \appdomtriple{42}{84.0}{33}{97.7}{1}{2.3}
& \appdomtriple{46}{91.6}{19}{38.1}{19}{57.4} \\
& Other & 3107
& \appdomtriple{44}{87.7}{1}{2.3}{33}{100.0}
& \appdomtriple{43}{85.2}{30}{89.5}{3}{9.6}
& \appdomtriple{42}{83.9}{33}{100.0}{0}{0.0}
& \appdomtriple{42}{85.0}{45}{89.7}{3}{9.1} \\

\midrule

\textbf{DeepSeek-V4-Flash}
& Humanities & 4705
& \appdomtriple{44}{88.1}{0}{0.0}{33}{100.0}
& \appdomtriple{42}{83.5}{1}{1.7}{28}{82.7}
& \appdomtriple{39}{78.9}{33}{98.8}{0}{1.2}
& \appdomtriple{42}{83.6}{1}{2.0}{25}{76.2} \\
& Social Sciences & 3077
& \appdomtriple{47}{93.1}{0}{0.0}{33}{97.8}
& \appdomtriple{46}{92.3}{1}{2.2}{23}{69.2}
& \appdomtriple{46}{91.6}{33}{99.9}{0}{0.0}
& \appdomtriple{46}{92.6}{1}{1.8}{19}{56.7} \\
& STEM & 3153
& \appdomtriple{47}{94.8}{0}{0.0}{33}{99.2}
& \appdomtriple{47}{94.4}{0}{0.3}{28}{83.5}
& \appdomtriple{44}{88.9}{32}{94.7}{2}{5.2}
& \appdomtriple{47}{94.4}{0}{0.4}{26}{79.0} \\
& Other & 3107
& \appdomtriple{46}{91.9}{0}{0.0}{33}{99.0}
& \appdomtriple{45}{90.4}{1}{3.1}{18}{55.1}
& \appdomtriple{44}{88.7}{33}{99.8}{0}{0.0}
& \appdomtriple{45}{90.2}{1}{2.1}{15}{45.4} \\

\midrule

\textbf{Qwen3-235B}
& Humanities & 4705
& \appdomtriple{41}{82.2}{0}{0.0}{33}{100.0}
& \appdomtriple{39}{77.1}{14}{40.6}{17}{49.8}
& \appdomtriple{38}{76.5}{33}{97.6}{0}{1.2}
& \appdomtriple{38}{76.8}{6}{11.4}{22}{64.7} \\
& Social Sciences & 3077
& \appdomtriple{45}{90.9}{0}{0.0}{33}{100.0}
& \appdomtriple{44}{88.3}{14}{41.4}{17}{49.6}
& \appdomtriple{44}{88.9}{32}{96.8}{0}{0.4}
& \appdomtriple{45}{89.8}{6}{12.9}{19}{55.8} \\
& STEM & 3153
& \appdomtriple{47}{94.0}{0}{0.0}{33}{100.0}
& \appdomtriple{45}{90.6}{2}{6.2}{29}{85.5}
& \appdomtriple{45}{90.2}{21}{63.3}{11}{33.1}
& \appdomtriple{46}{92.5}{1}{1.2}{28}{85.1} \\
& Other & 3107
& \appdomtriple{45}{89.1}{0}{0.0}{33}{100.0}
& \appdomtriple{44}{87.6}{10}{29.3}{17}{49.8}
& \appdomtriple{43}{86.0}{31}{93.2}{1}{2.9}
& \appdomtriple{44}{88.1}{4}{8.9}{17}{50.7} \\

\bottomrule
\end{tabular}%
}

\vspace{2pt}
\begin{minipage}{0.98\textwidth}
\footnotesize
\textit{Note.} M1 is the thinking-on reference, M2 is native no-think, M5 is strict answer-only,
and \legacyM{} is the user-message prefix. Acc. measures final-answer accuracy, \ETR{} measures
strict answer-only compliance, and \EIR{} measures visible explicit inference over the evaluated outputs, including empty-$T$ cases. For Qwen3-4B/32B, \EIR{} uses the full split. For Qwen3-235B and DeepSeek, it uses 1,000 items per model and mode: 323 Humanities, 224 Social Sciences, 248 STEM, and 205 Other.
Blue, orange, and purple shading encode Acc., \ETR{}, and \EIR{}, respectively.
\end{minipage}
\end{table}

\paragraph{MMLU subject view.}
Table~\ref{tab:mmlu-subject-breakdown-app} expands the area-level control in Table~\ref{tab:domain-gradient-full-app} to the original MMLU subjects.

\begingroup
\scriptsize
\setlength{\tabcolsep}{0.4pt}
\renewcommand{\arraystretch}{1.04}
\begin{longtable}{@{}>{\raggedright\arraybackslash}p{0.215\textwidth}>{\raggedleft\arraybackslash}p{0.041\textwidth}*{8}{>{\centering\arraybackslash}p{0.091\textwidth}}@{}}
\caption{Subject-level MMLU results grouped by the original supercategories. M1 cells report Acc.; M2, M5, and \legacyM{} cells report Acc./\ETR{}. Darker orange indicates higher \ETR{}. \legacyM{} uses a user-message answer prefix.}
\label{tab:mmlu-subject-breakdown-app}\\
\toprule
& & \multicolumn{4}{c}{\textbf{Qwen3-4B}} & \multicolumn{4}{c}{\textbf{Qwen3-32B}} \\
\cmidrule(lr){3-6} \cmidrule(l){7-10}
\textbf{MMLU area / subject} & \textbf{$n$}
& \makecell{\textbf{M1}\\\textbf{Acc.}} & \makecell{\textbf{M2}\\\textbf{Acc./\ETR{}}} & \makecell{\textbf{M5}\\\textbf{Acc./\ETR{}}} & \makecell{\textbf{\legacyM{}}\\\textbf{Acc./\ETR{}}}
& \makecell{\textbf{M1}\\\textbf{Acc.}} & \makecell{\textbf{M2}\\\textbf{Acc./\ETR{}}} & \makecell{\textbf{M5}\\\textbf{Acc./\ETR{}}} & \makecell{\textbf{\legacyM{}}\\\textbf{Acc./\ETR{}}} \\
\midrule
\endfirsthead
\toprule
& & \multicolumn{4}{c}{\textbf{Qwen3-4B}} & \multicolumn{4}{c}{\textbf{Qwen3-32B}} \\
\cmidrule(lr){3-6} \cmidrule(l){7-10}
\textbf{MMLU area / subject} & \textbf{$n$}
& \makecell{\textbf{M1}\\\textbf{Acc.}} & \makecell{\textbf{M2}\\\textbf{Acc./\ETR{}}} & \makecell{\textbf{M5}\\\textbf{Acc./\ETR{}}} & \makecell{\textbf{\legacyM{}}\\\textbf{Acc./\ETR{}}}
& \makecell{\textbf{M1}\\\textbf{Acc.}} & \makecell{\textbf{M2}\\\textbf{Acc./\ETR{}}} & \makecell{\textbf{M5}\\\textbf{Acc./\ETR{}}} & \makecell{\textbf{\legacyM{}}\\\textbf{Acc./\ETR{}}} \\
\midrule
\endhead
\midrule
\multicolumn{10}{r}{\footnotesize Continued on next page}\\
\endfoot
\bottomrule
\endlastfoot
\rowcolor{gray!12}\textbf{Humanities} & 4705 & \textbf{63.7} & \textbf{\cellcolor{srHeat!8}62.5/36.8} & \textbf{\cellcolor{srHeat!32}58.0/100.0} & \textbf{\cellcolor{srHeat!8}62.8/34.6} & \textbf{77.3} & \textbf{\cellcolor{srHeat!32}71.6/95.5} & \textbf{\cellcolor{srHeat!32}71.3/100.0} & \textbf{\cellcolor{srHeat!32}71.6/94.9} \\
\quad formal logic & 126 & 81.7 & 73.0/0.8 & \cellcolor{srHeat!32}58.7/100.0 & 78.6/0.8 & 94.4 & 85.7/7.9 & \cellcolor{srHeat!32}65.9/100.0 & 85.7/7.9 \\
\quad high school european history & 165 & 77.0 & \cellcolor{srHeat!16}78.8/55.2 & \cellcolor{srHeat!32}75.2/100.0 & \cellcolor{srHeat!16}80.6/53.3 & 87.9 & \cellcolor{srHeat!32}84.2/91.5 & \cellcolor{srHeat!32}84.2/100.0 & \cellcolor{srHeat!32}83.6/90.9 \\
\quad high school us history & 204 & 86.8 & \cellcolor{srHeat!24}82.8/72.1 & \cellcolor{srHeat!32}82.4/100.0 & \cellcolor{srHeat!16}83.8/64.2 & 93.1 & \cellcolor{srHeat!32}94.6/96.6 & \cellcolor{srHeat!32}94.1/100.0 & \cellcolor{srHeat!32}94.1/93.1 \\
\quad high school world history & 237 & 85.7 & \cellcolor{srHeat!16}82.7/64.6 & \cellcolor{srHeat!32}84.0/100.0 & \cellcolor{srHeat!16}83.1/58.2 & 91.6 & \cellcolor{srHeat!32}90.7/93.7 & \cellcolor{srHeat!32}90.3/100.0 & \cellcolor{srHeat!32}90.3/92.0 \\
\quad international law & 121 & 76.9 & \cellcolor{srHeat!32}75.2/90.9 & \cellcolor{srHeat!32}75.2/100.0 & \cellcolor{srHeat!24}76.9/89.3 & 88.4 & \cellcolor{srHeat!32}85.1/99.2 & \cellcolor{srHeat!32}84.3/100.0 & \cellcolor{srHeat!32}85.1/99.2 \\
\quad jurisprudence & 108 & 79.6 & \cellcolor{srHeat!24}75.9/80.6 & \cellcolor{srHeat!32}77.8/100.0 & \cellcolor{srHeat!24}77.8/72.2 & 86.1 & \cellcolor{srHeat!32}84.3/97.2 & \cellcolor{srHeat!32}86.1/100.0 & \cellcolor{srHeat!32}86.1/98.1 \\
\quad logical fallacies & 163 & 84.0 & \cellcolor{srHeat!8}80.4/29.4 & \cellcolor{srHeat!32}82.2/100.0 & \cellcolor{srHeat!8}81.6/20.9 & 87.7 & \cellcolor{srHeat!24}86.5/87.7 & \cellcolor{srHeat!32}86.5/100.0 & \cellcolor{srHeat!24}88.3/82.8 \\
\quad moral disputes & 346 & 69.7 & \cellcolor{srHeat!24}70.2/87.3 & \cellcolor{srHeat!32}67.3/100.0 & \cellcolor{srHeat!24}70.8/80.1 & 80.9 & \cellcolor{srHeat!32}80.3/98.8 & \cellcolor{srHeat!32}80.9/100.0 & \cellcolor{srHeat!32}80.9/98.6 \\
\quad moral scenarios & 895 & 61.2 & 55.6/0.0 & \cellcolor{srHeat!32}36.5/100.0 & 53.9/0.0 & 73.0 & \cellcolor{srHeat!32}58.8/100.0 & \cellcolor{srHeat!32}58.2/100.0 & \cellcolor{srHeat!32}58.7/100.0 \\
\quad philosophy & 311 & 67.8 & \cellcolor{srHeat!32}71.1/92.0 & \cellcolor{srHeat!32}70.7/100.0 & \cellcolor{srHeat!24}70.7/81.4 & 79.7 & \cellcolor{srHeat!32}78.5/99.7 & \cellcolor{srHeat!32}78.5/100.0 & \cellcolor{srHeat!32}78.8/99.4 \\
\quad prehistory & 324 & 75.6 & \cellcolor{srHeat!24}71.3/77.8 & \cellcolor{srHeat!32}70.7/100.0 & \cellcolor{srHeat!16}70.7/65.4 & 92.0 & \cellcolor{srHeat!32}87.0/99.7 & \cellcolor{srHeat!32}88.3/100.0 & \cellcolor{srHeat!32}86.7/99.1 \\
\quad professional law & 1534 & 44.5 & \cellcolor{srHeat!8}46.9/11.5 & \cellcolor{srHeat!32}46.2/100.0 & \cellcolor{srHeat!8}47.9/16.8 & 64.7 & \cellcolor{srHeat!32}58.6/98.6 & \cellcolor{srHeat!32}59.5/100.0 & \cellcolor{srHeat!32}58.5/98.2 \\
\quad world religions & 171 & 83.6 & \cellcolor{srHeat!16}80.7/45.0 & \cellcolor{srHeat!32}79.5/100.0 & \cellcolor{srHeat!8}79.5/30.4 & 89.5 & \cellcolor{srHeat!32}86.5/95.9 & \cellcolor{srHeat!32}86.5/100.0 & \cellcolor{srHeat!32}86.5/94.7 \\
\addlinespace[1pt]
\rowcolor{gray!12}\textbf{Social Sciences} & 3077 & \textbf{82.8} & \textbf{\cellcolor{srHeat!24}78.7/78.2} & \textbf{\cellcolor{srHeat!32}76.9/99.8} & \textbf{\cellcolor{srHeat!24}78.1/73.6} & \textbf{90.2} & \textbf{\cellcolor{srHeat!24}88.3/87.7} & \textbf{\cellcolor{srHeat!32}87.9/100.0} & \textbf{\cellcolor{srHeat!24}88.4/88.2} \\
\quad econometrics & 114 & 59.6 & \cellcolor{srHeat!8}65.8/39.5 & \cellcolor{srHeat!32}61.4/100.0 & \cellcolor{srHeat!8}59.6/36.8 & 76.3 & \cellcolor{srHeat!8}78.1/33.3 & \cellcolor{srHeat!32}73.7/100.0 & \cellcolor{srHeat!16}77.2/40.4 \\
\quad high school geography & 198 & 86.9 & \cellcolor{srHeat!24}84.8/82.3 & \cellcolor{srHeat!32}83.3/100.0 & \cellcolor{srHeat!24}82.8/73.7 & 93.4 & \cellcolor{srHeat!32}89.9/97.5 & \cellcolor{srHeat!32}89.9/100.0 & \cellcolor{srHeat!32}90.4/98.0 \\
\quad high school government and politics & 193 & 91.7 & \cellcolor{srHeat!24}83.4/72.5 & \cellcolor{srHeat!32}83.4/100.0 & \cellcolor{srHeat!16}84.5/65.3 & 98.4 & \cellcolor{srHeat!32}96.9/95.9 & \cellcolor{srHeat!32}96.4/100.0 & \cellcolor{srHeat!32}97.4/96.9 \\
\quad high school macroeconomics & 390 & 85.9 & \cellcolor{srHeat!24}79.5/73.1 & \cellcolor{srHeat!32}74.4/99.2 & \cellcolor{srHeat!24}78.7/72.8 & 93.6 & \cellcolor{srHeat!16}91.3/60.0 & \cellcolor{srHeat!32}89.5/100.0 & \cellcolor{srHeat!16}91.5/61.3 \\
\quad high school microeconomics & 238 & 94.1 & \cellcolor{srHeat!24}85.7/74.8 & \cellcolor{srHeat!32}80.7/100.0 & \cellcolor{srHeat!24}83.2/75.2 & 96.2 & \cellcolor{srHeat!16}94.5/63.0 & \cellcolor{srHeat!32}94.5/100.0 & \cellcolor{srHeat!16}95.0/62.6 \\
\quad high school psychology & 545 & 91.9 & \cellcolor{srHeat!24}87.5/71.0 & \cellcolor{srHeat!32}86.4/99.8 & \cellcolor{srHeat!16}88.3/64.6 & 95.6 & \cellcolor{srHeat!32}94.7/96.3 & \cellcolor{srHeat!32}94.7/100.0 & \cellcolor{srHeat!32}94.7/96.3 \\
\quad human sexuality & 131 & 81.7 & \cellcolor{srHeat!24}76.3/78.6 & \cellcolor{srHeat!32}74.0/100.0 & \cellcolor{srHeat!16}76.3/68.7 & 89.3 & \cellcolor{srHeat!32}89.3/95.4 & \cellcolor{srHeat!32}88.5/100.0 & \cellcolor{srHeat!32}89.3/96.2 \\
\quad professional psychology & 612 & 75.3 & \cellcolor{srHeat!24}71.6/82.2 & \cellcolor{srHeat!32}71.2/99.8 & \cellcolor{srHeat!24}71.6/78.1 & 87.1 & \cellcolor{srHeat!32}83.3/97.4 & \cellcolor{srHeat!32}83.0/100.0 & \cellcolor{srHeat!32}83.3/97.4 \\
\quad public relations & 110 & 71.8 & \cellcolor{srHeat!24}67.3/81.8 & \cellcolor{srHeat!32}64.5/100.0 & \cellcolor{srHeat!24}66.4/70.0 & 71.8 & \cellcolor{srHeat!32}75.5/100.0 & \cellcolor{srHeat!32}75.5/100.0 & \cellcolor{srHeat!32}74.5/100.0 \\
\quad security studies & 245 & 71.8 & \cellcolor{srHeat!32}66.5/95.5 & \cellcolor{srHeat!32}69.4/100.0 & \cellcolor{srHeat!32}66.9/94.3 & 79.6 & \cellcolor{srHeat!32}77.6/99.6 & \cellcolor{srHeat!32}78.8/100.0 & \cellcolor{srHeat!32}78.0/99.6 \\
\quad sociology & 201 & 83.1 & \cellcolor{srHeat!32}84.1/94.5 & \cellcolor{srHeat!32}81.6/100.0 & \cellcolor{srHeat!24}81.6/89.6 & 90.5 & \cellcolor{srHeat!32}88.6/100.0 & \cellcolor{srHeat!32}87.6/100.0 & \cellcolor{srHeat!32}88.6/99.5 \\
\quad us foreign policy & 100 & 81.0 & \cellcolor{srHeat!24}82.0/89.0 & \cellcolor{srHeat!32}78.0/100.0 & \cellcolor{srHeat!24}84.0/80.0 & 92.0 & \cellcolor{srHeat!32}89.0/99.0 & \cellcolor{srHeat!32}90.0/100.0 & \cellcolor{srHeat!32}89.0/98.0 \\
\addlinespace[1pt]
\rowcolor{gray!12}\textbf{STEM} & 3153 & \textbf{81.9} & \textbf{\cellcolor{srHeat!8}83.0/34.3} & \textbf{\cellcolor{srHeat!24}73.2/88.8} & \textbf{\cellcolor{srHeat!8}83.1/33.3} & \textbf{89.9} & \textbf{\cellcolor{srHeat!8}91.3/36.4} & \textbf{\cellcolor{srHeat!32}84.0/97.7} & \textbf{\cellcolor{srHeat!8}91.6/38.1} \\
\quad abstract algebra & 100 & 72.0 & 73.0/0.0 & \cellcolor{srHeat!32}57.0/97.0 & 77.0/0.0 & 90.0 & 92.0/0.0 & \cellcolor{srHeat!32}72.0/100.0 & 92.0/0.0 \\
\quad anatomy & 135 & 72.6 & \cellcolor{srHeat!16}63.0/60.7 & \cellcolor{srHeat!32}54.8/100.0 & \cellcolor{srHeat!24}59.3/71.1 & 87.4 & \cellcolor{srHeat!24}80.7/83.0 & \cellcolor{srHeat!32}80.7/100.0 & \cellcolor{srHeat!24}80.7/88.9 \\
\quad astronomy & 152 & 84.9 & \cellcolor{srHeat!24}84.9/71.1 & \cellcolor{srHeat!32}77.6/100.0 & \cellcolor{srHeat!16}86.2/65.1 & 94.1 & \cellcolor{srHeat!24}94.7/78.9 & \cellcolor{srHeat!32}92.8/100.0 & \cellcolor{srHeat!24}94.1/80.9 \\
\quad college biology & 144 & 91.7 & \cellcolor{srHeat!24}84.7/82.6 & \cellcolor{srHeat!32}81.9/100.0 & \cellcolor{srHeat!24}84.0/84.7 & 98.6 & \cellcolor{srHeat!24}97.2/83.3 & \cellcolor{srHeat!32}93.8/100.0 & \cellcolor{srHeat!24}97.9/84.7 \\
\quad college chemistry & 100 & 51.0 & \cellcolor{srHeat!8}63.0/27.0 & \cellcolor{srHeat!32}52.0/100.0 & \cellcolor{srHeat!8}60.0/25.0 & 67.0 & \cellcolor{srHeat!8}71.0/14.0 & \cellcolor{srHeat!32}66.0/100.0 & \cellcolor{srHeat!8}73.0/21.0 \\
\quad college computer science & 100 & 69.0 & \cellcolor{srHeat!8}77.0/25.0 & \cellcolor{srHeat!32}68.0/98.0 & \cellcolor{srHeat!8}77.0/26.0 & 85.0 & \cellcolor{srHeat!8}91.0/25.0 & \cellcolor{srHeat!32}81.0/100.0 & \cellcolor{srHeat!8}94.0/29.0 \\
\quad college mathematics & 100 & 64.0 & 71.0/0.0 & \cellcolor{srHeat!32}61.0/90.0 & 78.0/0.0 & 76.0 & 86.0/0.0 & \cellcolor{srHeat!32}66.0/97.0 & 85.0/0.0 \\
\quad college physics & 102 & 81.4 & \cellcolor{srHeat!8}87.3/15.7 & \cellcolor{srHeat!32}62.7/98.0 & \cellcolor{srHeat!8}91.2/16.7 & 91.2 & \cellcolor{srHeat!8}93.1/10.8 & \cellcolor{srHeat!32}73.5/100.0 & 94.1/8.8 \\
\quad computer security & 100 & 81.0 & \cellcolor{srHeat!24}81.0/70.0 & \cellcolor{srHeat!32}82.0/100.0 & \cellcolor{srHeat!16}80.0/62.0 & 84.0 & \cellcolor{srHeat!24}89.0/84.0 & \cellcolor{srHeat!32}85.0/100.0 & \cellcolor{srHeat!24}89.0/83.0 \\
\quad conceptual physics & 235 & 88.1 & \cellcolor{srHeat!16}81.3/46.4 & \cellcolor{srHeat!32}77.0/99.6 & \cellcolor{srHeat!8}81.3/36.2 & 94.9 & \cellcolor{srHeat!8}94.5/38.7 & \cellcolor{srHeat!32}93.2/100.0 & \cellcolor{srHeat!8}94.0/37.9 \\
\quad electrical engineering & 145 & 75.2 & \cellcolor{srHeat!16}73.1/49.0 & \cellcolor{srHeat!32}68.3/98.6 & \cellcolor{srHeat!16}69.7/53.1 & 84.8 & \cellcolor{srHeat!16}84.8/43.4 & \cellcolor{srHeat!32}82.1/100.0 & \cellcolor{srHeat!16}84.8/44.8 \\
\quad elementary mathematics & 378 & 96.8 & 96.0/0.0 & \cellcolor{srHeat!16}81.0/68.3 & 95.5/0.0 & 97.1 & 97.6/3.7 & \cellcolor{srHeat!32}89.4/99.7 & 97.4/5.3 \\
\quad high school biology & 310 & 92.6 & \cellcolor{srHeat!24}87.4/84.5 & \cellcolor{srHeat!32}86.8/100.0 & \cellcolor{srHeat!24}87.7/82.9 & 95.8 & \cellcolor{srHeat!24}95.2/85.8 & \cellcolor{srHeat!32}94.8/100.0 & \cellcolor{srHeat!24}95.5/89.4 \\
\quad high school chemistry & 203 & 75.4 & \cellcolor{srHeat!8}84.7/31.0 & \cellcolor{srHeat!32}70.4/98.5 & \cellcolor{srHeat!8}87.2/28.1 & 89.7 & \cellcolor{srHeat!8}91.1/20.2 & \cellcolor{srHeat!32}81.3/99.5 & \cellcolor{srHeat!8}91.1/21.2 \\
\quad high school computer science & 100 & 92.0 & \cellcolor{srHeat!8}91.0/33.0 & \cellcolor{srHeat!32}83.0/97.0 & \cellcolor{srHeat!8}91.0/27.0 & 96.0 & \cellcolor{srHeat!16}96.0/42.0 & \cellcolor{srHeat!32}92.0/100.0 & \cellcolor{srHeat!16}96.0/43.0 \\
\quad high school mathematics & 270 & 77.4 & 89.6/0.0 & \cellcolor{srHeat!8}76.3/37.4 & 91.1/0.0 & 84.4 & 92.2/0.0 & \cellcolor{srHeat!24}74.8/75.6 & 93.0/0.0 \\
\quad high school physics & 151 & 79.5 & \cellcolor{srHeat!8}82.1/18.5 & \cellcolor{srHeat!32}66.2/96.7 & \cellcolor{srHeat!8}81.5/19.9 & 87.4 & 89.4/7.3 & \cellcolor{srHeat!32}79.5/98.7 & \cellcolor{srHeat!8}91.4/11.3 \\
\quad high school statistics & 216 & 81.5 & \cellcolor{srHeat!8}85.2/15.7 & \cellcolor{srHeat!24}75.5/85.2 & \cellcolor{srHeat!8}83.8/16.2 & 88.0 & \cellcolor{srHeat!8}93.1/31.5 & \cellcolor{srHeat!32}86.6/100.0 & \cellcolor{srHeat!8}91.7/34.3 \\
\quad machine learning & 112 & 75.9 & \cellcolor{srHeat!8}75.0/30.4 & \cellcolor{srHeat!32}58.0/98.2 & \cellcolor{srHeat!8}70.5/32.1 & 88.4 & \cellcolor{srHeat!16}78.6/58.9 & \cellcolor{srHeat!32}74.1/100.0 & \cellcolor{srHeat!16}79.5/59.8 \\
\addlinespace[1pt]
\rowcolor{gray!12}\textbf{Other} & 3107 & \textbf{78.4} & \textbf{\cellcolor{srHeat!16}74.9/68.1} & \textbf{\cellcolor{srHeat!32}72.9/98.5} & \textbf{\cellcolor{srHeat!16}75.0/63.9} & \textbf{87.7} & \textbf{\cellcolor{srHeat!24}85.2/89.5} & \textbf{\cellcolor{srHeat!32}83.9/100.0} & \textbf{\cellcolor{srHeat!24}85.0/89.7} \\
\quad business ethics & 100 & 79.0 & \cellcolor{srHeat!16}69.0/43.0 & \cellcolor{srHeat!32}72.0/100.0 & \cellcolor{srHeat!16}71.0/45.0 & 81.0 & \cellcolor{srHeat!32}82.0/99.0 & \cellcolor{srHeat!32}81.0/100.0 & \cellcolor{srHeat!32}81.0/99.0 \\
\quad clinical knowledge & 265 & 81.9 & \cellcolor{srHeat!24}74.3/79.2 & \cellcolor{srHeat!32}73.2/99.2 & \cellcolor{srHeat!24}76.2/79.2 & 89.4 & \cellcolor{srHeat!24}87.2/88.7 & \cellcolor{srHeat!32}86.0/100.0 & \cellcolor{srHeat!24}86.8/89.8 \\
\quad college medicine & 173 & 79.2 & \cellcolor{srHeat!16}78.0/62.4 & \cellcolor{srHeat!32}75.7/99.4 & \cellcolor{srHeat!16}80.3/63.0 & 83.8 & \cellcolor{srHeat!16}85.0/67.1 & \cellcolor{srHeat!32}81.5/100.0 & \cellcolor{srHeat!16}82.7/68.8 \\
\quad global facts & 100 & 47.0 & \cellcolor{srHeat!24}44.0/73.0 & \cellcolor{srHeat!32}41.0/100.0 & \cellcolor{srHeat!8}43.0/27.0 & 58.0 & \cellcolor{srHeat!32}59.0/99.0 & \cellcolor{srHeat!32}57.0/100.0 & \cellcolor{srHeat!32}58.0/99.0 \\
\quad human aging & 223 & 71.3 & \cellcolor{srHeat!24}68.6/82.5 & \cellcolor{srHeat!32}68.6/100.0 & \cellcolor{srHeat!16}69.1/66.4 & 84.3 & \cellcolor{srHeat!32}82.1/100.0 & \cellcolor{srHeat!32}83.0/100.0 & \cellcolor{srHeat!32}81.2/100.0 \\
\quad management & 103 & 88.3 & \cellcolor{srHeat!24}79.6/88.3 & \cellcolor{srHeat!32}79.6/100.0 & \cellcolor{srHeat!24}79.6/77.7 & 90.3 & \cellcolor{srHeat!32}88.3/100.0 & \cellcolor{srHeat!32}87.4/100.0 & \cellcolor{srHeat!32}88.3/99.0 \\
\quad marketing & 234 & 88.9 & \cellcolor{srHeat!24}87.2/85.9 & \cellcolor{srHeat!32}86.8/100.0 & \cellcolor{srHeat!24}86.8/82.9 & 93.6 & \cellcolor{srHeat!32}93.6/98.7 & \cellcolor{srHeat!32}93.6/100.0 & \cellcolor{srHeat!32}93.6/98.3 \\
\quad medical genetics & 100 & 85.0 & \cellcolor{srHeat!16}82.0/67.0 & \cellcolor{srHeat!32}76.0/100.0 & \cellcolor{srHeat!16}79.0/69.0 & 97.0 & \cellcolor{srHeat!24}96.0/89.0 & \cellcolor{srHeat!32}95.0/100.0 & \cellcolor{srHeat!24}96.0/87.0 \\
\quad miscellaneous & 783 & 86.5 & \cellcolor{srHeat!16}83.1/50.2 & \cellcolor{srHeat!32}81.1/99.9 & \cellcolor{srHeat!16}83.9/43.2 & 94.4 & \cellcolor{srHeat!32}91.3/91.7 & \cellcolor{srHeat!32}90.5/100.0 & \cellcolor{srHeat!32}91.1/92.7 \\
\quad nutrition & 306 & 77.1 & \cellcolor{srHeat!24}73.5/83.7 & \cellcolor{srHeat!32}70.9/99.7 & \cellcolor{srHeat!24}72.2/80.1 & 91.8 & \cellcolor{srHeat!32}85.3/94.1 & \cellcolor{srHeat!32}85.3/100.0 & \cellcolor{srHeat!32}85.0/95.1 \\
\quad professional accounting & 282 & 71.6 & \cellcolor{srHeat!16}69.5/41.1 & \cellcolor{srHeat!24}62.4/85.1 & \cellcolor{srHeat!16}66.3/45.0 & 85.5 & \cellcolor{srHeat!16}80.9/56.4 & \cellcolor{srHeat!32}72.3/100.0 & \cellcolor{srHeat!16}81.9/54.3 \\
\quad professional medicine & 272 & 80.9 & \cellcolor{srHeat!24}76.8/83.8 & \cellcolor{srHeat!32}74.6/100.0 & \cellcolor{srHeat!32}76.8/95.2 & 93.0 & \cellcolor{srHeat!32}91.2/94.9 & \cellcolor{srHeat!32}89.3/100.0 & \cellcolor{srHeat!32}91.2/95.2 \\
\quad virology & 166 & 47.0 & \cellcolor{srHeat!24}48.8/88.0 & \cellcolor{srHeat!32}48.8/100.0 & \cellcolor{srHeat!24}49.4/80.1 & 55.4 & \cellcolor{srHeat!32}51.8/98.2 & \cellcolor{srHeat!32}56.0/100.0 & \cellcolor{srHeat!32}53.6/97.0 \\
\addlinespace[1pt]
\end{longtable}
\vspace{-0.5em}
\noindent\begin{minipage}{\textwidth}
\footnotesize
\textit{Note.} Area labels follow the MMLU paper's four supercategories: Humanities, Social Sciences, STEM, and Other. The rows below each area list the original MMLU subjects in that area. Color encodes \ETR{} only; accuracy is printed numerically.
\end{minipage}
\endgroup

\paragraph{Interpretation of Table~\ref{tab:mmlu-subject-breakdown-app}.}
The subject-level rows show the same split: quantitative subjects are harder to compress, while many fact-like subjects reach high \ETR{} without the same accuracy cost.

\paragraph{Surface perturbations.}
Table~\ref{tab:surface-control-app} checks whether the MMLU pattern can be explained by label names or option order.
Cons. is the percentage of paired base and perturbed outputs whose nonempty predictions agree after the perturbed labels are mapped back to the original choices. All matched pairs enter the denominator; an empty prediction on either side counts as inconsistent.

\begin{table}[H]
\centering
\scriptsize
\setlength{\tabcolsep}{3.0pt}
\renewcommand{\arraystretch}{1.07}
\caption{MMLU surface perturbations with the answer set fixed. Deltas are relative to the base version; darker colored cells indicate larger absolute changes. \legacyM{} denotes a user-message answer prefix.}
\label{tab:surface-control-app}
\resizebox{\textwidth}{!}{%
\begin{tabular}{@{}lllrrrrrrr@{}}
\toprule
\textbf{Model} & \textbf{Mode} & \textbf{Variant}
& Acc. & \ETR{} & \QRel{} & $\Delta$Acc. & $\Delta$\ETR{} & $\Delta$\QRel{} & Cons. \\
\midrule
\textbf{Qwen3-4B} & Mode~2 & base & 74.2 & 53.0 & 0.369 & -- & -- & -- & -- \\
& & choice shuffle & 72.3 & 52.7 & 0.372 & \appaccshade{46}{-1.9} & \appsrshade{10}{-0.3} & \appqrelshade{6}{+0.003} & 82.5 \\
& & numeric labels & 73.4 & 45.0 & 0.426 & \appaccshade{24}{-0.8} & \appsrshade{62}{-8.0} & \appqrelshade{60}{+0.057} & 89.9 \\
\cmidrule(lr){2-10}
& Mode~5 & base & 68.9 & 97.0 & 0.023 & -- & -- & -- & -- \\
& & choice shuffle & 67.0 & 96.8 & 0.024 & \appaccshade{46}{-1.9} & \appsrshade{6}{-0.1} & \appqrelshade{5}{+0.001} & 80.7 \\
& & numeric labels & 68.5 & 96.8 & 0.024 & \appaccshade{12}{-0.4} & \appsrshade{6}{-0.1} & \appqrelshade{5}{+0.001} & 93.0 \\
\cmidrule(lr){2-10}
& \legacyM{} & base & 74.5 & 49.9 & 0.385 & -- & -- & -- & -- \\
& & choice shuffle & 72.5 & 50.4 & 0.383 & \appaccshade{48}{-2.0} & \appsrshade{12}{+0.5} & \appqrelshade{6}{-0.002} & 81.8 \\
& & numeric labels & 73.2 & 45.6 & 0.414 & \appaccshade{32}{-1.2} & \appsrshade{44}{-4.3} & \appqrelshade{42}{+0.029} & 88.7 \\
\midrule
\textbf{Qwen3-32B} & Mode~2 & base & 82.4 & 79.6 & 0.164 & -- & -- & -- & -- \\
& & choice shuffle & 82.1 & 79.0 & 0.168 & \appaccshade{10}{-0.3} & \appsrshade{14}{-0.6} & \appqrelshade{7}{+0.004} & 90.2 \\
& & numeric labels & 82.0 & 84.2 & 0.126 & \appaccshade{12}{-0.4} & \appsrshade{46}{+4.6} & \appqrelshade{46}{-0.038} & 95.8 \\
\cmidrule(lr){2-10}
& Mode~5 & base & 79.8 & 99.4 & 0.005 & -- & -- & -- & -- \\
& & choice shuffle & 79.0 & 99.4 & 0.005 & \appaccshade{24}{-0.8} & \appsrshade{4}{+0.0} & \appqrelshade{4}{+0.000} & 88.3 \\
& & numeric labels & 78.5 & 99.5 & 0.004 & \appaccshade{34}{-1.3} & \appsrshade{6}{+0.1} & \appqrelshade{5}{-0.001} & 94.7 \\
\cmidrule(lr){2-10}
& \legacyM{} & base & 82.4 & 79.6 & 0.162 & -- & -- & -- & -- \\
& & choice shuffle & 82.0 & 79.1 & 0.165 & \appaccshade{12}{-0.4} & \appsrshade{12}{-0.5} & \appqrelshade{7}{+0.003} & 90.1 \\
& & numeric labels & 81.8 & 84.3 & 0.125 & \appaccshade{18}{-0.6} & \appsrshade{48}{+4.7} & \appqrelshade{44}{-0.037} & 95.1 \\
\bottomrule
\end{tabular}%
}
\end{table}

\paragraph{Interpretation of Table~\ref{tab:surface-control-app}.}
The modest shifts in Acc./\ETR{}/\QRel{} and the high consistency argue against a label-name or option-order explanation.

\Needspace{20\baselineskip}
\paragraph{Candidate visibility.}
Table~\ref{tab:candidate-shown-app} rewrites MMLU items as yes/no verification with one candidate answer shown.

\begin{table}[H]
\centering
\scriptsize
\setlength{\tabcolsep}{2.8pt}
\renewcommand{\arraystretch}{1.07}
\caption{MMLU boolean verification with a proposed answer shown. Darker cells indicate larger metric values or larger changes from MCQ. \legacyM{} denotes a user-message answer prefix.}
\label{tab:candidate-shown-app}
\resizebox{\textwidth}{!}{%
\begin{tabular}{@{}llcccccccccc@{}}
\toprule
& & \multicolumn{3}{c}{\textbf{MCQ base}}
& \multicolumn{5}{c}{\textbf{Boolean verification}}
& \multicolumn{2}{c}{\textbf{Change}} \\
\cmidrule(lr){3-5} \cmidrule(lr){6-10} \cmidrule(l){11-12}
\textbf{Model} & \textbf{Mode}
& Acc. & \ETR{} & \QRel{}
& Acc. & \ETR{} & \QRel{} & Pos. & Neg.
& $\Delta$Acc. & $\Delta$\QRel{} \\
\midrule
\textbf{Qwen3-4B} & Mode~1 & 77.6 & 0.1 & 0.852 & \appaccshade{48}{80.0} & \appsrshade{30}{65.6} & \appqrelshade{62}{0.833} & 82.4 & 77.6 & \appaccshade{60}{+2.4} & \appqrelshade{10}{-0.019} \\
& Mode~2 & 75.0 & 52.8 & 0.370 & \appaccshade{36}{73.5} & \appsrshade{32}{67.2} & \appqrelshade{28}{0.235} & 65.5 & 81.6 & \appaccshade{38}{-1.5} & \appqrelshade{46}{-0.135} \\
& Mode~5 & 69.4 & 97.0 & 0.023 & \appaccshade{26}{69.1} & \appsrshade{64}{100.0} & \appqrelshade{4}{0.000} & 52.9 & 85.2 & \appaccshade{8}{-0.3} & \appqrelshade{12}{-0.023} \\
& \legacyM{} & 74.6 & 50.4 & 0.382 & \appaccshade{34}{72.9} & \appsrshade{40}{76.8} & \appqrelshade{24}{0.179} & 63.9 & 81.8 & \appaccshade{42}{-1.7} & \appqrelshade{60}{-0.203} \\
\midrule
\textbf{Qwen3-32B} & Mode~1 & 86.2 & 3.1 & 0.835 & \appaccshade{62}{84.6} & \appsrshade{58}{94.8} & \appqrelshade{60}{0.822} & 88.4 & 80.9 & \appaccshade{40}{-1.6} & \appqrelshade{14}{-0.013} \\
& Mode~2 & 82.6 & 79.9 & 0.162 & \appaccshade{52}{81.2} & \appsrshade{60}{96.9} & \appqrelshade{7}{0.026} & 81.1 & 81.2 & \appaccshade{36}{-1.4} & \appqrelshade{42}{-0.136} \\
& Mode~5 & 80.0 & 99.6 & 0.004 & \appaccshade{50}{80.6} & \appsrshade{64}{100.0} & \appqrelshade{4}{0.000} & 80.3 & 80.8 & \appaccshade{12}{+0.5} & \appqrelshade{5}{-0.004} \\
& \legacyM{} & 82.5 & 80.4 & 0.155 & \appaccshade{52}{81.2} & \appsrshade{60}{96.8} & \appqrelshade{7}{0.026} & 81.2 & 81.0 & \appaccshade{34}{-1.3} & \appqrelshade{40}{-0.129} \\
\bottomrule
\end{tabular}%
}
\end{table}

\paragraph{Interpretation of Table~\ref{tab:candidate-shown-app}.}
Under no-think conditions, Boolean verification raises \ETR{} and lowers \QRel{} relative to the MCQ form. The Pos./Neg. columns show that models both accept true proposals and reject false ones.

\paragraph{Removing options from numeric-source items.}
Table~\ref{tab:open-numeric-app} tests the effect of removing the visible candidate set while keeping the answer source numeric.

\begin{table}[H]
\centering
\scriptsize
\setlength{\tabcolsep}{2.5pt}
\renewcommand{\arraystretch}{1.07}
\caption{Matched numeric-source items under MCQ, boolean verification, and open generation. Darker cells indicate larger open-generation values or larger Open--MCQ accuracy drops. \legacyM{} denotes a user-message answer prefix.}
\label{tab:open-numeric-app}
\resizebox{\textwidth}{!}{%
\begin{tabular}{@{}llcccccccccccc@{}}
\toprule
& & \multicolumn{3}{c}{\textbf{MCQ}}
& \multicolumn{3}{c}{\textbf{Bool verification}}
& \multicolumn{5}{c}{\textbf{Open numeric}}
& \textbf{Drop} \\
\cmidrule(lr){3-5} \cmidrule(lr){6-8} \cmidrule(lr){9-13} \cmidrule(l){14-14}
\textbf{Model} & \textbf{Mode}
& Acc. & \ETR{} & \QRel{}
& Acc. & \ETR{} & \QRel{}
& Acc. & \ETR{} & \QRel{} & Parse & Exact
& Open--MCQ \\
\midrule
\textbf{Qwen3-4B} & Mode~1 & 66.8 & 0.0 & 0.832 & 82.0 & 13.5 & 0.847 & \appaccshade{22}{40.6} & \appsrshade{4}{0.0} & \appqrelshade{56}{0.841} & 99.6 & 38.3 & \appdropshade{60}{-26.2} \\
& Mode~2 & 72.1 & 3.7 & 0.760 & 81.2 & 12.1 & 0.722 & \appaccshade{38}{50.3} & \appsrshade{4}{0.0} & \appqrelshade{64}{0.811} & 97.0 & 47.2 & \appdropshade{50}{-21.8} \\
& Mode~5 & 53.8 & 63.4 & 0.281 & 60.4 & 99.4 & 0.005 & \appaccshade{34}{48.6} & \appsrshade{12}{13.4} & \appqrelshade{50}{0.678} & 98.0 & 45.8 & \appdropshade{18}{-5.2} \\
& \legacyM{} & 72.0 & 1.4 & 0.771 & 81.6 & 12.9 & 0.717 & \appaccshade{38}{50.6} & \appsrshade{4}{0.0} & \appqrelshade{64}{0.811} & 97.9 & 47.5 & \appdropshade{50}{-21.4} \\
\midrule
\textbf{Qwen3-32B} & Mode~1 & 72.7 & 0.2 & 0.821 & 81.2 & 44.8 & 0.841 & \appaccshade{26}{43.9} & \appsrshade{4}{0.0} & \appqrelshade{54}{0.832} & 99.8 & 41.6 & \appdropshade{66}{-28.9} \\
& Mode~2 & 77.8 & 10.5 & 0.709 & 78.9 & 62.5 & 0.314 & \appaccshade{42}{52.6} & \appsrshade{4}{0.0} & \appqrelshade{62}{0.810} & 98.0 & 49.0 & \appdropshade{58}{-25.2} \\
& Mode~5 & 52.9 & 94.1 & 0.046 & 69.6 & 100.0 & 0.000 & \appaccshade{22}{40.4} & \appsrshade{58}{82.7} & \appqrelshade{10}{0.119} & 98.0 & 38.0 & \appdropshade{30}{-12.4} \\
& \legacyM{} & 78.2 & 10.8 & 0.707 & 78.9 & 62.5 & 0.315 & \appaccshade{42}{52.3} & \appsrshade{4}{0.0} & \appqrelshade{62}{0.811} & 98.1 & 48.9 & \appdropshade{60}{-25.9} \\
\bottomrule
\end{tabular}%
}
\end{table}

\paragraph{Interpretation of Table~\ref{tab:open-numeric-app}.}
Removing the options requires the model to generate a numeric value. Compared with MCQ, open numeric generation under M5 has lower accuracy and higher \QRel{} for both models.
For Open numeric, Parse is the percentage of extracted answers that can be parsed as decimals, integer fractions, or percentages. Acc. compares their exact numeric values, while Exact compares answer strings after removing outer mathematical wrappers, normalizing whitespace, and removing commas. All three rates use all matched Open outputs as the denominator.

\clearpage
\section{User-Message-Prefix Results}
\label{app:historical-user-prefix}

\paragraph{User-message prefix.}
In \legacyM{}, an answer-prefix template is appended to the \emph{user message}, after which the model starts a new assistant response.
M6 instead supplies the prefix as an unfinished assistant turn that the model continues.

\begin{figure}[H]
\centering
\includegraphics[width=\linewidth,height=0.72\textheight,keepaspectratio]{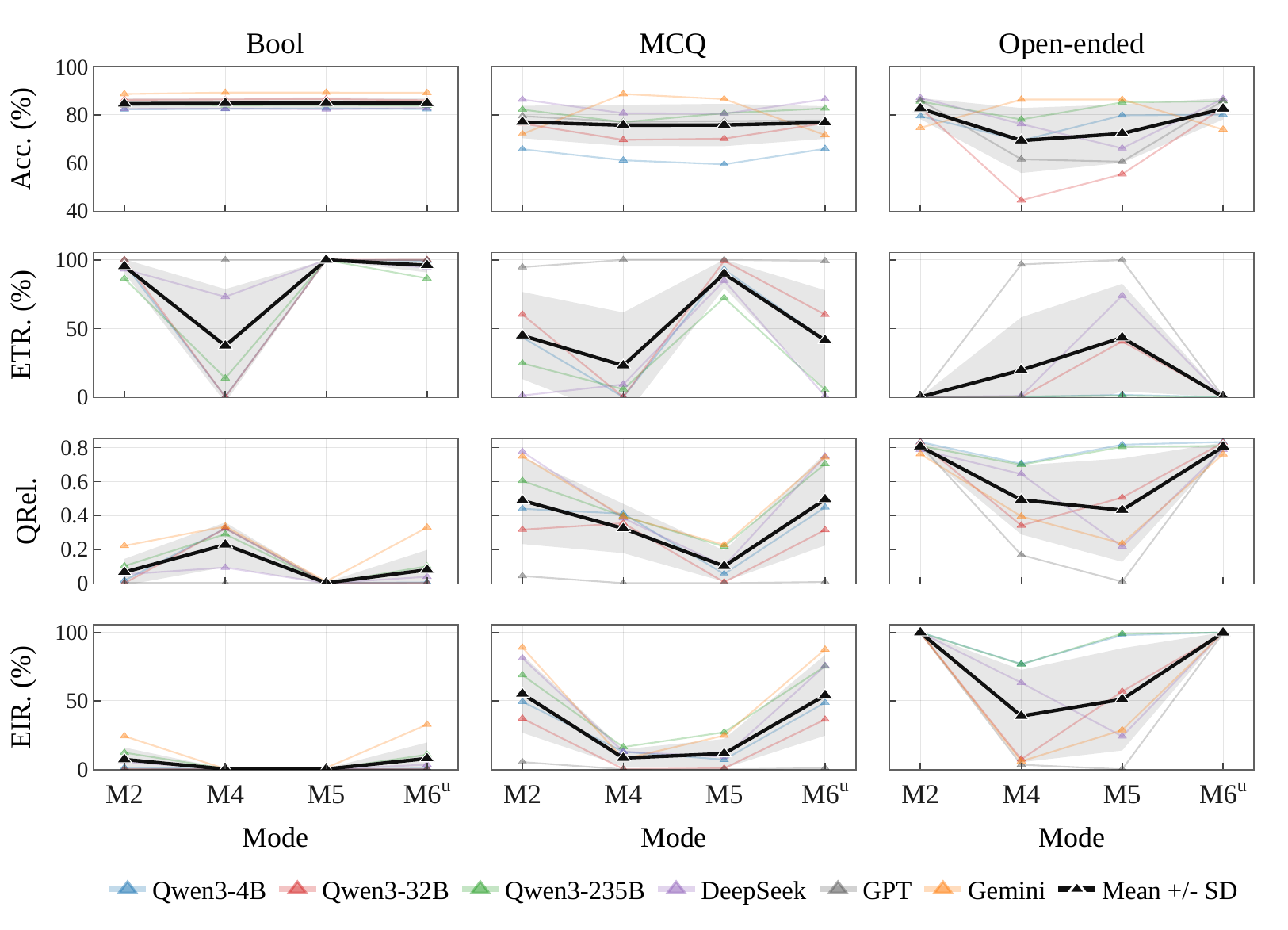}
\caption{\textbf{User-message-prefix results across answer spaces.} \legacyM{} supplies the answer prefix in the user message; M2/M4/M5 are the comparison modes. Colored lines show individual models; black lines and gray bands show their mean and one standard deviation.}
\label{fig:legacy-user-prefix}
\end{figure}

\paragraph{Accuracy comparison.}
Across M2, M4, M5, and \legacyM{}, Bool accuracy is relatively flat, MCQ accuracy varies more, and Open-ended tasks show the largest changes.
In the natural-language comparison from M2 to M5, greater answer-only compliance accompanied reduced accuracy on Open-ended tasks.
Under \legacyM{}, Open-ended accuracy returned toward M2 levels. Accuracy here uses the final answer selected by the pre-answer parser.

\paragraph{Response-behavior comparison.}
The user-message prefix did not strengthen answer-only compliance relative to M5: from M5 to \legacyM{}, mean \ETR{} fell on MCQ and Open-ended tasks, while mean \QRel{} and \EIR{} increased in all three families.
On Open-ended tasks, \ETR{} returned near zero and relevance, visible inference, and accuracy moved toward M2.
For \legacyM{}, \QRel{} and \EIR{} evaluate the text preceding the selected final answer. M6 uses the residual-text definition in Appendix~\ref{app:m6-prefill-protocol}, which also includes text after the first answer.

\paragraph{Experimental coverage.}
Table~\ref{tab:m6-legacy-scope-app} specifies the intervention and population used in each analysis.

\begin{table}[H]
\centering
\small
\setlength{\tabcolsep}{4pt}
\renewcommand{\arraystretch}{1.12}
\caption{\textbf{Interventions and evaluation scope.}}
\label{tab:m6-legacy-scope-app}
\begin{tabularx}{\linewidth}{@{}p{0.31\linewidth}X@{}}
\toprule
\textbf{Result block} & \textbf{Intervention and evaluation scope} \\
\midrule
Figure~\ref{fig:frontier-family-control}; Tables~\ref{tab:m6-format-audit-app}--\ref{tab:m6-format-examples-app} & Assistant-prefill M6: four models, 22,748 generations, first-box accuracy and residual-text metrics. \\
Table~\ref{tab:measurement-audit-app} & M1--M5 format/copy audit. \\
Tables~\ref{tab:cross-judge-eir-app}, \ref{tab:judge-paraphrase-audit-app}; native and M2/M5 comparisons & Native no-thinking and natural-language suppression comparisons under the pre-answer protocol. \\
Figure~\ref{fig:global-six-mode-app}; Table~\ref{tab:qwen-scaling-modes-app} & Global/scaling analyses with \legacyM{} as the user-message-prefix condition. \\
Tables~\ref{tab:domain-gradient-full-app}, \ref{tab:mmlu-subject-breakdown-app} & MMLU domain/subject analyses under the pre-answer protocol, including \legacyM{}; metric-specific sample sizes are given with the tables. \\
Tables~\ref{tab:surface-control-app}, \ref{tab:candidate-shown-app}, \ref{tab:open-numeric-app} & Surface, candidate, and numeric-source transformations under the pre-answer protocol, including \legacyM{}. \\
Table~\ref{tab:cross-judge-agreement-app}; encoder/human validation & Full-corpus encoder and judge comparisons: M1--M5 and \legacyM{}; human validation: the 600-output M2/M5 sample. \\
\bottomrule
\end{tabularx}
\end{table}

\subsection{Whole-Corpus Judge Agreement}
We compare the two judges on the 572,688-output corpus covering M1--M5 and \legacyM{}.
Opus returns usable labels for 572,645 outputs. The remaining 43 (0.0075\%) receive no usable label from the API and are excluded only from the direct agreement calculation.
Table~\ref{tab:cross-judge-agreement-app} reports agreement by model, benchmark, and mode. Five-way agreement requires identical rubric labels; binary agreement distinguishes explicit reasoning from all other categories.

\begin{table}[H]
\centering
\scriptsize
\setlength{\tabcolsep}{3.0pt}
\renewcommand{\arraystretch}{1.04}
\caption{\textbf{GPT-5.5--Opus 4.8 agreement for M1--M5 plus \legacyM{}.} Each group pools the other experimental dimensions. Percentages are calculated over outputs for which both judges have a usable label.}
\label{tab:cross-judge-agreement-app}
\begin{minipage}[t]{0.49\textwidth}
\vspace{0pt}
\centering
\begin{tabular}{@{}lrrr@{}}
\toprule
\textbf{Group} & \textbf{$N$} & \textbf{Five-way} & \shortstack{\textbf{Explicit vs.}\\\textbf{other}} \\
\midrule
\multicolumn{4}{@{}l}{\textbf{Overall}} \\
All outputs & 572,645 & 96.92\% & 97.30\% \\
Non-empty $T$ & 391,010 & 95.48\% & 96.05\% \\
\midrule
\multicolumn{4}{@{}l}{\textbf{Model}} \\
Qwen3-4B & 218,078 & 95.78\% & 96.23\% \\
Qwen3-32B & 218,079 & 98.47\% & 98.77\% \\
Qwen3-235B-A22B & 34,122 & 96.27\% & 96.43\% \\
DeepSeek-V4-Flash & 34,122 & 94.21\% & 94.78\% \\
GPT-5.4 & 34,122 & 99.54\% & 99.88\% \\
Gemini-3-Flash-Preview & 34,122 & 94.96\% & 95.56\% \\
\bottomrule
\end{tabular}
\end{minipage}
\hfill
\begin{minipage}[t]{0.49\textwidth}
\vspace{0pt}
\centering
\begin{tabular}{@{}lrrr@{}}
\toprule
\textbf{Group} & \textbf{$N$} & \textbf{Five-way} & \shortstack{\textbf{Explicit vs.}\\\textbf{other}} \\
\midrule
\multicolumn{4}{@{}l}{\textbf{Benchmark}} \\
BoolQ & 63,240 & 96.60\% & 98.02\% \\
StrategyQA & 24,732 & 96.30\% & 97.13\% \\
MMLU & 192,486 & 96.80\% & 96.88\% \\
MMLU-Pro & 168,359 & 96.05\% & 96.29\% \\
GSM8K & 39,828 & 99.62\% & 99.95\% \\
MATH & 84,000 & 98.04\% & 98.53\% \\
\midrule
\multicolumn{4}{@{}l}{\textbf{Mode}} \\
M1 & 95,433 & 98.91\% & 98.96\% \\
M2 & 95,448 & 96.24\% & 96.40\% \\
M3 & 95,434 & 92.93\% & 93.57\% \\
M4 & 95,434 & 98.69\% & 99.62\% \\
M5 & 95,448 & 99.30\% & 99.56\% \\
\legacyM{} & 95,448 & 95.43\% & 95.69\% \\
\bottomrule
\end{tabular}
\end{minipage}
\end{table}

Agreement remains high after removing empty-$T$ cases and within each model, benchmark, and mode; the matched \EIR{} rates in Table~\ref{tab:cross-judge-eir-app} reproduce the Level~3 ordering under the independent judge.

\Needspace{18\baselineskip}
\subsection{Whole-Corpus Format Audit}
This audit covers the M1--M5 plus \legacyM{} corpus.
Table~\ref{tab:measurement-audit-app} reports the M1--M5-only audit, and Table~\ref{tab:m6-format-audit-app} reports M6 response structures.
\begin{table}[H]
\centering
\scriptsize
\setlength{\tabcolsep}{3.2pt}
\renewcommand{\arraystretch}{1.08}
\caption{Whole-corpus format audit for M1--M5 plus \legacyM{}. Copy uses the same four-gram detector as Appendix~\ref{app:measurement-audits}. Answer-first requires the first box to occur after at most three lexical tokens and to be followed by at least three lexical tokens. Percentages for Copy use all successful responses; Answer-first percentages use boxed responses.}
\label{tab:legacy-measurement-audit-app}
\begin{tabular}{@{}lrrr@{}}
\toprule
\textbf{Model} & \textbf{Responses} & \textbf{Copy} & \textbf{Answer-first} \\
\midrule
Qwen3-4B & 218,100 & 235 (0.108\%) & 23/199,186 (0.012\%) \\
Qwen3-32B & 218,100 & 31 (0.014\%) & 27/205,146 (0.013\%) \\
Qwen3-235B-A22B & 34,122 & 4 (0.012\%) & 97/33,930 (0.286\%) \\
DeepSeek-V4-Flash & 34,122 & 19 (0.056\%) & 1/33,935 (0.003\%) \\
GPT-5.4 & 34,122 & 0 (0.000\%) & 0/34,104 (0.000\%) \\
Gemini-3-Flash-Preview & 34,122 & 21 (0.062\%) & 1/28,907 (0.003\%) \\
\midrule
\textbf{All} & \textbf{572,688} & \textbf{310 (0.054\%)} & \textbf{149/535,208 (0.028\%)} \\
\bottomrule
\end{tabular}
\end{table}

\paragraph{Paraphrase-only outputs.}
The primary judge classifies 655 outputs as paraphrase-only in the M1--M5 plus \legacyM{} corpus, including 477 from M1--M5.
These outputs do not contribute to \EIR{}, which counts only the explicit-reasoning category.